\documentclass[twoside,11pt]{article}

\usepackage[T1]{fontenc}
\usepackage{lmodern}

\usepackage{amsmath,amssymb,amsfonts,amsthm}
\usepackage{mathtools}
\usepackage{bm}

\usepackage{graphicx}
\usepackage{booktabs}
\usepackage{enumitem}
\usepackage{xcolor}

\usepackage{authblk}

\usepackage[a4paper,margin=3cm]{geometry}

\newcommand{\keywords}[1]{%
\par\noindent\textbf{Keywords:} #1}

\usepackage{hyperref}
\hypersetup{
colorlinks=true,
linkcolor=blue,
citecolor=blue,
urlcolor=blue,
anchorcolor=blue,
linkcolor=blue,
}

\usepackage{multirow}
\usepackage{xcolor}
\usepackage{comment}

\usepackage[round]{natbib}

\DeclareMathOperator{\dom}{dom}

\newcommand{\R}{\mathbb R}

\newcommand{\T}{\mathcal{T}}
\newcommand{\X}{\mathcal{X}}
\newcommand{\bx}{\mathbf{x}}
\newcommand{\by}{\mathbf{y}}

\newcommand{\bA}{\mathbf{A}}
\newcommand{\bb}{\mathbf{b}}
\newcommand{\bc}{\mathbf{c}}
\newcommand{\bt}{\mathbf{t}}

\newcommand{\bM}{\mathbf{M}}
\newcommand{\bu}{\mathbf{u}}
\newcommand{\bv}{\mathbf{v}}
\newcommand{\btheta}{\boldsymbol{\theta}}

\newcommand{\linOperator}{\mathbf{K}}
\newcommand{\objFun}{\mathcal{J}}

\newcommand{\inte}{\operatorname{int}}
\newcommand{\ip}[2]{\left\langle #1,#2\right\rangle}

\theoremstyle{plain}

\newtheorem{theorem}{Theorem}
\newtheorem{lemma}{Lemma}
\newtheorem{proposition}{Proposition}
\newtheorem{corollary}[theorem]{Corollary}
\newtheorem{definition}{Definition}

\theoremstyle{definition}
\newtheorem{remark}{Remark}

\usepackage{cleveref}

\crefname{remark}{remark}{remarks}
\Crefname{remark}{Remark}{Remarks}
\crefname{assumption}{assumption}{assumptions}
\Crefname{assumption}{Assumption}{Assumptions}

\crefname{appendix}{appendix}{appendices}
\Crefname{appendix}{Appendix}{Appendices}

\usepackage{comment}

\begin{document}

\title{Bregman geometry in regular exponential families: \\dual representations and relative smoothness}

\author[1]{Mame Diarra Fall}
\author[2]{Eric Barat}
\author[3]{Nicolas Dobigeon}

\affil[1]{Univ. Rouen Normandie, LITIS, Rouen, France}
\affil[2]{Univ. Paris-Saclay, CEA, List, Palaiseau, France}
\affil[3]{Univ. Toulouse, IRIT/INP-ENSEEIHT, Toulouse, France}

\date{}

\maketitle

\begin{abstract}%
Regular exponential-family distributions are ubiquitous in statistics due to their appealing and convenient properties. Notably, they can be expressed in terms of Bregman divergences, revealing fundamental connections between statistics, convex analysis, and information geometry. However, this geometric structure has not been systematically exploited to characterize these distributions with respect to their expectation parametrization. Building on a projective characterization, we establish a general construction of equivalent dual Bregman representations, together with normalization and uniqueness results. These representations express log-densities as ordinary Bregman divergences in dual coordinate systems, providing geometric characterizations of the parameter space. So-called $\beta$- and $\alpha$-scaled dual representations emerge as particular instances of this framework. We further derive dual Bregman representations of conjugate priors and establish that they preserve conjugacy. Applications to classical distributions illustrate the geometric structures underlying seemingly different statistical models. Moreover, we show that these representations provide natural reference geometries for establishing relative smoothness of the primal Bregman representation of the distribution, with explicit bounds that are sharp in the scalar and separable settings. An illustrative application to the Poisson distribution sheds new light on several works in the literature, providing complementary geometric justifications for reference geometries while deriving existing algorithms.
\end{abstract}%


\keywords
Bregman divergences, regular exponential families, Legendre duality, relative smoothness, information geometry    


\section{Introduction}

Exponential families constitute one of the most important classes of statistical models. Their broad applicability stems from a combination of analytical and statistical properties, including tractable likelihoods, finite-dimensional sufficient statistics, and the existence of natural parametrizations in terms of both natural and expectation parameters \citep{barndorff2014information,efron2023exponential}. They encompass many classical distributions, including Gaussian, Poisson, Bernoulli, Binomial, Gamma, and Beta distributions, and play a central role in statistical inference, information geometry, and machine learning.

An important connection between regular exponential families and convex analysis is provided by Bregman divergences. It is well established that, under standard regularity and minimality assumptions, the log-density of a regular exponential-family distribution can be expressed in terms of a Bregman divergence generated by the convex conjugate of the log-partition function \citep{Banerjee2005}. This representation provides a geometric viewpoint of exponential families and has proved useful well beyond statistical interpretation. In particular, Bregman divergences constitute fundamental tools in optimization and statistical learning, where they underlie generalized projection methods, mirror descent, Bregman proximal algorithms, and a variety of estimation and learning procedures \citep{bregman1967relaxation,bauschke1997legendre}. The Bregman representation of an exponential-family distribution also endows its data space with a natural geometry, with the sufficient statistics playing the role of geometric coordinates.

This representation, however, reveals an important asymmetry between the two arguments of the Bregman divergence. In the canonical representation, the first argument corresponds to the sufficient statistics, whereas the second one is associated with the expectation parameter. Convexity with respect to the first argument is guaranteed by the convexity of the Bregman generator. In contrast, convexity with respect to the second argument is not generally guaranteed. Thus, although the Bregman divergence naturally describes the geometry of the data space, its reverse form does not necessarily define an ordinary Bregman geometry on the expectation-parameter space. This difficulty is particularly pronounced when conducting asymptotically exact data augmentation \citep{Vono2020} and deriving splitting-based Monte Carlo algorithms \citep{vono2019split} with non-Euclidean and, in particular, Bregman kernels. With respect to these two tasks, respectively, it has recently been overcome by augmenting the statistical model with the same divergence evaluated with swapped arguments \citep{Faye2026} or, equivalently, by replacing the Bregman divergences by symmetrized counterparts defined as Jensen-Bregman divergences \citep{Dobigeon2026}. More generally and beyond these purely modeling and algorithmic workarounds, this divergence asymmetry raises a more fundamental question: can a reverse Bregman divergence itself be represented as an ordinary Bregman divergence after a suitable transformation of the coordinates? Then, can the Bregman geometry underlying a regular exponential-family distribution be transferred to its parameter space through an appropriate dual coordinate system? From an information geometry perspective, this question can be formulated as the search for a geometric structure underlying conformal divergences as studied by \cite{amari2000methods} and \cite{Nock2015}.

The objective of this paper is to answer this question. A sound, albeit incomplete, response is naturally provided by leveraging the role of Legendre duality in the geometry of exponential families \citep{rockafellar1997convex}. However, the term \emph{duality}, used hereafter and underlying the main rationale of this work, refers to a property intended to go beyond the sole Legendre duality. Thus, starting from the \emph{primal} Bregman representation of regular exponential-family distributions, we build on a projective characterization providing necessary and sufficient conditions for the existence of such dual Bregman representations. We further establish a general construction of the corresponding dual coordinates and Bregman generators, together with normalization and uniqueness results. Among admissible representations, we exhibit two particular models, distinguished by the form of an associated scaling function and, consequently, by the resulting Bregman generator and parameter coordinates. The first one, referred to as the $\beta$-scaled dual Bregman representation, recovers the self-evident representation induced by Legendre duality in exponential families. In this case, the resulting geometry is determined by the gradient of the original Bregman generator together with a scaled version of its Legendre conjugate. The second class, referred to as the $\alpha$-scaled dual Bregman representation, relies on a different transformed coordinate system and an associated Bregman generator. Both constructions arise as particular instances of the general projective framework. These two constructions therefore provide distinct but equivalent geometric descriptions of the same exponential-family distribution. In particular, different parametrizations of the distribution can induce different Bregman geometries on its parameter space while remaining equivalent at the level of the underlying probability model.

An important application of the proposed framework concerns conjugate priors for exponential families. By applying the two dual Bregman representations previously identified to the conjugate distributions associated with regular exponential families, we obtain alternative geometric representations of these priors. These representations differ from the primal Bregman formulation of conjugate priors established by \cite{Agarwal2010} in both their coordinate systems and associated Bregman generators, while retaining the defining conjugacy property under Bayesian updating. In this sense, the proposed framework does not modify the underlying probabilistic structure of conjugate priors, but provides alternative geometric descriptions of the same distributions. The resulting framework also recovers, as particular cases, dual Bregman representations associated with several classical statistical models and their conjugate priors. These examples illustrate how apparently different models can share common geometric structures and provide concrete interpretations of the proposed constructions.

Finally, the proposed framework also leads to consequences beyond the sole construction of these alternative Bregman representations. It is well known that relative smoothness generalizes the classical Lipschitz-gradient condition by measuring the variation of a function relative to a reference convex function, and has become an important notion in modern Bregman-based optimization \citep{Bauschke2017,lu2018relatively}. We show that the geometries induced by the $\beta$- and $\alpha$-scaled  models identified in this work provide natural reference geometries for establishing relative smoothness of the negative log-density associated with the original, primal Bregman representation. These results are extended to separable multidimensional models, thereby covering a broad class of statistical models with component-wise structure. For the geometries induced by both models, we derive explicit relative-smoothness bounds, with sharp constants in the scalar and separable settings. These results establish a direct connection between the geometry of exponential-family distributions and the convex-analytic notion of relative smoothness, which is central to modern Bregman-based optimization.

Addressing Poisson inverse problems leads to useful illustrations in which the proposed geometric framework provides a principled interpretation of choices that have appeared in previous optimization and statistical-learning approaches.

The main contributions of this work can be summarized as follows:

\begin{itemize}
\item Building on the projective characterization stated in \Cref{thm:proj-characterization}, which provides necessary and sufficient conditions for a reverse Bregman divergence to admit an ordinary Bregman representation in transformed coordinates, \Cref{cor:normalization-uniqueness} clarifies the normalization freedom and the associated uniqueness properties. Assuming the existence of admissible dual Bregman representations, \Cref{prop:matrix-reconstruction} then derives reconstruction formulas of such representations; 

\item \Cref{thm:dual-construction} then establishes the existence and uniqueness of dual Bregman representations for regular minimal exponential families, with explicit construction formulas given in \Cref{cor:explicit-construction}. \Cref{cor:dual-readings} specifies how these dual representations translate to the log-density of regular exponential family. \Cref{cor:beta_existence,cor:alpha_existence} identifies two classes of admissible dual representations, corresponding to the so-called $\beta$- and $\alpha$-scaled constructions, as particular instances of the general projective framework;

\item Starting from \Cref{prop:weighted:bregman:identity}, which establishes the primal representation of posterior distributions resulting from conjugate priors, \Cref{cor:prior_beta,prop:alternative} show that the $\beta$- and $\alpha$-scaled dual Bregman representations can be instantiated to derive as alternatives of these priors, while preserving conjugacy properties;

\item \Cref{tab:geometry} reports the main geometrical characteristics of these two particular representations for several classical exponential-family distributions, the corresponding derivations being detailed in Appendix \ref{appendix:additional-examples};

\item \Cref{prop:beta-relative-smoothness,prop:general-tau-tilde} show that both representations provide suitable reference geometries to ensure relative smoothness of the negative log-density associated with exponential-family distributions, with explicit bounds that are sharp in the scalar and separable settings;

\item We apply the proposed framework to Poisson inverse problems to recover several results from the literature. In particular, we show that Bregman gradient iterations under the geometries induced by the $\alpha$-scaled and $\beta$-scaled model recover the NoLips \citep{Bauschke2017} and the expectation-maximization \citep{shepp1982maximum} algorithms, respectively.

\end{itemize}

The remainder of the paper is organized as follows. \Cref{sec:background} recalls the necessary background on Legendre duality, Bregman geometry, relative smoothness and regular exponential families. \Cref{sec:dual_representations} states the projective characterization of dual Bregman representations and establishes their general construction, normalization and uniqueness properties. In particular, we instantiate these representations to exhibit dual representations of prior distributions required in a Bayesian framework and show that they naturally recover expected conjugacy properties. We also derive these representations for several classical exponential-family distributions. Section~\ref{sec:relative-smoothness-conjugacy} shows that these representations also provide suitable reference geometries to ensure relative smoothness properties. As an illustrative example, \Cref{sec:application} considers the case of Poisson inverse problems. Finally, Section~\ref{sec:conclusion} concludes the paper and discusses possible extensions.


\section{Background and preliminaries}
\label{sec:background}

\subsection{Legendre duality}
\label{subsec:legendre}

Notions of Legendre duality are recalled below following the terminology adopted by \cite{rockafellar1997convex}.

\begin{definition}[Legendre function]
Let $f:\mathcal{X} \subseteq \mathbb{R}^n\rightarrow \mathbb{R} \cup \{+\infty\}$ be a proper, lower semicontinuous, convex function. The function $f$ is said to be {essentially smooth} if it is differentiable throughout $\inte(\dom f)$ and $ \| \nabla f(x_k) \| \rightarrow +\infty $ whenever $x_k\in\inte(\dom f)$ converges to a boundary point of $\dom f$. Moreover, if $f$ is strictly convex on $\inte(\dom f)$, it is called a {Legendre function}. 
\end{definition}

\begin{definition}[Fenchel conjugate]
Let $f$ be a Legendre function with $\mathcal X=\inte(\dom f)$.  The Fenchel conjugate of $f$ is defined for $\btheta\in \dom  f^*$ by
\begin{equation*}
f^{*}(\btheta)
=
\sup_{\mathbf{x}\in \dom  f}
\{
\langle \btheta,\mathbf x\rangle-f(\mathbf x)
\}.
\label{eq:legendre}
\end{equation*}
The associated dual coordinate is
\[
\btheta=\nabla f(\mathbf x).
\]

\end{definition}

A Legendre function and its conjugate satisfy the following duality property.

\begin{proposition}[Legendre duality]\label{prop:legendre_duality}
    The function $f:\mathcal{X} \subseteq \mathbb{R}^n\rightarrow \mathbb{R} \cup \{+\infty\}$ is a Legendre function if and only if its  Fenchel conjugate  $f^*$ is also a Legendre function. Moreover, its gradient  $ \nabla f : \inte(\dom f) \rightarrow \inte(\dom f^*)$ is a bijection and satisfies the identity
    \[
\nabla f^{*}
=
(\nabla f)^{-1}
\]
or, equivalently,  $\forall \mathbf{x}\in \mathcal{X}$,  $\btheta = \nabla f(\mathbf{x}) \Leftrightarrow \mathbf{x} = \nabla f^*(\btheta)$. Moreover, one has
\[
f(\mathbf x)+f^{*}(\nabla f(\mathbf x))
=
\langle \mathbf x,\nabla f(\mathbf x)\rangle.
\]
\end{proposition}

Thus, for a Legendre function, the gradient establishes a one-to-one correspondence between the respective interiors of the effective domains of $f$ and $f^*$. These properties are central in the construction of divergences underlying the Bregman geometry.

\subsection{Bregman geometry}
\label{subsec:preliminaries_bregman}

\begin{definition}[Bregman divergences] \citep{bregman1967relaxation}
Let $f:\mathcal{X} \subseteq \mathbb{R}^n\rightarrow \mathbb{R} \cup \{+\infty\}$ be a proper convex function that is differentiable on
$\inte(\dom f)$. For $\mathbf{y}\in\dom f$ and $\mathbf{x}\in\inte(\dom f)$, the Bregman divergence induced by $f$ is defined as 
\begin{equation*} 
d_f(\mathbf{y},\mathbf{x}) = f(\mathbf{y})-f(\mathbf{x}) -\left\langle \nabla f(\mathbf{x}),\mathbf{y}-\mathbf{x}\right\rangle. \label{eq:bregman_divergence} 
\end{equation*} 
When $f$ is strictly convex, $d_f(\mathbf{y},\mathbf{x})\geq 0$, with 
\begin{equation*} 
d_f(\mathbf{y},\mathbf{x})=0 \quad\Leftrightarrow\quad \mathbf{y}=\mathbf{x}. 
\end{equation*}
\end{definition}

It is well-known that the Bregman divergences satisfy the duality identity recalled in the following proposition.

\begin{proposition}[Duality of Bregman divergences] \citep{bregman1967relaxation}
\label{prop:duality}
Let $f$ be a Legendre function. For every $\mathbf x,\mathbf y\in\inte(\dom f)$, one has
\begin{equation}
d_f(\mathbf y,\mathbf x)
=
d_{f^{*}}
(
\nabla f(\mathbf x),
\nabla f(\mathbf y)
).
\label{eq:dual}
\end{equation}
\end{proposition}

\subsection{Relative smoothness}
Relative smoothness  generalizes the definition of Lipschitz continuity of the gradient by replacing the Euclidean quadratic distance with a Bregman divergence \citep{Bauschke2017}.

\begin{definition}[Relative smoothness]\label{def:LC}
  Let $h$ be a Legendre function and let $g$ be a convex function, differentiable on
$\inte(\dom h)$, with
$\dom h\subseteq\dom g$, We say that the function $g$ is $L$-smooth relative to $h$ if $\exists L>0$ such that the function $Lh -g$ is convex on $\inte(\dom h)$.
\end{definition}
Interestingly, this so-called Lipschitz-like/convexity condition (LC) yields a Bregman descent lemma, as stated below.
\begin{proposition}\citep[Lemma 1]{Bauschke2017}\label{eq:bregman-descent}
Let $g$ be $L$-smooth relative to $h$. Then, for all
$\mathbf x,\mathbf{x}^\prime\in\inte(\dom h)$, one has
\begin{equation*}
g(\mathbf{x}^\prime)
\leq
g(\mathbf{x})
+
\left\langle
\nabla g(\mathbf{x}),
\mathbf x^\prime-\mathbf{x}
\right\rangle
+
L\,d_h(\mathbf x^\prime,\mathbf{x})
\end{equation*}
where $d_h$ is the Bregman divergence induced by the generator $h$.
\end{proposition}

While the Lipschitz continuity of the gradient imposes a Euclidean curvature bound, relative smoothness allows the Euclidean geometry to be replaced by a geometry adapted to the problem. In particular, when designing descent-based algorithms, the conventional yet restrictive gradient Lipschitz condition can be lifted by requiring instead that the curvature of the objective be controlled relative to that of a suitable reference function. For instance, one of the main algorithmic consequences of relative smoothness can be formulated as follows: for any $\gamma>0$ such that $\gamma\leq 1/L$, the Bregman gradient update \citep{auslender2006interior}
$$
\mathbf x^{+}
\in
\underset{\mathbf u}{\operatorname{argmin}}
\left\{
\langle \nabla g(\mathbf x),\mathbf u-\mathbf x\rangle
+\frac{1}{\gamma}d_h(\mathbf u,\mathbf x)
\right\}
$$
satisfies 
$$
g(\mathbf x^{+})
\leq
g(\mathbf x)
-
\left(\frac{1}{\gamma}-L\right)
d_h(\mathbf x^{+},\mathbf x).
$$
In particular, when $\gamma<1/L$, the objective decreases monotonically along the iterates, while $\gamma=1/L$ guarantees non-increase. This result is the Bregman analogue of the classical descent guarantee for gradient descent under a Lipschitz-gradient assumption: the Euclidean quadratic term $\frac{L}{2}\|\mathbf{u}-\mathbf{x}\|^2$ is replaced by the more general Bregman divergence $L d_h(\mathbf{u},\mathbf{x})$. Consequently, identifying a suitable reference geometry $h$ and establishing a relative-smoothness constant $L$ directly provides a valid step-size range for Bregman gradient or proximal-gradient methods. For instance, this formalism has proved particularly useful to solve estimation problems with Poisson distributed data, where it provides Bregman descent inequalities without requiring globally Lipschitz gradients. For Poisson linear inverse problems, Burg's entropy provides a suitable reference function $h$ satisfying the relative-smoothness condition \citep{Bauschke2017}. This geometry has subsequently been exploited in Poisson imaging problems \citep{Hurault2023,tibshirani2025laplace,Daniele2026} and Kullback-Leibler regression \citep{chirinos2026linear}.
These examples illustrate the importance of choosing a reference geometry adapted to the statistical model. In this work, we investigate such geometric structures in the broader setting of regular minimal exponential families, whose Bregman representations provide the starting point for our analysis.

\subsection{Regular exponential families}
\label{subsec:bregman}
In this paper, we consider regular exponential families whose natural parametrization is defined below.
Let $\boldsymbol\omega$ denote an observation and let $\mathbf y=T(\boldsymbol\omega)\in\mathbb R^n$ be its sufficient statistic.

\begin{definition}[Natural parametrization]
\label{def:natural_param_expo}

An exponential family admits the natural para-metrization
\begin{equation}
\label{eq:natural-exponential-family}
p(\mathbf y;\btheta) = \exp\left(\langle\btheta,\mathbf y\rangle -\psi(\btheta) \right)p_0(\mathbf y),
\end{equation}
where $\btheta \in \Theta \subseteq \mathbb{R}^n$ is the natural parameter, $p_0$ is a parameter-independent base density with respect to a reference measure $\mu$ (e.g., Lebesgue or counting measure), and
\begin{equation*}
\label{eq:log-partition}
\psi(\btheta)
=
\log\int
\exp\bigl(
\langle\btheta,\mathbf y\rangle
\bigr)
p_0(\mathbf y)\,\mu(d\mathbf y)
\end{equation*}
is the log-partition function. The natural parameter space is
\[
\Theta=\dom\psi
=
\left\{
\btheta\in\mathbb R^n:
\psi(\btheta)<+\infty
\right\}.
\]
The family is regular when $\Theta$ is open,
and minimal when no nonzero linear combination
of the components of $\mathbf y$ is constant
almost surely under the family.
\end{definition}

Throughout the paper, we assume that the exponential
family is regular and minimal. Under these assumptions,
the following standard properties hold
\citep{Banerjee2005,rockafellar1997convex,barndorff2014information}:
\begin{enumerate}[label=(E\arabic*)]
\item
The log-partition function $\psi$ is closed,
proper and convex. Consequently, setting
$f=\psi^*$ gives $f^*=\psi$.

\item
The function $\psi$ belongs to $C^\infty(\Theta)$,
and
\begin{equation*}
\label{eq:log-partition-hessian}
\nabla^2\psi(\btheta)
=
\operatorname{Cov}_{\btheta}(\mathbf y)
\succ0,
\qquad \btheta\in\Theta.
\end{equation*}
In particular, $\psi$ is strictly convex on
$\Theta$.

\item
The function $\psi$ is essentially smooth
and therefore of Legendre type.
Its gradient is a $C^\infty$ diffeomorphism
from $\Theta$ onto $\mathcal X
=\inte(\dom f),$
the interior of the convex support of the
sufficient statistic.
\end{enumerate}

The conjugate generator $f=\psi^*$ is also of Legendre type and belongs to $C^\infty(\mathcal X)$, with $\nabla^2f\succ0$ on $\mathcal X$. The natural and expectation parameters are related by
\begin{equation*}
\label{eq:natural_expectation_mapping}
\mathbf x
=
\nabla\psi(\btheta)
=
\mathbb E_{\btheta}[\mathbf y],
\end{equation*}
and conversely,
\begin{equation*}
\label{eq:natural_expectation_mapping_inverse}
\btheta=\nabla f(\mathbf x).
\end{equation*}
Thus,
\[
\nabla\psi:\Theta\mapsto \mathcal X,
\qquad
\nabla f:\mathcal X\mapsto \Theta
\]
are mutually inverse $C^\infty$ diffeomorphisms.
Moreover, for
$\btheta=\nabla f(\mathbf x)$,
\begin{equation*}
\label{eq:expectation-hessian}
\nabla^2 f(\mathbf x)
=
\left[
\nabla^2\psi(\btheta)
\right]^{-1}
=
\left[
\operatorname{Cov}_{\btheta}(\mathbf y)
\right]^{-1}.
\end{equation*}
Finally, let $\ell(\by;\bx)$ denote the log-density $\log p(\by;\btheta)$ hereafter expressed  in expectation parametrization 
\begin{align*}
\ell(\by;\bx) =\langle\nabla f(\bx),\mathbf y\rangle
-\psi(\nabla f(\bx))
+\log p_0(\mathbf y).
\end{align*}

This set of properties provides the regularity, strict convexity, and Legendre duality required in the sequel. They also establish the correspondence between the natural and expectation parametrizations, which underlies the Bregman representation of exponential-family
distributions recalled next.

\begin{theorem}\cite[Theorem 4]{Banerjee2005}\label{theo:derivation:banerjee:term}
Let $
p(\mathbf{y};\btheta)
$
denote the natural parametrization of the density of the regular exponential family. Under the expectation parametrization, let $\mathcal X \subseteq\mathbb R^n$ denote the mean-parameter space and let $\mathbf x\in\mathcal X$ denote the expectation parameter. Then, the log-density $\ell(\mathbf y;\mathbf x)$ is uniquely expressed as
\begin{equation}
\ell(\mathbf y;\mathbf x)
=
-d_f(\mathbf y,\mathbf x)+\phi_f(\mathbf y),
\label{eq:likelihood}
\end{equation}
where $f: \mathcal{X} \rightarrow \mathbb{R}$ is the Legendre conjugate of the cumulant function, i.e., $f=\psi^*$, $d_f$ denotes the associated Bregman divergence
and the function $\phi_f$ is independent of the parameter $\mathbf x$.
\end{theorem}

This theorem states that the log-density of a regular exponential family distribution can be expressed under the expectation parametrization through a Bregman divergence fully determined by the convex generator $f$. In what follows, this expression will  be referred to as the \emph{primal} Bregman representation of the distribution. For several classical distributions from the  regular exponential families,  Appendix~\ref{app:summary_bregman_exponential} reports the Legendre generator $f$ and the associated divergence $d_f$ for their primal Bregman representations, as well as the minimal sufficient statistic $\mathbf y$  for an observation $\bm{\omega}$ and the expectation parameter $\mathbf x$. 

\section{Dual Bregman representations of regular exponential families}
\label{sec:dual_representations}
\label{sec:problem}

As recalled in \Cref{theo:derivation:banerjee:term}, the log-density of a regular exponential-family
distribution admits the primal Bregman representation \eqref{eq:likelihood}. While this representation provides a natural geometry for the sufficient statistics, it does not directly extend to the expectation-parameter space.
Indeed, Bregman divergences are generally asymmetric: although $d_f(\mathbf y,\mathbf x)$ is convex in its first argument, convexity with respect to the second argument is not guaranteed. Consequently, the primal representation does not, in general, yield an ordinary Bregman divergence with the expectation parameter in the first argument.
 This may cause severe issues when conducting statistical inference of the parameter $\mathbf x \in \mathcal{X}$ within a frequentist or Bayesian framework.  This observation naturally raises the first central question addressed in this paper.
\begin{enumerate}
    \item[(\hypertarget{thequestion}{$\mathcal{Q}$})] Can the \emph{reverse} Bregman divergence defining a regular exponential family distribution be expressed as a potentially scaled \emph{ordinary} Bregman divergence in a suitable coordinate system? 
\end{enumerate}\label{quote:question}
Formally, this question can be answered by identifying a strictly convex function $h$ defined on a dual domain $\mathcal T$ together with a diffeomorphism $\tau:\mathcal{X}  \rightarrow\mathcal T$, such that
\begin{equation}
 d_f\left(\mathbf y,\mathbf x\right) = {g(\mathbf y)}d_h\left(\tau(\mathbf x),\tau(\mathbf y)\right)
\label{eq:problem}
\end{equation}
where $g : \mathcal{X} \to \mathbb{R}_+^{*}$ is a positive scaling function that does not depend on $\mathbf x$. If such a representation exists, then the regular exponential family distribution allows the geometry of the parameter space to be also characterized by an ordinary Bregman divergence, although expressed in a \emph{dual} coordinate system. Yet, to appreciate the scope of the objective pursued in this paper, it is worth emphasizing that the terms \emph{dual} and \emph{duality} used hereafter are not restricted to the notion of Legendre duality. Rather, Legendre duality will be shown to be recovered as a particular instance within the broader class of dual representations characterized by our approach.

\begin{remark}[Connection with conformal divergences]
The scaled divergence appearing in the right-hand side of \eqref{eq:problem} coincides with the $\tau$-conformal divergence $d_{h,g}^\tau(\mathbf{x},\mathbf{y})$ which generalizes the notion of ordinary Bregman divergence \citep{Nock2015}. By denoting $\xi = \nabla h \circ \tau$ and using the Bregman-Legendre duality identity (see Proposition \ref{prop:duality}), this identity can be rewritten as
\begin{align*}
d_f(\mathbf y, \mathbf x) &=
g(\mathbf {y})
d_{h^*}\bigl(\nabla h \circ \tau(\mathbf y),\nabla h\circ \tau(\mathbf x)\bigr)\\
 & = g(\mathbf {y}) d_{h^*}\bigl(\xi(\mathbf y),\xi(\mathbf x)\bigr)\\
 & = d_{h^*,g}^{\xi}(\mathbf y,\mathbf x).
\end{align*}
In other words, looking for the strictly convex function $h$ and the diffeomorphism $\tau$ such that the identity \eqref{eq:problem} holds amounts to identifying a $(\tau,\xi)$-geometric structure underlying conformal counterparts of the primal Bregman representation $d_f(\mathbf y,\mathbf x)$ \citep{amari2000methods}.
\end{remark}

We now turn to the projective characterization of  dual Bregman representations,  which provides
the theoretical foundation for the constructive results developed in the sequel.

\subsection{Projective characterization of dual Bregman representations}\label{subsec:general_formulation}

The following theorem gives necessary and sufficient conditions for a reverse Bregman divergence to admit  a scaled ordinary Bregman divergence in a suitable coordinate system.

\begin{theorem}
\label{thm:proj-characterization}
Let $\mathcal X,\mathcal T\subset\mathbb R^n$ be nonempty open convex sets.
Suppose that $f\in C^2(\mathcal X)$ and $h\in C^2(\mathcal T)$ are
strictly convex, $\tau:\mathcal X\to\mathcal T$ is a $C^1$
diffeomorphism, and $g:\mathcal X\to(0,\infty)$ is a positive function,
not assumed differentiable. Write $\boldsymbol\xi=\nabla h\circ\tau$.
Then the identity
\begin{equation}
\label{eq:proj-identity}
 d_f\left(\mathbf y,\mathbf x\right) = {g(\mathbf y)}d_h\left(\tau(\mathbf x),\tau(\mathbf y)\right)
\end{equation}
holds for all $\mathbf x,\mathbf y\in\mathcal X$
if and only if there exist $\mathbf A\in\mathbb R^{n\times n}$,
$\mathbf b,\mathbf c,\boldsymbol\zeta\in\mathbb R^n$ and
$d,\kappa\in\mathbb R$ such that
\begin{align}
 g(\mathbf x)&=\mathbf c^\top\mathbf x+d>0,\label{eq:proj-g}\\
 \boldsymbol\xi(\mathbf x)&=\frac{\mathbf A\mathbf x+\mathbf b}
 {g(\mathbf x)},\qquad
 \mathbf M=\begin{pmatrix}\mathbf A&\mathbf b\\\mathbf c^\top&d\end{pmatrix}, \quad \det\mathbf M\ne0,
 \nonumber \\
 f(\mathbf x)&=g(\mathbf x)h^*\bigl(\boldsymbol\xi(\mathbf x)\bigr)
 +\boldsymbol\zeta^\top\mathbf x+\kappa.\label{eq:proj-f}
\end{align}
Moreover, whenever these equivalent conditions hold,
the Hessians satisfy
\begin{equation*}\label{eq:proj-hessian}
\nabla^2 f(\mathbf x)
=
g(\mathbf x)\,
J_\tau(\mathbf x)^\top
\nabla^2 h(\tau(\mathbf x))
J_\tau(\mathbf x).
\end{equation*}
\end{theorem}
\begin{proof}
    See Appendix \ref{app:proof:thm:proj-characterization}
\end{proof}

The weighting function $g$ fixes the last row $(\bc^{\top},d)$ of $\bM$. For a given weight $g$, the choice of the remaining rows $(\bA,\bb)$ of $\bM$ will be referred to as the \emph{gauge} in the sequel.

\begin{remark}[Constructive interpretation]
\label{rem:proj-construction}
The proposed characterization can also be used in the reverse direction to construct
pairs of generators. Let $h\in C^2(\mathcal T)$ have positive-definite
Hessian, and choose $\mathbf A,\mathbf b,\mathbf c,d$ such that
$\det\mathbf M\ne0$ and $g(\mathbf x)=\mathbf c^\top\mathbf x+d>0$
on $\mathcal X$. If the projective map
$\boldsymbol\xi(\mathbf x)=(\mathbf A\mathbf x+\mathbf b)/g(\mathbf x)$
satisfies $\boldsymbol\xi(\mathcal X)=\nabla h(\mathcal T)$, then
$\tau=(\nabla h)^{-1}\circ\boldsymbol\xi$ is a diffeomorphism and
\begin{equation*}\label{eq:proj-construct}
 f(\mathbf x)=g(\mathbf x)h^*\bigl(\boldsymbol\xi(\mathbf x)\bigr)
 +\boldsymbol\zeta^\top\mathbf x+\kappa
\end{equation*}
defines a $C^2$ generator with positive-definite Hessian satisfying
\eqref{eq:problem}. Thus the projective characterization is also a
construction principle, subject to the stated range and domain conditions.
\end{remark}

The projective characterization determines the admissible representations up to certain changes of coordinates and affine terms. Before deriving explicit reconstruction formulas, we clarify this gauge freedom and adopt a convenient normalization. The following corollary also establishes the corresponding uniqueness property.

\begin{corollary}[Normalization and uniqueness]
\label{cor:normalization-uniqueness}
Assume that the hypotheses of   \Cref{thm:proj-characterization} hold. Let $(h_0,\tau_0)$ be an admissible representation
associated with the affine weight
\[
g(\mathbf x)=\mathbf c^\top\mathbf x+d>0.
\]
Then the following properties hold.
\begin{itemize}
    \item[i)]  \emph{Normalization}.
There exist $\mathbf a\in\mathbb R^n$ and $\kappa_1\in\mathbb R$ such that \[\widetilde\tau(\mathbf x) = \tau_0(\mathbf x)+\mathbf a, \qquad \widetilde h(\mathbf t) = h_0(\mathbf t-\mathbf a)-\kappa_1 \] define an admissible representation with the same projective matrix $\mathbf M$, the same map $\boldsymbol\xi$, and with $\boldsymbol\zeta=\mathbf 0$ and $\kappa=0$ in \eqref{eq:proj-f}. The associated Bregman divergence is unchanged.
   \item[ii)]  \emph{Uniqueness up to affine equivalence}. If $(h,\tau)$ is another admissible representation
associated with the same weight $g$, then there exist
an invertible matrix
$\mathbf G\in\mathbb R^{n\times n}$,
vectors $\mathbf w,\mathbf v\in\mathbb R^n$,
and a scalar $\kappa_0\in\mathbb R$ such that
\[
\tau(\mathbf x)
=
\mathbf G^{-\top}\tau_0(\mathbf x)+\mathbf v
\]
and
\[
h(\mathbf t)
=
h_0\bigl(\mathbf G^\top(\mathbf t-\mathbf v)\bigr)
+
\langle\mathbf w,\mathbf t-\mathbf v\rangle
+\kappa_0.
\]
Conversely, every pair of this form defines an admissible representation with the same weight $g$. In particular,
\[
d_h\bigl(\tau(\mathbf x),\tau(\mathbf y)\bigr)
=
d_{h_0}\bigl(\tau_0(\mathbf x),\tau_0(\mathbf y)\bigr).
\]
\end{itemize}
\end{corollary}

\begin{proof}
See Appendix~\ref{app:proof:normalization-uniqueness}.
\end{proof}
Exploiting the normalization property stated by \Cref{cor:normalization-uniqueness} \emph{i)}, we may and do
assume throughout the remainder of this section that $\boldsymbol\zeta=\mathbf 0$ and $\kappa=0$ while defining $f$ in \eqref{eq:proj-f}.\\

The projective characterization also provides a way to explicitly recover the dual coordinates and their associated Bregman generator from the projective
coefficients, as established in the following proposition.

\begin{proposition}[Block-matrix reconstruction]
\label{prop:matrix-reconstruction}
Assume that the hypotheses and conclusions of the projective characterization
established by \Cref{thm:proj-characterization} hold. In particular, let 
\begin{equation}\label{eq:projective-data}
 g(\mathbf x)=\mathbf c^\top\mathbf x+d>0,\qquad
 \boldsymbol\xi(\mathbf x)
 =\nabla h(\tau(\mathbf x))
 =\frac{\mathbf A\mathbf x+\mathbf b}{g(\mathbf x)},\qquad
 \mathbf M=\begin{pmatrix}\mathbf A&\mathbf b\\
                    \mathbf c^\top&d\end{pmatrix}
\end{equation}
such that $\det\mathbf M\ne0$.  Then
\begin{equation}
\label{eq:block-reconstruction}
\begin{pmatrix}
 \tau(\mathbf x)\\-h(\tau(\mathbf x))
\end{pmatrix}
=\mathbf M^{-\top}
\begin{pmatrix}
 \nabla f(\mathbf x)\\
 -F(\mathbf x)
\end{pmatrix}
\end{equation}
with $F(\mathbf x)=f^*(\nabla f(\mathbf x))$. Equivalently, when
$\mathbf M^{-\top}=\begin{pmatrix}
\mathbf P&\mathbf q\\\mathbf r^\top&s
\end{pmatrix}$,
then
\begin{equation}\label{eq:component-reconstruction}
\quad
\tau(\mathbf x)=\mathbf P\nabla f(\mathbf x)-\mathbf q F(\mathbf x)
\quad \text{and} \quad
h(\tau(\mathbf x))=-\mathbf r^\top\nabla f(\mathbf x)+sF(\mathbf x).
\quad
\end{equation}
\end{proposition}
\begin{proof}
   See Appendix~\ref{app:proof:prop:matrix-reconstruction}.
\end{proof}

It is worth noting that the reconstruction formulas stated in \Cref{prop:matrix-reconstruction} require the hypotheses of \Cref{thm:proj-characterization} to be satisfied, thereby presupposing the existence of an admissible dual Bregman representation. Going beyond these results, the following theorem establishes both the existence and uniqueness of such representations when $f$ is the expectation-parameter generator of a regular minimal exponential family.

\begin{theorem}[Existence and uniqueness of dual Bregman representations]
\label{thm:dual-construction}
Let $f=\psi^*$ be the expectation-parameter generator
of a regular minimal exponential family, where
$\psi$ denotes its log-partition function, and let
$ \mathcal X=\inte(\dom f)$. Let $\mathbf M$ be an invertible block matrix
of the form introduced in \eqref{eq:projective-data},
and let
\[
g(\mathbf x)=\mathbf c^\top\mathbf x+d>0,
\qquad
\boldsymbol\xi(\mathbf x)
=\frac{\mathbf A\mathbf x+\mathbf b}{g(\mathbf x)},
\quad \mathbf x\in\mathcal X.
\]
Then there exists a unique pair $(h,\tau)$ satisfying the conditions of \Cref{thm:proj-characterization} with this $\bM$, $\boldsymbol\zeta=0$ and $\kappa=0$, where $\tau$ is a $C^\infty$ diffeomorphism from $\X$ onto a nonempty open convex set $\T$, and $h$ is a closed convex function of Legendre type with $\dom h=\T$, $C^\infty$ on $\T$ with $\nabla^2h\succ0$. In particular,
\[
\nabla h\circ\tau=\boldsymbol\xi
\]
and, for $\mathbf x,\mathbf y\in\mathcal X$,
\begin{equation}
\label{eq:construction-conformal}
d_f(\mathbf y,\mathbf x)
=
g(\mathbf y)\,
d_h\bigl(\tau(\mathbf x),\tau(\mathbf y)\bigr).
\end{equation}
\end{theorem}
\begin{proof}
See Appendix~\ref{app:proof:dual-construction}.
\end{proof}

The following corollary provides explicit formulas for the dual representation established in
\Cref{thm:dual-construction}.

\begin{corollary}[Explicit construction]
\label{cor:explicit-construction}
Let $(h,\tau)$ be the pair exhibited in \Cref{thm:dual-construction}. Then, the map $\tau$ and $h\circ \tau$ are  explicitly given by
\eqref{eq:component-reconstruction} with
\begin{equation}
\label{eq:construction-formulas}
h(\mathbf t) =\sup_{\mathbf x\in\mathcal X} \left\{ \left\langle \boldsymbol\xi(\mathbf x),\mathbf t \right\rangle -\frac{f(\mathbf x)}{g(\mathbf x)} \right\},
\end{equation}
Moreover, we have $\inte(\dom h^*)=\boldsymbol\xi(\X)$ on which $h^*$ is given by
\begin{equation}
\label{eq:construction-conjugate}
h^*(\boldsymbol u) 
=
\left(\mathbf q^\top\boldsymbol u+s\right)
f\left(
\frac{\mathbf P^\top\boldsymbol u+\mathbf r}
{\mathbf q^\top\boldsymbol u+s}
\right)
\end{equation}
and
\[
\nabla h^*
=
\tau\circ\boldsymbol\xi^{-1}.
\]
Finally, for every $\mathbf t\in\mathcal T$, the value $\sigma=h(\mathbf t)$ is the unique solution of
\begin{equation}
\label{eq:construction-implicit}
\psi(\mathbf A^\top\mathbf t-\mathbf c\sigma)
=
d\sigma-\mathbf b^\top\mathbf t.
\end{equation}   
\end{corollary}

\begin{proof}
See Appendix~\ref{app:proof:explicit-construction}.
\end{proof}

Therefore, \Cref{thm:dual-construction} provides an affirmative answer to the question (\hyperlink{thequestion}{$\mathcal Q$}). Indeed, it establishes the existence of dual Bregman representations in transformed coordinates for every invertible projective matrix whose associated affine scaling function is positive on the mean-parameter space. Moreover, the result goes beyond the requirements of question (\hyperlink{thequestion}{$\mathcal Q$}), since the constructed Bregman generator $h$ is not merely convex but Legendre. Besides, it appears that the regularity assumptions on the exponential family play an important role in ensuring the global properties of the constructed representation. The following remark clarifies their scope.

\begin{remark}[Role of the assumptions]
\label{rem:role-assumptions}
The proof of \Cref{thm:dual-construction} relies on
the fact that $f$ is closed, proper, of Legendre type,
and $C^2$ with $\nabla^2 f\succ0$ on $\mathcal X$,
together with the openness of $\dom f^*$.
Under these assumptions, the construction remains
valid with $C^1$ regularity for $\tau$ and $C^2$
regularity for $h$, without requiring $f$ to be
$C^\infty$.

The openness of the natural parameter space
$\Theta=\dom f^*$, which is the
regularity assumption for the exponential family,
is essential in the proof of the global properties:
the convexity of $\mathcal T$, the implicit
normalization identity
\eqref{eq:construction-implicit}, and the
Legendre property of $h$.

Without regularity, $\tau$ remains a $C^1$ diffeomorphism onto the open set $\tau(\mathcal X)$, and $h$ is convex on $\mathbb R^n$ with $\nabla^2h\succ0$ on $\tau(\mathcal X)$. The identity \eqref{eq:construction-conformal} still holds, but $\tau(\mathcal X)$ need not be convex when $n\ge2$.

The equality $\tau(\X)=\dom h$ and the Legendre property can indeed fail, even for $n=1$, as shown in Appendix~\ref{app:inverse_gaussian} which discusses the case of the non-regular inverse Gaussian family with known shape parameter \cite[Ch.~8]{barndorff2014information}.
\end{remark}

The following corollary now translates the results established above when deriving the log-density of regular exponential family.

\begin{corollary}[Dual representations of regular exponential family density]
\label{cor:dual-readings}
Let $f=\psi^*$ be the expectation-parameter generator of a regular minimal exponential family, where $\psi$ denotes its log-partition function. Let $\mathcal Y$ denote the support of the sufficient statistic, and let $p_0$ be the base density of the exponential family. Let $(h,\tau)$ be the pair exhibited in \Cref{thm:dual-construction} and associated with
the block matrix
\[
\mathbf M=
\begin{pmatrix}
\mathbf A & \mathbf b\\
\mathbf c^\top & d
\end{pmatrix}.
\]
For every $\mathbf x\in\mathcal X$ and
$\mathbf y\in\mathcal Y$, the log-density admits
the dual exponential representation
\begin{equation}
\label{eq:dual-logdensity-exponential}
\ell(\mathbf y;\mathbf x)=
\left\langle
\tau(\mathbf x),\mathbf A\mathbf y+\mathbf b
\right\rangle
-
g(\mathbf y)h(\tau(\mathbf x))
+
\log p_0(\mathbf y).
\end{equation}
Moreover, Eq. \eqref{eq:construction-implicit} evaluated at $\mathbf t=\tau(\mathbf x)$ coincides with the normalization condition of this dual parametrization. Finally, for $\mathbf y\in\mathcal X$, the log-density also admits the dual Bregman representation
\begin{equation}
\label{eq:dual-logdensity-bregman}
\ell(\mathbf y;\mathbf x)
=
-g(\mathbf y)
d_h\bigl(\tau(\mathbf x),\tau(\mathbf y)\bigr)
+\phi_f(\mathbf y).
\end{equation}
\end{corollary}

\begin{proof}
See Appendix~\ref{app:proof:dual-readings}.
\end{proof}

\begin{remark}[Exponential vs. Bregman representations]
    Unlike the dual exponential representation, the dual Bregman representation requires $\mathbf y\in\mathcal X$. Observations on the boundary of the mean-parameter
space, as may occur for discrete families, are covered by the dual exponential representation.
\end{remark}

\begin{remark}[Graphs of $\psi$ and $h$]\label{rem:graphs} The construction admits a geometric interpretation in terms of graphs. Define $\mathbf D=\mathrm{diag}(\mathbf I,-1)$ and, for $\bx\in\X$, $\btheta=\nabla f(\bx)\in\Theta$ and $\bt=\tau(\bx)\in\T$, so
that $\psi(\btheta)=f^*(\nabla f(\bx))$. Since $\mathbf D^2=\mathbf I$, the identity \eqref{eq:block-reconstruction} in \Cref{prop:matrix-reconstruction} reads
\begin{equation}\label{eq:graph}
  \begin{pmatrix}\bt\\ h(\bt)\end{pmatrix}=\mathbf D\bM^{-\top}\mathbf D\begin{pmatrix}\btheta\\ \psi(\btheta)\end{pmatrix}.
\end{equation}
As $\bx$ ranges over $\X$, $\btheta$ ranges over $\Theta$ \cite[Thm.~26.5]{rockafellar1997convex} and $\bt$ over $\T$. Thus the graph of
$h$ on $\T$ is the image of the graph of $\psi$ on $\Theta$ under the invertible linear map $\mathbf D\bM^{-\top}\mathbf D$.

The log-density pairs these graph points with the statistics $\mathbf D(\by;1)=(\by;-1)$. For $\by\in\mathcal Y$, the log-density \eqref{eq:natural-exponential-family} expressed in natural parameter can be rewritten as
\begin{align*}
  \log p(\by;\btheta) &  =\Bigl\langle\begin{pmatrix}\btheta\\ \psi(\btheta)\end{pmatrix},\, \begin{pmatrix}\by\\-1\end{pmatrix}\Bigr\rangle +\log p_0(\by).
\end{align*}
From \eqref{eq:dual-logdensity-exponential} with $\bt=\tau(\bx)$, this log-density also admits the dual exponential representation 
\begin{align*}\label{eq:pairing}
  \log p(\by;\btheta) 
  &=\Bigl\langle\begin{pmatrix}\bt\\ h(\bt)\end{pmatrix},\,\mathbf D\bM \mathbf D\begin{pmatrix}\by\\-1\end{pmatrix}\Bigr\rangle +\log p_0(\by)
\end{align*}
Graph points and statistics are thus transformed contragrediently, by $\mathbf D\bM^{-\top}\mathbf D$ and $\mathbf D\bM\mathbf D$, which preserves the log-likelihood.
\end{remark}

The preceding results establish a general framework for constructing dual Bregman representations of
regular exponential families. We now consider two particularly simple choices of the projective
coefficients, leading to the so-called $\beta$-scaled and $\alpha$-scaled representations. These constructions
will also provide the reference geometries used later to study conjugate priors and relative smoothness.

\begin{corollary}[$\beta$- and $\alpha$-scaled dual models]
\label{cor:scaled-representations}
\label{cor:beta_existence}
\label{cor:alpha_existence}
Let $f=\psi^*$ be the expectation-parameter generator
of a regular minimal exponential family, where
$\psi$ denotes its log-partition function, and let
$\mathcal X=\inte(\dom f)$.
Let $\alpha,\beta>0$. The following are two particular instances of the dual Bregman representation
\eqref{eq:construction-conformal}.
\begin{itemize}
    \item[i)] \emph{$\beta$-scaled dual model.} In arbitrary dimension, the constant weight
\[
g(\mathbf y)=\frac1\beta
\]
gives 
\[
\tau=\nabla f,
\qquad
\mathcal T
=\inte(\dom f^*),
\qquad
h=\beta f^*.
\]
Consequently, for $\mathbf x,\mathbf y\in\mathcal X$,
\begin{equation*}
\label{eq:logdensity_beta}
\ell(\mathbf y;\mathbf x) = -\frac1\beta d_{h}\bigl( \tau(\mathbf x),\tau(\mathbf y) \bigr) +\phi_f(\mathbf y).
\end{equation*}
\item[ii)] \emph{$\alpha$-scaled dual model.} Suppose that $n=1$ and
$\mathcal X\subseteq(0,\infty)$. 
The weight
\[
g(y)=\frac y\alpha
\]
gives
\[
\tau(x)
=
xf'(x)-f(x)
=
f^*(f'(x)),
\qquad
\mathcal T
=
f^*\bigl(\inte(\dom f^*)\bigr),
\]
and
\[
h(t)
=
-\alpha
\left[
f^*
\right]^{-1}(t),
\qquad t\in\mathcal T.
\]
Consequently, for $x,y\in\mathcal X$,
\begin{equation*}
\label{eq:logdensity_alpha}
\ell(y;x)  = -\frac y\alpha d_{h}\bigl( \tau(x),\tau(y) \bigr) +\phi_f(y).
\end{equation*}
\end{itemize}
In the first case, $\tau$ is strictly monotone, whereas in the second case, $\tau$ is strictly increasing. Both dual generators are of Legendre type under the assumptions of \Cref{thm:dual-construction}.
\end{corollary}

\begin{proof}
See Appendix~\ref{app:proof:scaled-representations}.
\end{proof}

The two scaled models illustrate how the projective construction yields different, yet equivalent, dual Bregman representations. This corollary also shows that this construction encompasses the classical Legendre duality as a particular instance of the $\beta$-scaled dual representation with $\beta=1$. We next investigate the implications of these representations while expressing conjugate prior distributions within a Bayesian framework.

\subsection{Dual representations of conjugate priors}\label{subsec:application_priors}

When conducting parameter inference in a statistical model induced by a distribution belonging to the exponential family, \Cref{theo:derivation:banerjee:term} shows that maximizing the likelihood function is equivalent to minimizing the Bregman divergence generated by $f$. Interestingly, this primal Bregman representation also proves particularly useful to derive conjugate prior distributions widely adopted in Bayesian inference. Indeed, suppose now that the prior information on the parameter $\mathbf x$ is governed by  a positive precision parameter $\beta>0$ together with a location parameter $\alpha/\beta\in\mathcal X$. For a given likelihood function, a  prior distribution $p(\mathbf x|\alpha,\beta)$ is said to be conjugate when it preserves the analytical form of the posterior distribution after observing\footnote{Without loss of generality, we consider a single $\mathbf y$.} $\mathbf  y$. It appears that this conjugacy systematically emerges when resorting to a primal Bregman representation of the prior distribution, i.e., by considering a log-prior on expectation coordinates of the form  \citep{Agarwal2010} 
\begin{equation}
\log p(\mathbf x|\alpha,\beta)
=
-\beta\, d_f\left(\frac{\alpha}{\beta},\mathbf x\right)+\varphi_f(\alpha, \beta)
\label{eq:prior}
\end{equation}
where $\varphi_f$ denotes a normalization potential  independent of $\mathbf{x}$. Throughout this section, prior and posterior densities in expectation coordinates are understood with respect to Lebesgue measure. The hyperparameters are assumed to be such that the prior density is normalizable. The conjugate nature of such prior models is confirmed by the following proposition.

\begin{proposition}\label{prop:weighted:bregman:identity}
   Consider the Bayesian model defined by the log-likelihood
\begin{equation*}
\ell(\mathbf y;\mathbf x) = -d_f(\mathbf y,\mathbf x)+\phi_f(\mathbf y)
\end{equation*}
and the Bregman representation of its conjugate prior
\begin{equation*}
\log p(\mathbf x|\alpha,\beta) = -\beta\, d_f\left(\frac \alpha \beta,\mathbf x\right)+\varphi_f(\alpha,\beta).
\end{equation*}
Then the log-posterior can be written as
\begin{equation}
\log p(\mathbf x|\mathbf y,\alpha,\beta) = - (1+\beta)\, d_f\left(\frac{\mathbf y+\alpha}{1+\beta},\mathbf x\right)+\Psi_f(\mathbf y,\alpha,\beta)
\label{eq:weighted}
\end{equation}
where
$\Psi_f(\mathbf y,\alpha,\beta)$ is independent of $\mathbf x$.
\end{proposition}
\begin{proof}
   See Appendix \ref{app:weighted:bregman:identity}.   
\end{proof}

This property  shows that the posterior \eqref{eq:weighted} retains the same parametric form \eqref{eq:prior} as the prior with updated hyperparameters
\begin{equation*}
\alpha^{+} = \mathbf y+\alpha, \qquad \beta^{+} = 1+\beta,
\label{eq:update}
\end{equation*}
which is the defining property of conjugacy in the present parametrization. 

The dual Bregman representations of regular exponential family distributions identified in \Cref{subsec:general_formulation} are particularly useful for deriving dual representations of these conjugate priors, as alternatives to the primal Bregman representation \eqref{eq:prior}. Indeed, after explicitly introducing the location parameter 
\begin{equation}
\mathbf z = \frac \alpha \beta
\end{equation}
the primal Bregman representation \eqref{eq:prior} exactly  matches the forms inducing $\alpha$- and $\beta$-scaled dual models identified in \Cref{subsec:general_formulation}, leading to the two following corollaries.

\begin{corollary}
\label{cor:prior_beta}
Let $f=\psi^*$ be the expectation-parameter generator
of a regular minimal exponential family, where
$\psi$ denotes its log-partition function, and let
$\mathcal X=\inte(\dom f),\,\mathcal T=\inte(\dom f^*)$. For any 
$(\mathbf x,\mathbf z) \in\mathcal X^2$ and $\beta>0$,  consider the Bregman representation of the log-likelihood associated with a regular exponential family distribution
\begin{equation*}
\ell(\mathbf y;\mathbf x) = -d_f(\mathbf y,\mathbf x) +\phi_f(\mathbf y)
\end{equation*}
and the Bregman representation of its conjugate prior
\begin{equation}\label{eq:prior_beta_proof}
\log p(\mathbf x|\mathbf z, \beta) = -\beta d_f(\mathbf z,\mathbf x) +\varphi_f(\beta\,\mathbf z,\beta).
\end{equation}
Then, this conjugate prior admits the equivalent dual Bregman representation
\begin{equation}
\label{eq:priorrev}
\log p(\mathbf x|\mathbf z, \beta)
=
-d_h\bigl(\tau(\mathbf x),\tau(\mathbf z)\bigr)
+\varphi_f(\beta\,\mathbf z,\beta)
\end{equation}
with
\begin{equation*}
\tau =\nabla f :\mathcal X \to \mathcal T \quad \text{and} \quad h=\beta f^*.
\end{equation*}
Moreover the resulting log-posterior can be written as
\begin{equation}
\log p(\mathbf x|\mathbf y,\mathbf z,\beta) = - \frac{(1+\beta)}{\beta}\, d_h\left(\tau(\mathbf x),\tau\left(\frac{\mathbf y+\beta \mathbf z}{1+\beta}\right)\right)+\Psi_h(\mathbf y,\mathbf z,\beta)
\label{eq:weighted_dual}
\end{equation}
where
$\Psi_h(\mathbf y,\mathbf z,\beta)$ is independent of $\mathbf x$.
\end{corollary}
\begin{proof}
   The existence of the dual representation \eqref{eq:priorrev} of the prior follows directly from applying \Cref{cor:beta_existence} to the conjugate prior \eqref{eq:prior_beta_proof}. Similarly, the expression of the log-posterior distribution \eqref{eq:weighted_dual} follows from applying \Cref{cor:beta_existence} to the log-posterior \eqref{eq:weighted} derived in Proposition \ref{prop:weighted:bregman:identity}. 
\end{proof}

It is worth noting that \Cref{cor:prior_beta} does not introduce a new family of conjugate priors associated with likelihoods derived from the regular exponential family. Rather, it provides a systematic way to derive Bregman representations of those priors, as an alternative of the representation already identified by \cite{Agarwal2010}. Equation~\eqref{eq:weighted_dual} shows that the posterior retains the same parametric form as the dual formulation \eqref{eq:priorrev} of the prior with updated hyperparameters $\tau(\mathbf z^+)$ and $\beta^{+}/\beta$ where $\mathbf z^+=\alpha^{+}/\beta^+$. Hence, \Cref{cor:prior_beta} shows that the conjugacy property also emerges from the dual representations of the likelihood and the prior. Similar properties can also be exhibited when considering a conjugate prior of the form of an $\alpha$-scaled model.

\begin{corollary}
\label{prop:alternative}
Let $f=\psi^*$ be the expectation-parameter generator
of a regular minimal exponential family, where
$\psi$ denotes its log-partition function, and let
$\mathcal X=\inte(\dom f)\subseteq\mathbb R_{+}^*,\,\mathcal T=f^*(\inte(\dom f^*))$.
For any  $( x, z) \in\mathcal X^2$ and $\alpha>0$, consider the Bregman representation of the log-likelihood associated with a regular exponential family distribution
\begin{equation*}
\ell( y; x) = -d_f( y, x) +\phi_f( y)
\end{equation*}
and the Bregman representation of its conjugate prior
\begin{equation}
\label{eq:prior_alpha_proof}
\log p(x| z,\alpha)
=
-\frac{\alpha}{z}d_f(z,x)
+\varphi_f(\alpha,\frac \alpha z).
\end{equation}
Then, this conjugate prior admits the equivalent dual Bregman representation
\[
\log p(x| z,\alpha)
=
-d_{{h}}
\left(
{\tau}(x),
{\tau}(z)
\right)
+\varphi_f(\alpha,\frac \alpha z)
\]
where
\begin{align*}
    {\tau}\ :\  &\mathcal X\to{\mathcal T}       & \text{and} &\  & {h}\ :\  &{\mathcal T} \to \mathbb{R} \cup \left\{+\infty\right\}\\
                        &x \mapsto f^*(f'(x))                          &            &       &                   & t \mapsto -\alpha [f^*]^{-1}\!\left(t\right).         
\end{align*}
Then ${\tau}$ is a strictly increasing diffeomorphism and ${h}$ is twice continuously differentiable and strictly convex on ${\mathcal T}$. Moreover, the log-posterior can be written as
\begin{equation}
\log p( x| y, z,\alpha) = - \frac{y+\alpha}{\alpha}\, d_{{h}}\left({\tau}( x),{\tau}\left(\frac{ y+\alpha}{1+\frac \alpha z}\right)\right)+{\Psi}_{{h}}( y,z,\alpha)
\label{eq:weighted_dual_alpha}
\end{equation}
where
${\Psi}_{{h}}(y, z,\alpha)$ is independent of $x$.
\end{corollary}
\begin{proof}
The existence of the dual representation of the prior easily follows from the application of \Cref{cor:alpha_existence} to the conjugate prior \eqref{eq:prior_alpha_proof}. The expression of the log-posterior distribution \eqref{eq:weighted_dual_alpha} results from the application of \Cref{cor:alpha_existence} applied to the log-posterior \eqref{eq:weighted} derived in Proposition \ref{prop:weighted:bregman:identity}. 
\end{proof}

\Cref{prop:alternative} shows that the scaling factor $\alpha/z$ involved in the primal Bregman representation of the $\alpha$-scaled prior model can be absorbed into the coordinate mapping ${\tau}$ and the convex generator ${h}$.

As expected, these results confirm that prior conjugacy is systematically ensured by first identifying the Bregman generator $f$ associated with the likelihood and then adopting the $\beta$- or $\alpha$- scaled dual Bregman representations induced by $(h,\tau)$. Contrary to the primal Bregman representation of conjugate priors introduced by  \cite{Agarwal2010}, the proposed representations rely on ordinary forms of Bregman divergences in dual coordinates, characterizing the geometry of the parameter space. Besides, this geometric characterization is independent of the particular probability distribution and relies only on the convex geometry induced by the likelihood generator. This suggests that conjugate Bayesian models may be organized according to their underlying Bregman geometry, rather than solely according to the form of their probability distributions.

\subsection{Dual representations of classical exponential family distributions}\label{subsec:examples}

The general construction established in \Cref{thm:dual-construction} provides a systematic way of deriving dual Bregman representations for regular minimal exponential families. The $\beta$- and $\alpha$-scaled constructions of \Cref{cor:scaled-representations} are particular instances of this projective framework. Table~\ref{tab:geometry} illustrates these two constructions for several classical exponential family models, with $\alpha=\beta=1$ for simplicity. The corresponding calculations are detailed in Appendix~\ref{appendix:additional-examples}. The inverse-Gaussian family is also included as a non-regular example, illustrating the limitations of the global Legendre conclusions when the regularity assumption is not met.

\begin{table}
\centering
\renewcommand{\arraystretch}{2.2}
\resizebox{\textwidth}{!}{%
\begin{tabular}{l|c|ccc|ccc}
\toprule
\multirow{2}{*}{\textbf{Distribution}}   & \multirow{2}{*}{$\mathcal{X}$} & \multicolumn{3}{c|}{\textbf{$\beta$-scaled dual model}} & \multicolumn{3}{c}{\textbf{$\alpha$-scaled dual model}} \\
                              &     &  $\tau({x})$      &  $h(t)$    &  ${\mathcal{T}}$  &  ${\tau}({x})$      & ${h}(t)$ & ${\mathcal{T}}$\\ 
\hline
Gaussian $\mathcal{N}(\mu,\textcolor{gray}{\sigma^2})$    & $\mathbb{R}$      & $\displaystyle \frac{x}{\sigma^2}$    & $\displaystyle \frac{\sigma^2 t^2}{2}$ & $\mathbb{R}$    & \multicolumn{3}{c}{ N/A}\\
Poisson  $\mathcal{P}(\lambda)$                           & $\mathbb{R}_+^*$  &  $\log x$               & $e^t$                 & $\mathbb{R}$       & $x$            & $-\log t $             & $\mathbb{R}_+^*$\\
Exponential $\mathcal{E}(b)$                  & $\displaystyle \mathbb{R}_+^*$  &  $\displaystyle -\frac{1}{x}$         &  $-1-\log(-t)$        &  $\mathbb{R}_-$    & $\log x -1$    & $ e^{-t-1}$             & $\mathbb{R}$\\
Gamma $\mathcal{G}(\textcolor{gray}{k},b)$       & $\mathbb{R}_+^*$  &   $\displaystyle-\frac{k}{x}$        &  $-k \log(-t)$        &  $\mathbb{R}_-$    & $\displaystyle k\log x - k$  & $\displaystyle k e^{-(t+k)/{k}}$ & $\mathbb{R}$ \\
Bernoulli  $\mathcal{B}er(p)$  & $(0,1)$           &   $\displaystyle\log \frac{x}{1-x}$  & $ \log(1+ e^t)$       &  $\mathbb{R}$      & $-\log(1-x)$   & $\displaystyle -\log(e^t -1)$        & $\mathbb{R}_+^*$\\
Binomial   $\mathcal{B}(\textcolor{gray}{k},p)$  & $(0,k)$           &   $\displaystyle \log{\frac{x}{k-x}}$ &  $\displaystyle k \log(1+e^t)$      &  $\mathbb{R}$      & $\displaystyle -k\log (k-x)$ & $ - \log(ke^{t/k}-1)$     
            & $ (-k \log k, +\infty)$ \\
Negative Binomial $\mathcal{NB}(\textcolor{gray}{r},p)$   & $\mathbb{R}_+^*$  & $\displaystyle \log \frac{x}{r+x}$    & $\displaystyle -r \log(1-e^t)$      &  $\mathbb{R}_-$    & $\displaystyle r \log(r+x)$  &  $\displaystyle \frac{t}{r} - \log(e^{t/r} -r)$  
            & $ (r \log r, +\infty)$ \\
Inverse-Gaussian $\mathcal{IN}(\mu,\textcolor{gray}{\lambda})$  
       & $\mathbb{R}_+^*$  &  $\displaystyle -\frac{\lambda}{2x^2}$& $-\sqrt{-2\lambda t }$& $\mathbb{R}_-$     & $\displaystyle -\frac{\lambda}{x}$ & $\displaystyle \frac{t^2}{2\lambda}$ & $\mathbb{R}_-^*$ \\
Gaussian $\mathcal{N}(\mu,{\sigma^2})$& $\{ x_1 \in \mathbb{R},\ x_2>x_1^2\}$    &   $\displaystyle \frac{1}{x_2-x_1^2} \left(\begin{smallmatrix} x_1 \\ -\frac{1}{2} \end{smallmatrix}\right)    
$      &    see \eqref{eq:gaussian_h} 
            &    $\mathbb{R} \times \mathbb{R}^*_-$     & \multicolumn{3}{c}{ N/A}\\
Gamma $\mathcal{G}(k,b)$& $\{ x_2 \in \mathbb{R}_+^*, \ x_1< \log x_2\}$      
                     &        no closed-form              &   see \eqref{eq:gamma_h}  
                     & $\left\{ t_1> -1, \ t_2 <0 \right\}$         & \multicolumn{3}{c}{ N/A}  \\
Beta $\mathcal{B}e(a,b)$   & $\displaystyle \operatorname{ran}\nabla f^*$ &  no closed-form  &  see \eqref{eq:beta_h} 
                            & $\left\{t_1>-1, t_2 > -1\right\}$ & \multicolumn{3}{c}{ N/A} \\
\bottomrule
\end{tabular}
}
\caption{Dual Bregman representations of classical exponential family models. Fixed parameters appear in gray. Whenever well-identified, the divergences resulting from the $\alpha$-scaled or $\beta$-scaled generators ${h}$ are specified in Appendix~\ref{appendix:additional-examples}.}
\label{tab:geometry}
\end{table}

The examples reported in this table  reveal several structural properties of the proposed geometric framework.  First, they highlight that several apparently different exponential-family models exhibit closely related dual Bregman geometries. This occurs whenever the primal Bregman representations of their densities are induced by the same convex generator $f$, or by  generators differing only by a positive scaling factor.  For instance, the Exponential and Gamma distributions with known shape parameter both induce (scaled versions of) the same Itakura-Saito geometry, while the Bernoulli and Binomial distributions exhibit closely related entropic geometries. 

The $\alpha$-scaled construction of \Cref{cor:alpha_existence} provides explicit dual Bregman representations for one-dimensional regular minimal exponential families with expectation-parameter domain $\mathcal X\subseteq(0,\infty)$.
For the corresponding models in Table~\ref{tab:geometry}, the dual generator $h$
is of Legendre type, consistently with \Cref{thm:dual-construction}.

The inverse-Gaussian family provides a contrasting example. Although this family is non-regular, it
admits an $\alpha$-scaled dual Bregman representation. However, its generator is not essentially smooth on the transformed domain $\tau(\mathcal X)=(-\infty,0)$.
This illustrates the distinction between the existence of a dual representation and the stronger
Legendre properties guaranteed by regularity.

These examples also illustrate how different scalings give rise to distinct dual Bregman representations of the same underlying likelihood divergence.
In particular, the $\alpha$-scaled construction of \Cref{cor:alpha_existence} expresses a conformally weighted Bregman divergence as an
ordinary Bregman divergence in transformed coordinates, through a suitable dual mapping
$\tau$ and generator $h$.
For the Poisson distribution, this construction yields an Itakura--Saito geometry, whereas the $\beta$-scaled construction involves the
exponential generator in natural coordinates. These representations are related by the
coordinate transformations and conformal weights established in the general projective framework. This illustrates how the geometric representation of conjugacy depends on the chosen scaling, without changing the underlying statistical model.

Finally, the Gaussian, Gamma, and Beta examples with multiple unknown parameters demonstrate that the general projective framework is not restricted to one-dimensional exponential families. In particular, the $\beta$-scaled construction extends directly to multidimensional models, even when the corresponding Legendre mappings cannot be expressed in elementary closed form. For the Gamma distribution with unknown shape
and rate parameters and the Beta distribution with two unknown shape parameters, the relevant dual coordinate mappings generally require numerical evaluation. This computational requirement does not affect the validity of the dual Bregman representation or the underlying Legendre properties.

\begin{remark}(Application to conjugate priors)
    Adopting the dual Bregman representations proposed in \Cref{subsec:general_formulation} and specifically instantiated to derive conjugate priors in \Cref{subsec:application_priors}  recovers well-established conjugate models, as emphasized in Appendix~\ref{appendix:additional-examples} whenever appropriate.
\end{remark}

These examples demonstrate that the Bregman geometry associated with regular exponential families is not merely a model-specific representation, but reflects a common convex-analytic structure across these families. This raises another natural question: beyond providing an explicit ordinary Bregman formulation of regular exponential family log-density, do these $\beta$- and $\alpha$-scaled dual Bregman representations also induce a suitable geometry for analyzing the regularity of this log-density? We address this question in the next section through the prism of relative smoothness.


\section{Relative smoothness of regular exponential families in dual Bregman geometries}
\label{sec:relative-smoothness-conjugacy}

This section investigates whether the dual Bregman geometry identified in \Cref{sec:dual_representations} also provides a natural reference geometry for relative smoothness. More precisely, we characterize the relative smoothness of the negative log-likelihood associated with the $\beta$- and $\alpha$-scaled models exhibited in \Cref{subsec:general_formulation}. This establishes a connection between the geometry induced by the dual Bregman representations of regular exponential family distributions and their regularity properties required by Bregman-based optimization methods.

\subsection{Geometry induced by \texorpdfstring{$\beta$}{beta}-scaled dual Bregman representations}
\label{subsec:smoothness_geometry_beta}
We first investigate the geometry induced by the $\beta$-scaled $(h,\tau)$-dual representations identified in \Cref{cor:beta_existence}. Recall that, under such a parametrization, the dual coordinates are defined by $\mathbf t=\tau(\mathbf x)=\nabla f(\mathbf x)$ and the dual Bregman generator is  $h=\beta f^*$. Interestingly, under such a geometry, the negative log-likelihood associated with a regular exponential family distribution satisfies the Lipschitz-like/convexity condition enounced in \Cref{def:LC}.

\begin{proposition}
\label{prop:beta-relative-smoothness} 
Consider the Bregman representation $\ell(\mathbf y;\mathbf x) = -d_f(\mathbf y,\mathbf x) +\phi_f(\mathbf y)$ of the log-likelihood associated with a regular exponential family distribution. Let $\lambda(\mathbf y;\cdot)$ denote the negative log-likelihood expressed in dual coordinates induced by a $\beta$-scaled dual Bregman representation, i.e.,
\begin{equation}\label{eq:lambda_beta}
\lambda(\mathbf y;\mathbf{t}) = -\ell(\mathbf y;\tau^{-1}({\mathbf{t})}) \quad \text{with} \quad \mathbf t=\tau(\mathbf x)=\nabla f(\mathbf x).
\end{equation}
For $\beta>0$, define $h(\mathbf{t})=\beta f^*(\mathbf{t})$. Then the function $\lambda(\mathbf y;\cdot)$ is $L$-smooth relative to $h$  for every  $L\geq\frac{1}{\beta}$.
\end{proposition}
\begin{proof} 
 See Appendix \ref{app:proof_beta-relative-smoothness} 
\end{proof}

\begin{remark} 
Since $f^*$ is strictly convex on $\mathcal T$, the threshold
$L=1/\beta$ is sharp: for $L>1/\beta$, the function
$Lh-\lambda$ is strictly convex, whereas for $L=1/\beta$ it is affine. For $L<1/\beta$, the function $Lh-\lambda$ is strictly concave.
\end{remark}

This result shows that, for a given negative log-likelihood, a reference function $h$ leading to a suitable reference geometry can be systematically exhibited by resorting to the $\beta$-scaled dual Bregman representation exhibited in \Cref{cor:beta_existence}. Then, as a specific instance of \Cref{eq:bregman-descent}, it yields the Bregman descent lemma in dual  coordinates  
\begin{equation*} 
\lambda(\mathbf y;\mathbf t^{\prime}) \leq \lambda({\mathbf y};\mathbf t) + \left\langle \nabla\lambda(\mathbf y;{\mathbf t}), \mathbf t^{\prime}-{\mathbf t} \right\rangle + L\,d_h(\mathbf t^{\prime},{\mathbf t})
\end{equation*} 
for all $\mathbf t,{\mathbf t}^{\prime}\in\mathcal T$ and every
$L\geq1/\beta$. Moreover, whenever the natural and expectation parameters coincide, such as for Gaussian location model with unit variance, this descent lemma also holds in expectation coordinates. \Cref{prop:beta-relative-smoothness} also admits a component-wise formulation, in the case of a Bregman representation of the log-likelihood induced by a separable generator, as stated in the following corollary.

\begin{corollary} 
\label{prop:multidimensional-separable_beta}
Consider the Bregman representation $\ell( \mathbf y; \mathbf x) = -d_f( \mathbf y, \mathbf  x) +\phi_f( \mathbf y)$ of the log-likelihood associated with a regular exponential family distribution. Assume the underlying generator $f$ is separable, i.e., 
\begin{equation*}
f(\mathbf x) = \sum_{i=1}^{n} f_i(x_i).
\end{equation*}
Let ${\lambda}(\mathbf y;{\mathbf{t}})$ denote the negative log-likelihood expressed in dual coordinates  ${\mathbf{t}}={\boldsymbol{\tau}}( \mathbf{x})$ induced by a component-wise $\beta$-scaled dual Bregman representation, i.e.,  with
\begin{align*}
{\mathbf{t}} &= ({t}_1,\ldots,{t}_{n}) \\
{\boldsymbol\tau}(\mathbf x) &= \left(\tau_1(x_1),\ldots,\tau_{n}(x_{n}) \right)\\
\tau_i(x_i) &= f_i'(x_i).
\end{align*}
For $\boldsymbol{\beta} = (\beta_1,\ldots,\beta_n)^\top$ and $\beta_i>0$, let define $ h_{\mathrm{sep}}({\mathbf t})
= \sum_{i=1}^{n}  \beta_i f^*_i\!\left( t_i\right)$. Then the function $\lambda(\mathbf{y};\cdot)$ is $L$-smooth relative to $ h_{\mathrm{sep}}$ for every $L\geq\|{\boldsymbol{\beta}}^{-1}\|_{\infty}$ where the division in ${\boldsymbol{\beta}}^{-1}$ should be understood componentwise.
 \end{corollary}
\begin{proof}
The proof easily follows from the convexity of nonnegative linear combination of convex functions, combined with the proof of \Cref{prop:beta-relative-smoothness}.
\end{proof}

\begin{remark} 
Since the negative log-likelihood associated with a regular exponential family distribution and its primal Bregman divergence only differs by the constant $-\phi_{f}(\mathbf{y})$, this property also provides a reference geometry for the Bregman divergence itself, yet expressed in the dual coordinates. In other words, the function $\mathbf{t} \mapsto d_f(\mathbf{y};\tau^{-1}(\mathbf{t}))$ is $L$-smooth relative to $h$ for the same range of $L$ values.
\end{remark}

\subsection{Geometry induced by \texorpdfstring{$\alpha$}{alpha}-scaled dual Bregman representations}
\label{subsec:smoothness_alpha}
We now consider the geometry induced by the $\alpha$-scaled $({\tau},{h})$-dual Bregman representations  introduced in  \Cref{cor:alpha_existence}. Under such a parametrization, contrary to the previous one, the dual coordinates ${t}$ do not, in general, coincide with the natural parameter of the regular exponential family model. Instead, they write as ${t}={\tau}( x)= f^*(f'(x))$ while the dual generator is ${h}(t)= -\alpha (f^*)^{-1}({t})$. Again, adopting such a geometry ensures relative smoothness of the log-likelihood expressed in dual coordinates, as stated above.

\begin{proposition} 
\label{prop:general-tau-tilde} 
Consider a regular exponential family with one-dimensional expectation-parameter domain
$\mathcal X\subset(0,\infty)$, and let
\[ \ell(y;x)=-d_f(y,x)+\phi_f(y) \]
be its Bregman log-likelihood representation.

Let ${\lambda}(y;\cdot)$ denote the negative log-likelihood expressed in dual coordinates induced by an $\alpha$-scaled dual Bregman representation, i.e.,
\begin{equation}\label{eq:lambda_alpha}
{\lambda}( y;{t}) = -\ell( y;{\tau}^{-1}({{t})}) \quad \text{with} \quad {t}={\tau}( x)= f^*(f'(x)).
\end{equation}
For  $\alpha>0$, define ${h}({t}) = -\alpha f'\!\left({\tau}^{-1}({t})\right)$. If the function ${h}$ is a Legendre function, then
\begin{itemize}
    \item if $y\neq 0$, the function ${\lambda}(y;\cdot)$ is $L$-smooth relative to ${h}$ for every $L\geq\frac{y}{\alpha}$.
    \item if $y=0$, for any $\epsilon>0$, the function ${\lambda}(y;\cdot)$ is $L$-smooth relative to ${h}$ for every $L\geq\epsilon$.
\end{itemize}
 \end{proposition}
\begin{proof}
See Appendix \ref{app:proof_general-tau-tilde}
\end{proof}

\begin{remark}
Since ${h}$ is strictly convex on
$\inte(\dom{h})$, the threshold
$L=y/\alpha$ is sharp: for $L>y/\alpha$, the function
$L{h}-{\lambda}$ is strictly convex, whereas for
$L=y/\alpha$ it is affine. For $L<y/\alpha$, it is strictly concave.
\end{remark}

The corresponding Bregman descent lemma in the transformed coordinates reads
\begin{equation} 
\begin{aligned} 
{\lambda}(y;{t}^{\prime}) \leq{}& {\lambda}(y;{{t}}) + ({t}^{\prime}-{t}) {\lambda}'(y;{t}) + L\,d_{{h}} ({t}^{\prime},{t}),
\end{aligned}
\end{equation} 
for all ${t},{t}^{\prime} \in\inte(\dom{h})$ and every $L\geq y/\alpha$. Moreover, whenever ${\tau}=\mathrm{Id}$, the primal and dual coordinates coincide and this descent lemma also holds in expectation coordinates. An archetypal example is the Poisson model, which will be deeply investigated in \Cref{sec:application}. This particular example also reveals important features when \Cref{prop:general-tau-tilde} extends to a separable multidimensional model, as stated in the following corollary (see Remark \ref{remark:poisson:bauschke}).

\begin{corollary}
\label{prop:multidimensional-separable}
Consider the Bregman representation $\ell( \mathbf y; \mathbf x) = -d_f( \mathbf y, \mathbf  x) +\phi_f( \mathbf y)$ of the log-likelihood associated with a regular exponential family distribution. Assume the underlying generator $f$ is separable, i.e., 
\begin{equation}
f(\mathbf x)
=
\sum_{i=1}^{n} f_i(x_i).
\end{equation}
Assume moreover that each component has an
expectation-parameter domain $\mathcal X_i\subset(0,\infty)$.
Let ${\lambda}(\mathbf y;{\mathbf{t}})$ denote the negative log-likelihood expressed in dual coordinates  ${\mathbf{t}}={\boldsymbol{\tau}}( \mathbf{x})$ induced by a component-wise $\alpha$-scaled dual Bregman representation, i.e., 
with
\begin{align}
{\mathbf{t}} &= ({t}_1,\ldots,{t}_{n}) \\
{\boldsymbol\tau}(\mathbf x) &= \left({\tau}_1(x_1),\ldots,{\tau}_{n}(x_{n}) \right)\\
{\tau}_i(x_i) &= f_i^*(f_i'(x_i)).
\end{align}
For $\boldsymbol{\alpha} = (\alpha_1,\ldots,\alpha_n)^\top$ and $\alpha_i>0$, define ${h}_{\mathrm{sep}}({\mathbf{t}})
= \sum_{i=1}^{n}{h}_i({t}_i)$ with ${h}_i({t}_i) = -\alpha_i f_i'\!\left({\tau}_i^{-1}({t}_i)\right)$. If all functions ${h}_i$ are Legendre functions then
\begin{itemize}
    \item if $\|\mathbf{y}\|_{\infty} \neq 0$, the function ${\lambda}(\mathbf{y};\cdot)$ is $L$-smooth relative to ${h}_{\mathrm{sep}}$ for every $L\geq\|{\mathbf{y}}/{\boldsymbol{\alpha}}\|_{\infty}$ where the division by $\boldsymbol{\alpha}$ should be understood componentwise,
    \item if $\|\mathbf{y}\|_{\infty} =0$, for any $\epsilon>0$, the function  ${\lambda}(\mathbf{y};\cdot)$ is $L$-smooth relative to ${h}_{\mathrm{sep}}$ for every $L\geq \epsilon$,
\end{itemize}
Moreover, the threshold $ L_{\min}=\|\mathbf y/\boldsymbol\alpha\|_\infty$ is sharp. 
 \end{corollary}

\begin{proof}
See Appendix \ref{app:proof_multidimensional-separable}
\end{proof}

\begin{remark}
\label{remark:poisson:bauschke}    
For Poisson linear inverse problems, \citep{Bauschke2017} established the relative-smoothness condition using the Burg entropy as reference function, with a sufficient
constant $ L\geq \|\mathbf y\|_1$. For the separable model considered above with $\alpha_i=\alpha=1$, the componentwise structure instead leads to the bound $ L\geq{\|\mathbf y\|_\infty}. $
These bounds arise from different model structures and reference geometries: the result enounced by \cite{Bauschke2017} is associated to a Poisson inverse problem which involves a linear operator coupling the components whereas the result above exploits a separable structure. \Cref{prop:multidimensional-separable} may prove useful when deriving splitting-based iterative schemes, where the introduction of auxiliary variables can restore a componentwise structure in conditional models \citep{Vono2020,Faye2026}.
\end{remark}

\begin{remark} 
Since the negative log-likelihood $-\ell(\mathbf{y};x)$ associated with a regular exponential family distribution and its primal Bregman divergence only differs by the constant $-\phi_{f}(\mathbf{y})$, this property also provides a reference geometry for the Bregman divergence itself, yet expressed in the dual coordinates. In other words, the function ${\mathbf{t}} \mapsto d_f(\mathbf{y};{\tau}^{-1}({\mathbf{t}}))$ is $L$-smooth relative to ${h}$ for the  same range of $L$ values.
\end{remark}

\subsection{Discussion: from dual Bregman representation to relative smoothness}

The results established in this section establish a direct connection between dual Bregman representations and relative smoothness in regular exponential families. For instance, this  connection is particularly relevant to address Poisson inverse problems, where the negative log-likelihood typically has a gradient that is not globally Lipschitz on its natural domain. \citep{Bauschke2017} showed that this issue can be overcome by replacing the classical  Lipschitz-gradient condition with a Bregman descent condition. On an \emph{ad hoc} basis, they  identified the Burg entropy as one appropriate reference function to ensure the relative smoothness. This framework has subsequently been exploited in Poisson imaging, including Bregman plug-and-play methods \citep{Hurault2023} and mirror-descent-based deep equilibrium models \citep{Daniele2026}. It will be discussed more deeply in \Cref{sec:application}.

Conversely, \Cref{prop:beta-relative-smoothness}  and \Cref{prop:general-tau-tilde} offer a different perspective on this quest for an appropriate reference geometry. More precisely, for regular exponential family distributions, a reference geometry ensuring relative smoothness of the negative log-likelihood in an appropriate coordinate system arises directly from the dual Bregman representation of this distribution. In the $\beta$-scaled formulation, passing to the natural coordinates
$\mathbf t=\nabla f(\mathbf x)$ and using the generator $h=\beta f^*$ yields relative smoothness of the negative log-likelihood with respect to the same Bregman geometry. Likewise, the $\alpha$-scaled formulation induces its own transformed coordinate system and Bregman generator, which also provide a reference geometry
for relative smoothness. Thus, in both representations, the geometry arising from the dual Bregman representation of the density is naturally adapted to its own regularity.
Interestingly, within a Bayesian context, the dual Bregman representation of conjugate priors discussed in \Cref{subsec:application_priors} also reveals an intrinsic compatibility between the geometry of this prior and the geometry of the likelihood.

The resulting relative smoothness constants are explicitly determined by the scaling function and depend on the corresponding coordinate representation. In the $\beta$-scaled formulation, the sharp threshold is $L_{\min}=1/\beta$ and $L_{\min}=\left\|\boldsymbol{\beta}^{-1}\right\|_{\infty}$ in the separable multidimensional setting. In the $\alpha$-scaled formulation, the dual coordinates
${\mathbf{t}}={\tau}(\mathbf x)$ yield the sharp threshold $L_{\min}=y/\alpha$ in the scalar case and $L_{\min}=\|\mathbf y / \boldsymbol{\alpha} \|_\infty$ in the separable multidimensional setting. Hence, the two conjugate parametrizations lead to distinct, yet explicitly characterized, Bregman geometries in which the negative log-likelihood admits a quantitative descent control.

Finally, these results have a direct computational interpretation. Relative smoothness yields a Bregman descent inequality, which provides a basic estimate underlying Bregman and mirror-based first-order methods \citep{beck2003mirror,Bauschke2017}. The geometry induced by the dual Bregman representation is therefore not only relevant to the characterization of the parameter space, as stated in \Cref{sec:dual_representations}, but also provides a natural framework for controlling the associated likelihood. In this sense, these representations connect the statistical structure of regular exponential family models with the convex-analytic regularity properties that underlie Bregman-based computational methods. This connection is illustrated in the next section by addressing Poisson linear inverse problems in the proposed framework.

\section{Application to  Poisson linear inverse problems}
\label{sec:application}
To illustrate the relevance of the proposed geometrical framework, we propose to address Poisson linear inverse problems from the scope of the dual Bregman representations derived in Section \ref{sec:dual_representations} and further explored in Section \ref{sec:relative-smoothness-conjugacy}. In this context, the data consists in observed counts $\mathbf y\in\mathbb N^m$  related to an unknown parameter $\mathbf{x}\in\mathbb{R}^n$ to be recovered through the degradation model 
\begin{equation*}
    y_i \sim \mathcal{P}\bigl([{\linOperator} \mathbf x]_i\bigr),
    \qquad i=1,\ldots,m,
\end{equation*}
where ${\linOperator} \in\mathbb R_+^{m\times n}$ is a forward operator satisfying
${\linOperator} \mathbf 1_n>\mathbf 0$. Adopting a penalized maximum likelihood paradigm, the problem consists in solving the following minimization problem 
\begin{equation}\label{eq:poisson_problem}
   \operatornamewithlimits{min}_{\mathbf{x}\in\mathbb{R}^n_{++}} \objFun(\mathbf x) \triangleq -\ell({\mathbf y};{\linOperator} \mathbf x)+r(\mathbf x)
\end{equation}
where $-\ell({\mathbf y};\cdot)$ is the Poisson negative log-likelihood and $r(\cdot)$ is a regularization function. This problem is particularly relevant for the present framework since the Poisson distribution is naturally associated with non-Euclidean geometries. Indeed, \Cref{theo:derivation:banerjee:term} states that the Poisson negative log-likelihood admits the primal Bregman representation
\begin{equation*}
 -\ell(\mathbf{y};{\linOperator} \mathbf x) = d_f(\mathbf{y},{\linOperator} \mathbf{x}) -\phi_f(\mathbf{y})
 \end{equation*}
where the Bregman generator is the negative Boltzmann-Shannon entropy function $f(x) = x \log x -x $ on $\dom f=\R_+$ (with $0\log0=0$) and $\X=\R_+^*$, inducing the generalized Kullback-Leibler divergence (see \Cref{tab:bregman_families}). In particular, when ${\linOperator}  = \mathbf{I}$, according to \Cref{prop:beta-relative-smoothness,prop:general-tau-tilde}, the dual Bregman representations identified in \Cref{sec:dual_representations} provide two different reference geometries for establishing relative smoothness of the negative log-likelihood expressed in dual coordinate systems. 
To appreciate the different relative-smoothness structures induced by the two resulting geometries, 
we analyze the properties of the Bregman proximal gradient method (BPGM) as studied by \cite{Bauschke2017} where the regularization denoted by $r(\cdot)$ is  chosen as the $\ell_1$-norm sparsity-promoting penalization 
\begin{equation*}
 r(\mathbf x) = \mu \left\|\mathbf{x}\right\|_1 = \mu \sum_{i=1}^n x_i.
 \end{equation*}
When ${h}$ is a reference function ensuring that the negative log-likelihood satisfies the LC condition in the primal domain, BPGM minimizes a composite functional such as in \eqref{eq:poisson_problem} by iterating the updating rule
\newcommand{\stepsize}{\delta}
\begin{equation}
    \label{eq:PGD_definition}
    \mathbf x^+
    =
    \arg\min_{\mathbf u>0}
    \left\{ 
        \left\langle
            -\nabla \ell(\mathbf{y};{\linOperator} \mathbf x),
            \mathbf u-\mathbf x
        \right\rangle
        +
        r(\mathbf{u})
        +
        \frac{1}{\stepsize}
        d_{{h}}(\mathbf u,\mathbf x)
    \right\}
\end{equation}
where $\delta>0$ denotes the stepsize. In what follows, we propose to instantiate the BPGM in the dual coordinate system when these dual coordinates and the reference functions are derived from the  $\alpha$- or $\beta$-scaled dual Bregman representation of the Poisson distribution, i.e., with the two  generators $h$ specified in Table \ref{tab:geometry}. To easily assess the impact of the chosen geometry,  we will first consider the case ${\linOperator} =\mathbf I$ which perfectly matches the framework introduced in this paper. By removing the coupling between coordinates, this simple setting will also allow one to derive the solution of the problem \eqref{eq:poisson_problem} explicitly. It will enable the explicit characterization of the induced iterations, relative-curvature constants, and convergence rates of the algorithms.  Then, we will explore the case ${\linOperator}  \neq \mathbf I$, which sheds striking light on some existing results from the literature.

\subsection{Geometry induced by the \texorpdfstring{$\alpha$}{alpha}-scaled dual Bregman representation}
For the Poisson distribution, the $\alpha$-scaled dual Bregman representation  with $\alpha=1$  yields (see Table \ref{tab:geometry} and \Cref{app:subsec:poisson})
\begin{equation*}
    {\tau}(\mathbf x)=\mathbf x,
    \qquad
    {h}(\mathbf t)
    =-\sum_{j=1}^n\log t_j
\end{equation*}
and the Itakura-Saito divergence
\begin{equation*}
    \label{eq:IS_divergence}
    d_{{h}}(\mathbf u,\mathbf t)
    =
    \sum_{j=1}^n
    \left[
        \frac{u_j}{t_j}
        -
        \log\left(\frac{u_j}{t_j}\right)
        -1
    \right].
\end{equation*}

Interestingly, the choice of the Burg entropy as a suitable Legendre function ${h}$ was already considered in several works from the literature to derive BPGM for solving Poisson inverse problems \citep{Bauschke2017,tibshirani2025laplace,chirinos2026linear}. However it is worth noting that, in those works, this choice was never properly motivated but rather resulted from an ad hoc approach, driven by the search of one possible reference geometry ensuring the LC condition. In contrast, in our work, this reference geometry naturally emerges from a principled construction, as a direct consequence of the dual Bregman representation exhibited in \Cref{cor:alpha_existence} and underlying \Cref{prop:general-tau-tilde}.

\subsubsection{Case \texorpdfstring{${\linOperator} =\mathbf I$}{K=I}}
We first assume that 
\begin{equation*}
    y_j>0, \qquad j=1,\ldots,n.
\end{equation*}
The particular case of zero-count will be discussed later (see \Cref{rem:alpha_zero_count}). When ${\linOperator}  = \mathbf I$, the objective function writes 
\begin{equation*}
    \objFun(\mathbf x)
    =
    \sum_{j=1}^n
    \left[
        (1+\mu)x_j-y_j\log x_j
    \right]
\end{equation*}
and its unique minimizer is 
\begin{equation}
    \label{eq:identity_xstar_comparison}
    \mathbf{x}^\star = \frac{1}{1+\mu}\mathbf{y}.
\end{equation}
Since ${\tau}=\mathrm{Id}$, according to \Cref{prop:general-tau-tilde}, the Burg entropy ${h}$ provides a suitable reference geometry to ensure relative smoothness of the Poisson negative log-likelihood function in the primal and the dual coordinate systems equivalently. Thus, the BPGM iteration \eqref{eq:PGD_definition} can be implemented in the primal domain and, given that $    \nabla_{\mathbf u}     d_{{h}}(\mathbf u,\mathbf x)     =     \mathbf x^{-1}-\mathbf u^{-1}$, the first-order optimality condition, leads to the following  multiplicative update rules
\begin{equation}
    \label{eq:alpha_PGD}
    \mathbf x^+
    =
    \frac{\mathbf x}
    {
        \mathbf 1
        +
        \stepsize         
           ( (\mu+1) \mathbf x -\mathbf y)        
    }
\end{equation}
where the division should be understood component-wise.  Assume now that
\begin{equation}
    \label{eq:alpha_stepsize}
    0<\stepsize\leq\frac{1}{y_{\max}}
\end{equation}
with $y_{\max}= \left\| \mathbf y \right\|_{\infty}$. Since $y_j>0$ for every $j$, this implies
\begin{equation}\label{eq:alpha_stepsize_constraint}
    0\leq1-\stepsize y_j<1
\end{equation}
and the denominator in \eqref{eq:alpha_PGD} satisfies
\begin{align*}
     1+\stepsize((1+\mu)x_j -y_j)>0
\end{align*}
for every $x_j>0$. This condition guarantees that the BPGM iterations are geometrically compliant, i.e., they are well defined and remain in the positive orthant for every initialization $\mathbf x^0\in\mathbb R_{++}^n$. To further analyze this iterative scheme which is component-wise separable, consider the reciprocal variable
\begin{equation*}
    {v}_j = \frac{1}{{x}_j}.
\end{equation*}
By using \eqref{eq:identity_xstar_comparison}, the updating rule \eqref{eq:alpha_PGD} becomes
\begin{equation}\label{eq:alpha_exact_error}
    {v}_j^+ =  {v}_j^\star +  (1- \stepsize {y}_j) ( {v}_j -{v}_j^\star).
\end{equation}
and, after $k$ iterations,
\begin{equation}
\label{eq:alpha_exact_error_k}
    {v}_j^k =  {v}_j^\star +  (1- \stepsize {y}_j)^{k}  ( {v}_j^0 -{v}_j^\star).
\end{equation}
Moreover, let define
\begin{equation*}
 q_{\alpha} = 1- \stepsize y_{\min}
\end{equation*}
with $y_{\min}=\min_j y_j>0$. Since the stepsize $\stepsize$ satisfies \eqref{eq:alpha_stepsize_constraint}, we have 
\begin{equation*}
    0\leq1-\stepsize y_j
    \leq
    1-\stepsize y_{\min}
    <1
\end{equation*}
and it follows from~\eqref{eq:alpha_exact_error_k} that
\begin{equation*}
    \|
        \mathbf v^k-\mathbf v^\star
    \|_\infty
    \leq
    q_{\alpha}^k
    \|
        \mathbf v^0-\mathbf v^\star
    \|_\infty.
\end{equation*}
Thus the iteration converges globally and linearly in the reciprocal coordinates, with exact worst-case contraction factor $q_{\alpha}$. At the largest stepsize $\stepsize=\frac{1}{y_{\max}}$ allowed by~\eqref{eq:alpha_stepsize}, the exact worst-case contraction factor is
\begin{equation}
\label{eq:alpha_canonical_rate}
    q_{\mathrm{IS}}=  1-\frac{y_{\min}}{y_{\max}}.
\end{equation}
The continuity of the reciprocal mapping towards the original domain on $(0,\infty)$ gives $\lim_{k \rightarrow +\infty}  \mathbf x^k = \mathbf x^\star$ for any $\mathbf x^0\in\mathbb R_{++}^n$. Hence the convergence statement in the original variables follows directly from the exact linear convergence established in the reciprocal coordinates.

\begin{remark}
This result also admits a natural interpretation in terms of relative geometry. Indeed, 
\begin{equation*}
    \nabla^2\objFun(\mathbf x)
    =\operatorname{Diag} \left(\frac{y_j}{x_j^2}  \right),
\end{equation*}
whereas the Burg generator induced by the $\alpha$-scaled dual Bregman representation
satisfies
\begin{equation*}
    \nabla^2{h}(\mathbf x)
    =\operatorname{Diag}
    \left( \frac{1}{x_j^2} \right).
\end{equation*}
Hence the two Hessians exhibit the same dependence on the variable $x_j$, and differ only through the coefficients $y_j$. Consequently,
\begin{equation} \label{eq:alpha_convexity_smoothness}
    y_{\min}\nabla^2{h}(\mathbf x)
    \preceq \nabla^2 \objFun(\mathbf x) \preceq  y_{\max}\nabla^2{h}(\mathbf x).
\end{equation}
The exact relative strong-convexity and relative-smoothness constants are thus 
\begin{equation*}
    \mu_{\alpha}=y_{\min} \quad
    \text{and} \quad L_{\alpha}=y_{\max}.
\end{equation*}
Introducing the corresponding relative condition number
\begin{equation*}
    \kappa_{\alpha}= \frac{L_\alpha}{\mu_{\alpha}}
    = \frac{y_{\max}}{y_{\min}},
\end{equation*}
the exact contraction factor obtained in
\eqref{eq:alpha_canonical_rate} can equivalently be written as
\begin{equation*}
    q_{\alpha}=  1-\frac{\mu_{\alpha}}{L_{\alpha}}
    = 1-\frac{1}{\kappa_{\alpha}}.
\end{equation*}
This is the familiar dependence of a gradient-type contraction factor
on a condition number, here expressed in the relative Burg geometry
rather than in the Euclidean geometry. In particular, $\kappa_{\alpha}=1$ corresponds to perfect relative conditioning and gives $q_{\alpha}=0$, whereas the convergence becomes progressively slower as $\kappa_{\alpha}$ increases. It is worth emphasizing that, in the present case with ${\linOperator} = \mathbf{I}$, this expression is not obtained by applying a generic convergence bound. The contraction factor follows directly from the exact scalar recursion~\eqref{eq:alpha_exact_error}. The relative-curvature analysis
shows, independently, that this exact factor can be expressed as
$1-\mu_{\alpha}/L_{\alpha}$.
\end{remark}

\begin{remark}[Case of zero-count]\label{rem:alpha_zero_count}
  If $y_j=0$, the corresponding scalar objective reduces to
\begin{equation*}
\objFun_j(x_j) = (1+\mu)x_j
\end{equation*}
The infimum over $x_j>0$ is approached at the boundary
$x_j^\star=0$, where the continuous extension attains its minimum.
The BPGM iteration \eqref{eq:alpha_PGD} becomes
\begin{equation*}
x_j^+ = \frac{x_j}{1+\stepsize(1+\mu)x_j},
\end{equation*}
or equivalently
\begin{equation*}
 \frac1{x_j^+} =\frac1{x_j} + \stepsize(1+\mu).
\end{equation*}
and after $k$ iterations
\begin{equation*}
\frac1{x_j^k} = \frac1{x_j^0}  + k\stepsize(1+\mu).
\end{equation*}
Consequently, $\lim_{k\rightarrow +\infty} x_j^k =0$ and the convergence towards the boundary point is sublinear of order $O(k^{-1})$.
\end{remark}

\subsubsection{Case \texorpdfstring{${\linOperator} \neq \mathbf I$}{K not identity}}
When ${\linOperator} \neq \mathbf{I}$, the linear operator couples the components of $\mathbf{x}$ in the likelihood function, which makes the divergence $d_f(\mathbf{y},{\linOperator}\mathbf{x})$ not separable. Thus the assumptions required by \Cref{prop:general-tau-tilde} or \Cref{prop:multidimensional-separable} are not met and these statements are not sufficient to ensure relative smoothness of the Poisson negative log-likelihood. However, as already highlighted (see Remark \ref{remark:poisson:bauschke}), \cite{Bauschke2017} filled this gap by stating the following property.

\begin{lemma}\citep[Lemma 7]{Bauschke2017}\label{Lsmooth_poisson_alpha_AnotI}
Consider the Bregman representation $\ell(\mathbf{y};{\linOperator} {\mathbf{x}}) = - d_f(\mathbf{y};{\linOperator} {\mathbf{x}}) + \phi_f(\mathbf{y})$ of the log-likelihood associated with a Poisson linear inverse problem. 
Let $\check\lambda(\mathbf{y};\mathbf{x})$ denote the negative log-likelihood function, i.e.,
    \begin{align*}
        \check\lambda(\mathbf{y};\mathbf{x}) &= d_f(\mathbf{y};{\linOperator} {\mathbf{x}}) - \phi_f(\mathbf{y}).
    \end{align*}
    Let define $h(\mathbf{x}) = - \sum_{i=1}^n \log{x_i}$.  Then the function $\check\lambda(\mathbf{y};\mathbf{x})$ is $L$-smooth relative to $h$ for every $L\geq  L_{\mathrm B} =\|\mathbf y\|_1 $.
\end{lemma}

This proposition shows that the reference geometry identified by applying \Cref{prop:general-tau-tilde} and corresponding to the case ${\linOperator} =\mathbf{I}$ remains suitable to ensure relative smoothness of the negative log-likelihood in the case ${\linOperator} \neq\mathbf{I}$, yet with a different constant $L_{\mathrm B}$. The BPGM iterations \eqref{eq:PGD_definition} can thus be derived in this case to generalize \eqref{eq:alpha_PGD}. They write
\begin{equation*}
    \mathbf{x}^+ =\frac{\mathbf{x}}{
 1+ \stepsize \mathbf{x} \odot\left(\mu \mathbf{I} + \mathbf{s} -{\linOperator} ^\top \frac{\mathbf{y}}{{\linOperator}\mathbf{x}})
 \right) }
\end{equation*}
where $\mathbf{s} = \left(s_1,\ldots,s_n\right)$ with $s_j = \sum_{i=1}^m k_{i,j}$. As expected this iterative scheme perfectly matches the so-called Nolips algorithm derived by \cite{Bauschke2017} to solve \eqref{eq:poisson_problem}. In other words, even when ${\linOperator}  \neq \mathbf{I}$, the $\alpha$-scaled Bregman representation recovers the Itakura-Saito geometry and the corresponding NoLips update.  Besides, given the convexity and smoothness characterization \eqref{eq:alpha_convexity_smoothness}, when $\mathbf y>0$, recall that the exact relative constants in the identity case are
\begin{equation*}
\mu_{\alpha}=y_{\min},
    \qquad  L_{\alpha}=y_{\max}.
\end{equation*}
In particular, $y_{\max}\leq\|\mathbf y\|_1$, so the general relative-smoothness bound $L_{\mathrm B}$ identified by \cite{Bauschke2017} in the general case can be sharpened, in the
identity case, to the exact constant $y_{\max}$.

\subsection{Geometry induced by the \texorpdfstring{$\beta$}{beta}-scaled dual Bregman representation}
We now consider the geometry induced by the $\beta$-scaled dual Bregman representation as discussed in \Cref{subsec:smoothness_geometry_beta}. More precisely, as in \Cref{prop:multidimensional-separable_beta}, we consider a separable generator $h$ parametrized by non-negative weights $\beta_j>0$, which leads to (see Table \ref{tab:geometry} and \Cref{app:subsec:poisson})
\begin{equation*}
    \tau(\mathbf x)=\log\mathbf x, \quad h(\mathbf t)= \sum_{j=1}^n \beta_j e^{t_j}.
\end{equation*}
and the generalized Kullback-Leibler divergence $\text{KL}(\by\|\bx)$ expressed in the dual coordinates
\begin{equation*}
    d_h(\mathbf{u},\mathbf t)
    = \sum_{j=1}^n \beta_j \left[e^{u_j} -e^{t_j}-  e^{t_j}({u_j}-{t_j})\right].
\end{equation*}
Several choices of the weights $\boldsymbol{\beta}=(\beta_1,\ldots,\beta_n)^\top$ will be discussed along this section.

\subsubsection{Case \texorpdfstring{${\linOperator} =\mathbf I$}{K=I}}
We first assume that 
\begin{equation*}
    y_j>0, \qquad j=1,\ldots,n.
\end{equation*}
The particular case of zero-count will be discussed later (see \Cref{rem:beta_zero_count}). In the identity case, the objective function \eqref{eq:poisson_problem} expressed in the dual coordinates is
\begin{align}
    \widetilde \objFun(\mathbf t) & \triangleq  \objFun(e^{\mathbf t}) \nonumber \\
    &=  \lambda (\mathbf{y};\mathbf t) + r(e^\mathbf{t})
    \label{eq:beta_natural_objective}
\end{align}
where $\lambda (\mathbf{y};\mathbf t)$ has been defined in \eqref{eq:lambda_beta} as the negative log-likelihood expressed in the dual coordinates, which leads to
\begin{align}
    \widetilde \objFun(\mathbf t) =\sum_{j=1}^n \left[(1+\mu)e^{t_j}-y_jt_j \right]    =(1+\mu)h(\mathbf t) -\langle\mathbf y,\mathbf t\rangle.
    \label{eq:beta_natural_objective_bis}
\end{align}
According to \Cref{prop:multidimensional-separable_beta}, the generator $h$ provides a suitable reference geometry
to ensure relative smoothness of the Poisson negative log-likelihood function in the dual coordinate system. Under the choice $\beta_j=\beta=1$, the BPGM iteration \eqref{eq:PGD_definition} can thus be rewritten in this dual domain as 
\begin{equation}\label{eq:beta_PGD_dual}
    \mathbf t^+
    = \arg\min_{\mathbf u}
    \left\{  \left\langle
            \nabla\lambda (\mathbf{y};\mathbf t),
            \mathbf u-\mathbf t
        \right\rangle
        + r(\tau^{-1}(\mathbf{u})) +
        \frac{1}{\stepsize}
        d_h(\mathbf u,\mathbf t).
    \right\}
\end{equation}
This iteration expressed in the dual domain has been recently proposed by \cite{modrzyk2026right} as the \emph{right}  Bregman proximal gradient. The first-order optimality condition leads to the following updating rule
\begin{equation*}
    e^{\mathbf t^+}
    =  \frac{1-\stepsize}{1+\mu \stepsize} e^{\mathbf t} + \frac{\stepsize}{1+\mu \stepsize} \mathbf{y}.
\end{equation*}
or, in the primal domain, 
\begin{equation}\label{eq:beta_iterate_x}
    {\mathbf x^+}
    =  \frac{1-\stepsize}{1+\mu \stepsize} {\mathbf x} + \frac{\stepsize}{1+\mu \stepsize} \mathbf{y}.
\end{equation}
Note that the positivity of the primal variable $\mathbf{x}$ is ensured by construction. By using \eqref{eq:identity_xstar_comparison}, this update can be rewritten as
\begin{equation*}
    {\mathbf x^+}
    =  \mathbf{x}^\star + \frac{1-\stepsize}{1+\mu \stepsize} ({\mathbf x} -\mathbf x^\star)
\end{equation*}
and, after $k$ iterations,
\begin{equation*}
    {\mathbf x^k}
    =  \mathbf x^\star + \left(\frac{1-\stepsize}{1+\mu \stepsize}\right)^k ({\mathbf x}^0 -\mathbf x^\star).
\end{equation*}
Let define
\begin{equation*}
    q_{\beta} = \frac{1-\stepsize}{1+\mu \stepsize}
\end{equation*}
with
\begin{equation} \label{eq:beta_stepsize}
    0 < \stepsize \leq 1.
\end{equation}
Then
\begin{equation*}
    \|
        \mathbf x^k-\mathbf x^\star
    \|_\infty
    \leq
    q_{\beta}^k
    \|
        \mathbf x^0-\mathbf x^\star
    \|_\infty.
\end{equation*}
Thus the iteration converges globally and linearly in the primal domain, with exact contraction factor $q_{\beta}$. At the largest stepsize $\stepsize=1$ allowed by~\eqref{eq:beta_stepsize}, the contraction factor vanishes and $\mathbf{x}^k = \mathbf{x}^\star$, for every $k\ge 1$. Thus, when ${\linOperator}  = \mathbf{I}$, the BPGM implemented in the dual
coordinates reaches the minimizer in a single step. This one-step property follows from the exact geometric matching in \eqref{eq:beta_natural_objective_bis}: using the dual coordinates
$\mathbf t=\log\mathbf x$, the nonlinear part of the objective function is driven by $h(\mathbf t)$, while the remaining term is affine.

\begin{remark}
This result also admits a direct interpretation in terms of relative smoothness in the natural coordinates. Indeed, from \eqref{eq:beta_natural_objective_bis}, we have
\begin{equation*}
\nabla^2\widetilde \objFun(\mathbf t) =(1+\mu)\nabla^2 h(\mathbf t).
\end{equation*}
Thus, the objective function in the dual coordinate system is both relatively strongly convex and relatively smooth with respect to $h$, with exact constants
\begin{equation*}
\mu_{\beta}=L_{\beta} =1+\mu.
\end{equation*}
The corresponding relative condition number is therefore
\begin{equation*}
\kappa_{\beta}=\frac{L_{\beta}}{\mu_{\beta}}
= 1.
\end{equation*}
This perfect relative conditioning reflects the exact geometric matching between the objective and the reference geometry induced by the Bregman generator $h$. As in the $\alpha$-scaled case, this property is not stated after merely applying a generic convergence bound. The contraction factor follows directly from the exact recursion \eqref{eq:beta_iterate_x}; the relative-curvature analysis shows, independently, that the one-step behavior corresponds to perfect conditioning in the natural $\beta$-scaled geometry.
\end{remark}

\begin{remark}[Case of zero-count]\label{rem:beta_zero_count}
For $y_j=0$, the BPGM iteration \eqref{eq:beta_iterate_x} reduces to
\begin{equation*}
x_j^+ = \frac{1-\stepsize}{1+\mu \stepsize}x_j
\end{equation*}
and, after $k$ iterations,
\begin{equation*}
x_j^k = \left(\frac{1-\stepsize}{1+\mu \stepsize}\right)^kx_j^0.
\end{equation*}
Hence, for $0<\stepsize<1$ the iterates remain strictly positive and converge geometrically to the boundary. At the limiting stepsize $\stepsize=1$, the formal primal recursion gives $x_j^+=0$  in a single iteration. However, this value does not correspond to a finite
BPGM iterate in the dual domain. Indeed, when $y_j=0$ and $\stepsize=1$, the corresponding dual subproblem has no finite minimizer: its infimum is approached as $t_j\to-\infty$. Since $x_j=e^{t_j}$, this limiting behavior corresponds to $x_j\to 0$ in the primal domain. Hence, the
one-step statement at $\stepsize=1$ should be understood as a limiting statement in the primal representation, rather than as an attained
finite BPGM iterate in the dual domain.
\end{remark}

\subsubsection{Case \texorpdfstring{${\linOperator} \neq \mathbf I$}{K not identity}}
When ${\linOperator}  \neq \mathbf{I}$, the negative log-likelihood is not separable with respect to the parameter components. This coupling is not compatible with the assumptions required by \Cref{prop:multidimensional-separable_beta}, which cannot ensure relative smoothness of the negative log-likelihood with respect to the Bregman generator $h$. However, we can show that this generator still provides a suitable reference geometry, as stated in the following lemma which extends the result of \Cref{prop:multidimensional-separable_beta} for the Poisson distribution specifically.

\begin{lemma}\label{Lsmooth_poisson_beta_AnotI}
Consider the Bregman representation $\ell(\mathbf{y};{\linOperator} {\mathbf{x}}) = - d_f(\mathbf{y};{\linOperator} {\mathbf{x}}) + \phi_f(\mathbf{y})$ of the log-likelihood associated with a Poisson linear inverse problem. Let $\check\lambda(\mathbf{y};\mathbf{t})$ denote the negative log-likelihood function expressed in the dual coordinates $\mathbf{t} =\log (\mathbf{x})$ induced by a component-wise $\beta$-scaled dual Bregman representation, i.e.,
    \begin{align*}
        \check\lambda(\mathbf{y};\mathbf{t}) &= d_f(\mathbf{y};{\linOperator} e^{\mathbf{t}}) - \phi_f(\mathbf{y}).
    \end{align*}
    For $\boldsymbol{\beta} = (\beta_1,\ldots,\beta_n)^\top$ and $\beta_i>0$, let define $h(\mathbf{t}) = \sum_{i=1}^n \beta_i e^{t_i}$. Set $\mathbf{s} = \left(s_1,\ldots,s_n\right) = {\linOperator} ^\top \boldsymbol{1}$, i.e., $s_j = \sum_{i=1}^m k_{i,j}$, with $s_j>0$ for every $j=1,\dots,n$. 
    Then the function $\check\lambda(\mathbf{y};\mathbf{t})$ is $L$-smooth relative to $h$ for every $L\geq L_{\inf}$ with
    \begin{equation*}
        L_{\inf} = \left\{
        \begin{matrix}
            1   &  \text{when}  &\boldsymbol{\beta} = \boldsymbol{s}, \\
           \left\|{\linOperator} \right\|_1  & \text{when}  &\boldsymbol{\beta} = \boldsymbol{1}.
        \end{matrix}
        \right.
    \end{equation*}
\end{lemma}

\begin{proof}
    See Appendix \ref{app:proof_Lsmooth_poisson_beta_AnotI}
\end{proof}
In other words, \Cref{prop:multidimensional-separable_beta} still provides a suitable reference geometry for the Poisson distribution, despite the linear coupling of the parameter components. This result can thus lead to two different valid BPGM iteration schemes, depending on the choice of the weights. Whatever this choice, the BPGM update \eqref{eq:beta_PGD_dual} with
\begin{equation*}
    \nabla\check\lambda(\mathbf y ; \mathbf{t}) = e^{\mathbf{t}} \odot {\linOperator} ^\top \left(\boldsymbol{1}-\frac{\mathbf{y}}{{\linOperator} e^{\mathbf t}}\right)
\end{equation*}
writes
\begin{equation*}
    e^{\mathbf{t}^+} = e^{\mathbf{t}} \odot \frac{\left[\boldsymbol{\beta} -  \stepsize  {\linOperator} ^{\top} \left(\boldsymbol{1}-\frac{\mathbf{y}}{{\linOperator}  e^{\mathbf t}}\right) \right] }{\boldsymbol{\beta} +  \stepsize\mu \boldsymbol{1} }
\end{equation*}
or, in the primal domain,
\begin{equation}\label{eq:appli:beta:primal}
  {\mathbf{x}}^+ = {\mathbf{x}} \odot \frac{\left[ \boldsymbol{\beta} -  \stepsize \left( {\linOperator} ^{\top} \left(\boldsymbol{1}-\frac{\mathbf{y}}{{\linOperator}  {\mathbf x}}\right)\right) \right]}{\boldsymbol{\beta} +\stepsize\mu\boldsymbol{1}  }.
\end{equation}

This BPGM update is well defined provided that the right-hand side of \eqref{eq:appli:beta:primal} belongs to  $(\mathbb R_+^*)^n$. Interestingly, when $\mu =0$ (i.e., in the absence of regularization), this update simplifies to
\begin{equation*}
    {\mathbf{x}^+} ={\mathbf{x}} \odot \left[ \boldsymbol{1} - \frac{\stepsize}{\boldsymbol{\beta}} \odot {\linOperator} ^{\top} \left(\boldsymbol{1}-\frac{\mathbf{y}}{{\linOperator}  {\mathbf x}}\right) \right].
\end{equation*}
Now, let assume that this iteration is  implemented with the choice $\boldsymbol{\beta} = \mathbf{s}$ and associated canonical stepsize $\stepsize=1$, this iteration then reduces to
\begin{align}\label{eq:step_EM}
    \mathbf{x}^+
    &=
    \frac{\mathbf{x}}{\mathbf{s}}
    \odot
    \left(
    {\linOperator}^{\top}
    \frac{\mathbf{y}}{{\linOperator} \mathbf{x}}
    \right).
\end{align}
A similar update would be have obtained with the alternative choice $\boldsymbol{\beta} = \boldsymbol{1}$ and using an anisotropic stepsize, i.e., $\stepsize_i = \frac{1}{s_i}$ or, equivalently, a diagonal preconditioning
matrix
\begin{equation*}
    \mathbf{P}= \operatorname{diag}\left(
    \frac{1}{s_1},\ldots,\frac{1}{s_n}
    \right).
\end{equation*}
With both choices, the step \eqref{eq:step_EM} coincides with the expectation-maximization (EM) iteration \citep{shepp1982maximum}, or equivalently with the Richardson-Lucy iteration \citep{lucy1974iterative,richardson1972bayesian}, for Poisson data. Thus, the adopted $\beta$-scaled dual geometry turns BPGM expressed in the dual domain into a multiplicative iterative scheme in the primal domain and recovers the EM/Richardson-Lucy iteration. Connections between EM and Bregman or mirror-descent methods have also been investigated, for instance, by \cite{kunstner2021homeomorphic} and \cite{clerc2026reinterpreting}. It is also interesting to compare this construction with the so-called \emph{right} BPGM proposed by \cite{modrzyk2026right} and applied to Poisson
linear inverse problems. In their approach,  the vector
$\mathbf{s}={\linOperator} ^{\top}\boldsymbol{1} $ also serves to define a weighted reference (negative-entropy) geometry, akin to the separable construction of $\beta$-scaled models proposed in \Cref{prop:multidimensional-separable_beta}. Their approach leads, in the
unregularized Poisson linear inverse problem, to
\begin{equation*}
    \mathbf{x}^{k+1}
    = (1-\stepsize)\mathbf{x}^{k} +
    \stepsize\, \frac{\mathbf{x}^{k}}{\mathbf{s}}
    \odot {\linOperator} ^{\top}
    \frac{\mathbf{y}}{{\linOperator} \mathbf{x}^{k}},
\end{equation*}
which reduces to the EM/Richardson-Lucy iteration for $\stepsize=1$. Hence, both approaches recover the same EM/Richardson-Lucy iteration, despite different starting points. \cite{modrzyk2026right} devise a new variant of BPGM to solve the  problem \eqref{eq:poisson_problem} in the primal domain and leverages the Legendre duality to reformulate the iterations with respect to the dual variables. On the contrary, we apply the canonical instance of BPGM to solve the  problem \eqref{eq:poisson_problem} in the dual domain induced by the $\beta$-scaled Bregman representation.

\subsection{Summary of the case \texorpdfstring{${\linOperator}  = \mathbf{I}$}{K=I}}
For ${\linOperator} =\mathbf I$ and $\mathbf y>0$, the two derivations of BPGM iterations associated with the two geometries lead to explicit iterations with distinct convergence
properties. Adopting the geometry induced by the $\alpha$-scaled dual Bregman representation, BPGM yields the Burg/Itakura-Saito iterations in the primal domain. Introducing the reciprocal
variable $v_j=1/x_j$ yields the exact affine error recursion 
\[v_j^+-v_j^\star = (1-\stepsize y_j)(v_j-v_j^\star). \]
Hence, for $0<\stepsize\leq 1/y_{\max}$, the iteration is globally linearly convergent in reciprocal coordinates, with exact worst-case coordinate-wise contraction factor
\[q_{\alpha} =1-\stepsize y_{\min}. \]
At the largest simple globally positive stepsize $\stepsize=1/y_{\max}$, this becomes \[q_{\alpha}=1-\frac{y_{\min}}{y_{\max}}. \]
This exact rate is consistent with the relative geometry of the identity problem, for which the relative strong-convexity and smoothness constants are respectively $y_{\min}$ and $y_{\max}$.

Adopting the geometry induced by the $\beta$-scaled dual Bregman representation leads to a BPGM iteration in dual coordinates.  The objective function is rewritten as
\[\widetilde \objFun(\mathbf t)
    =(1+\mu)h(\mathbf t) -\langle\mathbf y,\mathbf t\rangle \]
where the  nonlinear part of this objective function  matches the Bregman generator $h(\mathbf t)=\sum_{j=1}^n e^{t_j}$ of the $\beta$-scaled dual representation exactly: the objective function and the reference function share the same geometry, since
\[\nabla^2\widetilde \objFun(\mathbf t)=(1+\mu)\nabla^2 h(\mathbf t). \]
The resulting BPGM can be mapped back to
the original variable of interest,
\[\mathbf x^+ =  \frac{1-\stepsize}{1+\mu \stepsize} \mathbf x+\frac{\stepsize}{1+\mu\stepsize}\mathbf y,\]
and  satisfies the exact global error recursion
\[\mathbf x^+-\mathbf x^\star=\frac{1-\stepsize}{1+\mu \stepsize}
    (\mathbf x-\mathbf x^\star).\]
At the canonical stepsize $\stepsize=1$, the contraction factor vanishes and the minimizer is reached in a single iteration: \[\mathbf x^+=\frac{\mathbf y}{1+\mu}    =\mathbf x^\star.\]

The identity case shows that the two geometries induced by the proposed dual Bregman representations provide substantially different convergence dynamics and that a given geometry does not determine a unique algorithmic implementation. In particular, the one-step convergence property of the $\beta$-scaled construction results from the derivation of BPGM in the dual coordinates together with the exact matching of the objective and the reference function. Table~\ref{tab:identity_summary} summarizes these findings, highlighting
the distinct updates, convergence properties, and limiting-step
behavior of the BPGM in the $\alpha$-scaled and $\beta$-scaled geometries. This table also specifies the particular behaviors of the two algorithms when the observations lie on the boundary of the corresponding parameter space, i.e., $y_i=0$.

\begin{table}[ht!]\small
\centering
\renewcommand{\arraystretch}{1.6}
\resizebox{0.8\textwidth}{!}{%
\begin{tabular}{c|c|c|c}
\toprule
\multicolumn{2}{c|}{} & \textbf{$\alpha$-scaled geometry} & \textbf{$ \beta$-scaled geometry}
\\ 
\midrule
 \multicolumn{2}{c|}{{Coordinates}}  & $\mathbf x$   & $ \mathbf t=\log\mathbf x$ \\
 
 \multicolumn{2}{c|}{{Bregman divergence}} & $\displaystyle d_{{h}}(\mathbf u,\mathbf x)$ & $\displaystyle d_h(\mathbf u,\mathbf t)$ \\

 \multicolumn{2}{c|}{{Natural analysis variable}}
& $ \mathbf v=\frac{1}{\mathbf{x}}= | \nabla {h}(\mathbf x)|$
& $\displaystyle \mathbf x=e^{\mathbf t}=\nabla h(\mathbf t)$
\\

 \multicolumn{2}{c|}{{Initialization}}
& $\mathbf x^0>0$ arbitrary 
& $\mathbf x^0>0$ arbitrary \\
\hline
\multirow{8}{*}{\rotatebox{90}{$\mathbf{y}>0$}} & {{Update in $\mathbf x$}}  
  &  $\displaystyle \mathbf x\oslash \left[ \mathbf 1+ \stepsize((\mu+1)\mathbf x - \mathbf y) \right]$  
  &  $ \frac{1}{1+\mu\delta} \left((1-\delta )\mathbf{x} + \delta \mathbf{y}\right)$ \\

 & Dynamics 
&$\displaystyle  v_j^+-v_j^\star = (1-\stepsize y_j)(v_j-v_j^\star)$
&  $     { x_j^+} -  {x}_j^\star = \frac{1-\stepsize}{1+\mu \stepsize} ({ x_j} - x_j^\star)$
\\

& Stepsize condition  
& $  0<\stepsize\leq\frac{1}{y_{\max}}$ 
& $\displaystyle 0<\stepsize\leq 1$\\
 
& Limiting stepsize $\delta_{\mathrm{lim}}$ 
 & $ \stepsize_{\mathrm{lim}}=\frac{1}{y_{\max}}$ 
 & $\displaystyle \stepsize_{\mathrm{lim}}=1$ \\
 
& Convergence ($\delta < \delta_{\mathrm{lim}}$)
 & global linear in $\mathbf x^{-1}$ 
 & global linear in $\mathbf x$ \\
 
& \multirow{2}{*}{{Contraction / rate}}
 & $\displaystyle q_{\alpha} = 1-\stepsize y_{\min}$ 
 & $ q_{\beta} = \frac{1-\stepsize}{1+\mu \stepsize}$ \\[-2mm]
& & {\footnotesize exact worst-case coordinate-wise} 
 & {\footnotesize exact global, $0<\stepsize<1$}\\

& Behavior at $\delta = \delta_{\mathrm{lim}}$
&$ q_{\alpha} = 1-\frac{y_{\min}}{y_{\max}}$
& { one-step convergence}
\\
\hline

\multirow{8}{*}{\rotatebox{90}{${y_i}=0$}} & {{Update in $x$}}  
 &  $\displaystyle x_j / \left(  1+ \stepsize(\mu+1) x_j \right)$  
  &  $ \frac{1-\delta}{1+\mu\delta} x_j $   \\
  
& Dynamics 
&$\displaystyle  v_j^+ = v_j+\delta\left(1+\mu\right)$
&  $     { x_j^+} = \frac{1-\stepsize}{1+\mu \stepsize} { x_j}$  \\

& Stepsize condition  
& $0<\delta$
& $0<\delta\le 1$\\

& Limiting stepsize $\delta_{\mathrm{lim}}$
 & None
 & $\delta_{\mathrm{lim}}=1$ \\

& Convergence ($\delta < \delta_{\mathrm{lim}}$)
 & sublinear in $x^{-1}$
 & global linear in $x$\\
 
& \multirow{2}{*}{{Contraction / rate}}
 &  $x_j^k \sim \frac{1}{\delta k}$
 &  $q_{\beta} = \frac{1-\stepsize}{1+\mu \stepsize}$\\[-2mm]
& & {\footnotesize no contraction }
 & {\footnotesize exact global, $0<\stepsize<1$}\\

& Behavior at $\delta = \delta_{\mathrm{lim}}$
& N/A
&  {one-step convergence}\\  
\bottomrule
\end{tabular}}
\caption{{Summary of the main convergence properties of the BPGM iterations implemented 
in the $\alpha$-scaled  and $\beta$-scaled geometries.}}
\label{tab:identity_summary}
\end{table}

\section{Conclusion}
\label{sec:conclusion}

In this paper, we developed a geometric framework for devising dual Bregman representations of distributions from regular exponential families. Building upon the primal Bregman representation of these distributions, we showed that the arguments involved in the Bregman divergence can be swapped after passing to a new coordinate system. Building on a projective characterization of such representations,
we established a general construction of the dual coordinates and their associated Bregman generators, together with normalization and uniqueness results. We also identified two particular forms of admissible dual representations. One of them, referred to as the $\beta$-scaled dual Bregman representation, is directly induced
by Legendre duality, with the gradient of the density generator defining the dual coordinates and its scaled Legendre conjugate defining the corresponding Bregman geometry. A second family of models, referred to as the $\alpha$-scaled dual Bregman representation, adopts explicit yet distinct coordinates and Bregman geometry.
Both constructions arise as particular instances of the general projective framework. One particular application of these representations was the derivation of alternative forms of conjugate priors within a Bayesian inference context.
The proposed framework was also instantiated on several classical statistical models, showing that different exponential-family models may share the same Bregman geometry, up to a positive scaling.

Beyond the construction of these dual Bregman representations, we established a direct connection between these representations and relative smoothness. Precisely, the Bregman geometries induced by these representations provide natural reference geometries for the associated negative log-density, with explicit relative
smoothness constants determined by the conjugate scaling.
In the scalar and separable settings, the corresponding constants are sharp.

These results connect statistical modeling, convex duality, and Bregman-based regularity within a common geometric framework. Beyond their theoretical interest, the geometries induced by these representations may provide useful reference structures
for the development and analysis of Bregman- and mirror-based computational methods for exponential-family models. This was illustrated by addressing Poisson linear inverse problems. The derivation of Bregman proximal gradient methods
under the two geometries induced by the dual representations recovers popular algorithms from the literature, shedding new light on their constructions and properties.

Several directions deserve further investigation. Extending the present framework beyond regular exponential families and exploiting the induced geometries for the design and analysis of Bayesian computational algorithms constitute promising directions for future research.

\section*{Acknowledgements}

The authors would like to thank Pr. Isma\"el Castillo (LPSM, Sorbonne Université, France) for his helpful comments and suggestions, and Dr. Emmanuel Soubies (IRIT, Univ. Toulouse, France) for stimulating discussions regarding this work and for pointing out the reference \citep{modrzyk2026right}.

M. D. Fall acknowledges partial financial support from the BACKUP project (ANR-23-CE40-0018-01). N. Dobigeon acknowledges partial financial support from the Artificial and Natural Intelligence Toulouse Institute (ANITI), funded by the France 2030 program under the grant agreement ANR-23-IACL-0002.

\appendix

\section{Proofs}

\subsection{Proof of \texorpdfstring{\Cref{thm:proj-characterization}}{Theorem 2}\label{app:proof:thm:proj-characterization}}

We establish two elementary identities before proving the theorem.

\begin{lemma}[Projective Jacobian]\label{lem:proj-jac}
Let $\mathcal X\subset\mathbb R^n$ be open, and let $g(\mathbf x)=\mathbf c^\top\mathbf x+d>0$ and
$\boldsymbol\xi(\mathbf x)=(\mathbf A\mathbf x+\mathbf b)/g(\mathbf x)$,
where $\mathbf M=\begin{pmatrix}\mathbf A&\mathbf b\\\mathbf c^\top&d\end{pmatrix}$.
Then
\begin{equation}\label{eq:proj-jac}
 J_{\boldsymbol\xi}(\mathbf x)
 =\frac{\mathbf A-\boldsymbol\xi(\mathbf x)\mathbf c^\top}{g(\mathbf x)},
 \qquad
 \det J_{\boldsymbol\xi}(\mathbf x)=\frac{\det\mathbf M}{g(\mathbf x)^{n+1}}.
\end{equation}
If $\det\mathbf M\ne0$, $\boldsymbol\xi$ is a smooth diffeomorphism
onto its open image.
\end{lemma}

\begin{proof}
The first identity is the quotient rule. Right-multiplication of
$\mathbf M$ by $\bigl(\begin{smallmatrix}\mathbf I&\mathbf x\\0&1\end{smallmatrix}\bigr)$
does not change its determinant, and produces the last column
$\bigl(g(\mathbf x)\boldsymbol\xi(\mathbf x),g(\mathbf x)\bigr)^\top$.
Subtracting $\boldsymbol\xi(\mathbf x)$ times the final row from the
first $n$ rows gives
\[
 \det\mathbf M=g(\mathbf x)
 \det\bigl(\mathbf A-\boldsymbol\xi(\mathbf x)\mathbf c^\top\bigr),
\]
which proves \eqref{eq:proj-jac}. If
$\boldsymbol\xi(\mathbf x)=\boldsymbol\xi(\mathbf y)=\mathbf v$,
then $\mathbf M(\mathbf x,1)^\top=g(\mathbf x)(\mathbf v,1)^\top$
and likewise for $\mathbf y$. Invertibility of $\mathbf M$ implies
that $(\mathbf x,1)^\top$ and $(\mathbf y,1)^\top$ are proportional;
comparison of the last entries yields $\mathbf x=\mathbf y$.
The inverse function theorem and injectivity give the asserted
smooth inverse on the image.
\end{proof}

\begin{lemma}[Perspective--Bregman identity]\label{lem:proj-perspective}
Suppose $\mathcal X,\mathcal T\subset\mathbb R^n$ are open convex sets, $h\in C^2(\mathcal T)$ has positive-definite Hessian,
$g>0$ and $g\boldsymbol\xi$ are affine on $\mathcal X$, and
$\boldsymbol\xi(\mathcal X)\subset\nabla h(\mathcal T)$. Set
$\tau=(\nabla h)^{-1}\circ\boldsymbol\xi$ and
$\rho=g h^*\circ\boldsymbol\xi$. Then $r\in C^2(\mathcal X)$ and
\begin{equation}\label{eq:proj-perspective}
 d_\rho(\mathbf y,\mathbf x)=g(\mathbf y)
 d_h\bigl(\tau(\mathbf x),\tau(\mathbf y)\bigr).
\end{equation}
\end{lemma}

\begin{proof}
Because $\mathcal T$ is convex and $\nabla^2h\succ0$, the gradient
$\nabla h$ is injective with everywhere invertible derivative and hence
a diffeomorphism onto its image. For $\mathbf t\in\mathcal T$, the
supporting-hyperplane inequality implies that the supremum defining
$h^*$ at $\mathbf v=\nabla h(\mathbf t)$ is attained at $\mathbf t$.
Consequently, on $\nabla h(\mathcal T)$,
\[
 h^*(\mathbf v)=\mathbf v^\top\mathbf t-h(\mathbf t),\qquad
 \nabla h^*(\mathbf v)=\mathbf t.
\]
These local conjugacy relations require no essential smoothness.
Write $g\boldsymbol\xi=\mathbf A\mathbf x+\mathbf b$ and
$\nabla g=\mathbf c$. Differentiating $\rho$ and using Fenchel equality,
\begin{align*}
 \nabla \rho
 &=\mathbf c\,h^*(\boldsymbol\xi)
 +gJ_{\boldsymbol\xi}^\top\tau\\
 &=\mathbf c(\boldsymbol\xi^\top\tau-h(\tau))
 +(\mathbf A-\boldsymbol\xi\mathbf c^\top)^\top\tau\\
 &=\mathbf A^\top\tau-\mathbf c\,h(\tau).
\end{align*}
The affine relations imply
$\mathbf A(\mathbf y-\mathbf x)
 =g(\mathbf y)\boldsymbol\xi(\mathbf y)-g(\mathbf x)\boldsymbol\xi(\mathbf x)$
and $\mathbf c^\top(\mathbf y-\mathbf x)=g(\mathbf y)-g(\mathbf x)$.
Substitution into the definition of $d_\rho$ gives
\begin{align*}
 d_\rho(\mathbf y,\mathbf x)
 &=g(\mathbf y)\bigl[h^*(\boldsymbol\xi(\mathbf y))
 -h^*(\boldsymbol\xi(\mathbf x))
 -\tau(\mathbf x)^\top(\boldsymbol\xi(\mathbf y)-\boldsymbol\xi(\mathbf x))\bigr]\\
 &=g(\mathbf y)d_{h^*}(\boldsymbol\xi(\mathbf y),\boldsymbol\xi(\mathbf x))\\
 &=g(\mathbf y)d_h(\tau(\mathbf x),\tau(\mathbf y)),
\end{align*}
where the last equality follows by expanding the two divergences
and applying Fenchel equality at both points.
\end{proof}

We are now ready to prove \Cref{thm:proj-characterization}.\\

\noindent \emph{Necessity ---} Differentiating \eqref{eq:proj-identity} with respect to $\mathbf x$
while holding $\mathbf y$ fixed yields
\begin{equation}\label{eq:proj-diff}
 \nabla^2f(\mathbf x)(\mathbf y-\mathbf x)
 =g(\mathbf y)J_\tau(\mathbf x)^\top
 [\boldsymbol\xi(\mathbf y)-\boldsymbol\xi(\mathbf x)].
\end{equation}
This step does not differentiate $g$. As $J_\tau$ is invertible, throughout the proof, define
$\mathbf K(\mathbf x)=J_\tau(\mathbf x)^{-\top}\nabla^2f(\mathbf x)$.
Then
\begin{equation}\label{eq:proj-separation}
 g(\mathbf y)[\boldsymbol\xi(\mathbf y)-\boldsymbol\xi(\mathbf x)]
 =\mathbf K(\mathbf x)(\mathbf y-\mathbf x).
\end{equation}
Strict convexity of differentiable $h$ on the convex set $\mathcal T$
implies injectivity of $\nabla h$: the strict supporting-hyperplane
inequalities at two distinct points imply strict monotonicity of its
gradient. Hence $\boldsymbol\xi=\nabla h\circ\tau$ is injective.
Choose $\mathbf x_1\ne\mathbf x_2$. Subtracting
\eqref{eq:proj-separation} at these base points gives
\begin{equation}\label{eq:proj-two-points}
 g(\mathbf y)[\boldsymbol\xi(\mathbf x_2)-\boldsymbol\xi(\mathbf x_1)]
 =\mathbf K(\mathbf x_1)(\mathbf y-\mathbf x_1)
 -\mathbf K(\mathbf x_2)(\mathbf y-\mathbf x_2).
\end{equation}
The right-hand side is affine in $\mathbf y$, and the fixed vector
multiplying $g(\mathbf y)$ is nonzero. Taking a nonzero component
shows that $g(\mathbf y)=\mathbf c^\top\mathbf y+d$ is affine.
Returning to \eqref{eq:proj-separation} with $\mathbf x$ fixed shows
that $g(\mathbf y)\boldsymbol\xi(\mathbf y)=\mathbf A\mathbf y+\mathbf b$
is affine as well.

To establish nonsingularity of $\mathbf M$, suppose instead that it is
singular. There is a nonzero $(\mathbf u,v)$ with
$\mathbf u^\top\mathbf A+v\mathbf c^\top=\mathbf0^\top$ and
$\mathbf u^\top\mathbf b+vd=0$. Since $g>0$, necessarily
$\mathbf u\ne\mathbf0$. It follows that
$\mathbf u^\top\boldsymbol\xi(\mathbf x)+v=0$ for all $\mathbf x$.
Thus the image of $\boldsymbol\xi$ lies in a proper affine hyperplane,
contradicting invariance of domain, because $\boldsymbol\xi$ is a
continuous injection on the open set $\mathcal X\subset\mathbb R^n$.
Therefore $\det\mathbf M\ne0$.

Lemma~\ref{lem:proj-jac} now shows that $J_{\boldsymbol\xi}$ is
invertible. From
$J_{\boldsymbol\xi}(\mathbf x)=\nabla^2h(\tau(\mathbf x))J_\tau(\mathbf x)$
and invertibility of $J_\tau$, the Hessian of $h$ is invertible at
every point of $\mathcal T$. It is positive semidefinite by convexity,
and therefore positive definite. Since $g$ is now known to be affine,
we may differentiate \eqref{eq:proj-diff} with respect to $\mathbf y$
and set $\mathbf y=\mathbf x$. The derivative of $g$ multiplies
a vanishing difference on the diagonal, giving
\[
 \nabla^2f(\mathbf x)
 =g(\mathbf x)J_\tau(\mathbf x)^\top
 \nabla^2h(\tau(\mathbf x))J_\tau(\mathbf x)\succ0.
\]
Finally, Lemma~\ref{lem:proj-perspective} applies to
$\rho=g h^*\circ\boldsymbol\xi$ and yields $d_\rho=d_f$.
Differentiating this equality with respect to $\mathbf y$ gives
$\nabla(f-\rho)(\mathbf y)=\nabla(f-\rho)(\mathbf x)$ for all points.
Hence $f-\rho$ is affine on the connected set $\mathcal X$, proving
\eqref{eq:proj-f}.\\

\noindent \emph{Sufficiency ---} 
Conversely, suppose \eqref{eq:proj-g}--\eqref{eq:proj-f} hold under the
initial hypotheses. Lemma~\ref{lem:proj-jac} and the invertibility of
$J_\tau$ imply that $\nabla^2h$ is positive definite. The perspective
identity then gives $d_{g h^*\circ\boldsymbol\xi}=g d_h$; the affine
term in \eqref{eq:proj-f} has zero Bregman divergence. This proves
\eqref{eq:proj-identity}.

\subsection{Proof of \texorpdfstring{\Cref{cor:normalization-uniqueness}}{Corollary 1}}
\label{app:proof:normalization-uniqueness}

\textit{i) Normalization.}
Let $(h_0,\tau_0)$ be an admissible representation associated with the projective matrix
\[
\mathbf M=
\begin{pmatrix}
\mathbf A&\mathbf b\\
\mathbf c^\top&d
\end{pmatrix},
\qquad
\mathbf M^{-\top} =
\begin{pmatrix}
\mathbf P&\mathbf q\\
\mathbf r^\top&s
\end{pmatrix}.
\]
By \Cref{thm:proj-characterization}, there exist $\boldsymbol\zeta\in\mathbb R^n$ and $\kappa\in\mathbb R$ such that
\begin{equation}
\label{eq:normalization-initial}
f(\mathbf x) = g(\mathbf x) h_0^*(\boldsymbol\xi(\mathbf x)) + \boldsymbol\zeta^\top\mathbf x+\kappa
\end{equation}
where
\[
g(\mathbf x)=\mathbf c^\top\mathbf x+d,
\qquad
\boldsymbol\xi(\mathbf x) = \frac{\mathbf A\mathbf x+\mathbf b}{g(\mathbf x)}.
\]
Define
\[
\mathbf a = \mathbf P\boldsymbol\zeta+\mathbf q\kappa, \qquad \kappa_1 = \mathbf r^\top\boldsymbol\zeta+s\kappa,
\]
and consider
\[
\widetilde\tau(\mathbf x) = \tau_0(\mathbf x)+\mathbf a,\qquad \widetilde h(\mathbf t) = h_0(\mathbf t-\mathbf a)-\kappa_1. 
\]
The Fenchel conjugate of $\widetilde h$ satisfies
\begin{equation}
\label{eq:normalization-conjugate}
\widetilde h^*(\mathbf u)
=
h_0^*(\mathbf u)
+
\langle\mathbf a,\mathbf u\rangle
+\kappa_1.
\end{equation}
Since
\[
\begin{pmatrix}
\mathbf a\\
\kappa_1
\end{pmatrix}
=
\mathbf M^{-\top}
\begin{pmatrix}
\boldsymbol\zeta\\
\kappa
\end{pmatrix},
\]
we have
\[
\begin{pmatrix}
\boldsymbol\zeta\\
\kappa
\end{pmatrix}
=
\mathbf M^\top
\begin{pmatrix}
\mathbf a\\
\kappa_1
\end{pmatrix}.
\]
Consequently,
\[
\boldsymbol\zeta
=
\mathbf A^\top\mathbf a
+
\mathbf c\kappa_1,
\qquad
\kappa
=
\mathbf b^\top\mathbf a+d\kappa_1.
\]
It follows that
\begin{align*}
g(\mathbf x)
\bigl(
\langle\mathbf a,\boldsymbol\xi(\mathbf x)\rangle
+\kappa_1
\bigr)
&=
\langle\mathbf a,\mathbf A\mathbf x+\mathbf b\rangle
+\kappa_1(\mathbf c^\top\mathbf x+d)\\
&=
\boldsymbol\zeta^\top\mathbf x+\kappa.
\end{align*}
Combining this identity with \eqref{eq:normalization-initial} and \eqref{eq:normalization-conjugate} yields 
\[
f(\mathbf x) = g(\mathbf x) \widetilde h^*(\boldsymbol\xi(\mathbf x)).
\]
Thus the affine terms in the projective representation have been eliminated. Moreover,
\[
\nabla\widetilde h(\widetilde\tau(\mathbf x)) = \nabla h_0(\tau_0(\mathbf x)) = \boldsymbol\xi(\mathbf x)
\]
so the projective matrix and the map
$\boldsymbol\xi$ remain unchanged. Finally, translations of the argument and
additive constants do not affect Bregman
divergences. Therefore,
\[
d_{\widetilde h} \bigl( \widetilde\tau(\mathbf x), \widetilde\tau(\mathbf y) \bigr) = d_{h_0} \bigl( \tau_0(\mathbf x), \tau_0(\mathbf y) \bigr).
\]
This establishes the normalization property.\\

\noindent \textit{ii) Uniqueness up to affine equivalence.}
Let $(h,\tau)$ and $(h_0,\tau_0)$ be two
admissible representations associated with the same weight $g(\mathbf x)=\mathbf c^\top\mathbf x+d$. By part~\emph{i)}, we may first work with their normalised forms, for which the affine terms $\boldsymbol\zeta$ and $\kappa$ vanish. Let $\mathbf M$ and $\mathbf M_0$ denote
the corresponding projective matrices. Since the two representations have the same weight $g$ on the open set $\mathcal X$, their matrices have  the same last row $(\mathbf c^\top,d)$. It follows that
\[
\mathbf M\mathbf M_0^{-1}
=
\begin{pmatrix}
\mathbf G&\mathbf w\\
\mathbf 0^\top&1
\end{pmatrix}
\]
for some $\mathbf G\in\mathbb R^{n\times n}$ and $\mathbf w\in\mathbb R^n$. Moreover,
\[
\det\mathbf G
=
\frac{\det\mathbf M}{\det\mathbf M_0}
\neq0,
\]
so $\mathbf G$ is invertible. By \Cref{prop:matrix-reconstruction},
the two normalised representations satisfy
\[
\begin{pmatrix}
\tau(\mathbf x)\\
-h(\tau(\mathbf x))
\end{pmatrix}
=
\mathbf M^{-\top}
\begin{pmatrix}
\nabla f(\mathbf x)\\
-F(\mathbf x)
\end{pmatrix}
\]
and
\[
\begin{pmatrix}
\tau_0(\mathbf x)\\
-h_0(\tau_0(\mathbf x))
\end{pmatrix}
=
\mathbf M_0^{-\top}
\begin{pmatrix}
\nabla f(\mathbf x)\\
-F(\mathbf x)
\end{pmatrix}.
\]
Consequently,
\[
\begin{pmatrix}
\tau(\mathbf x)\\
-h(\tau(\mathbf x))
\end{pmatrix}
=
(\mathbf M\mathbf M_0^{-1})^{-\top}
\begin{pmatrix}
\tau_0(\mathbf x)\\
-h_0(\tau_0(\mathbf x))
\end{pmatrix}.
\]
A direct computation gives
\[
(\mathbf M\mathbf M_0^{-1})^{-\top}
=
\begin{pmatrix}
\mathbf G^{-\top}&\mathbf 0\\
-\mathbf w^\top\mathbf G^{-\top}&1
\end{pmatrix}.
\]
Hence
\begin{equation}
\label{eq:uniqueness-coordinate}
\tau(\mathbf x)
=
\mathbf G^{-\top}\tau_0(\mathbf x)
\end{equation}
and
\[
h(\tau(\mathbf x))
=
h_0(\tau_0(\mathbf x))
+
\left\langle
\mathbf w,\tau(\mathbf x)
\right\rangle.
\]
Since
$\tau_0(\mathbf x)=\mathbf G^\top\tau(\mathbf x)$,
we obtain
\begin{equation}
\label{eq:uniqueness-generator}
h(\mathbf t)
=
h_0(\mathbf G^\top\mathbf t)
+
\langle\mathbf w,\mathbf t\rangle.
\end{equation}
Undoing the translations introduced in part \emph{i)} yields the general expressions 
\[
\tau(\mathbf x)
=
\mathbf G^{-\top}\tau_0(\mathbf x)
+\mathbf v
\]
and
\[
h(\mathbf t)
=
h_0\bigl(
\mathbf G^\top(\mathbf t-\mathbf v)
\bigr)
+
\langle\mathbf w,\mathbf t-\mathbf v\rangle
+\kappa_0,
\]
for suitable $\mathbf v,\mathbf w\in\mathbb R^n$ and $\kappa_0\in\mathbb R$. 

Conversely, suppose that $(h,\tau)$ is obtained from $(h_0,\tau_0)$ by such an invertible affine change of coordinates and an affine  modification of the generator. Then $\tau$ remains a diffeomorphism onto an open convex set, and $h$ remains strictly convex. Furthermore, Bregman divergences are invariant under invertible affine coordinate changes accompanied by the corresponding transformation of their generators. The additional affine term in $h$ does not affect its Bregman divergence. Thus 
\[
d_h\bigl(
\tau(\mathbf x),\tau(\mathbf y)
\bigr)
=
d_{h_0}\bigl(
\tau_0(\mathbf x),\tau_0(\mathbf y)
\bigr).
\]
The original dual representation therefore remains valid with the same weight $g$. This proves uniqueness up to affine equivalence.

\subsection{Proof of \texorpdfstring{\Cref{prop:matrix-reconstruction}}{Proposition 4}}\label{app:proof:prop:matrix-reconstruction}
By Fenchel equality, $h^*(\boldsymbol\xi)=\langle\boldsymbol\xi,\tau\rangle-h(\tau)$. The normalized perspective identity gives
\begin{equation}\label{eq:perspective-expanded}
 f(\mathbf x)
 =(\mathbf A\mathbf x+\mathbf b)^\top\tau(\mathbf x)
 -g(\mathbf x)h(\tau(\mathbf x)).
\end{equation}
Differentiating the perspective identity, and using
$\nabla h^*(\boldsymbol\xi(\mathbf x))=\tau(\mathbf x)$,
yields
\begin{equation}\label{eq:first-block}
 \nabla f(\mathbf x)=\mathbf A^\top\tau(\mathbf x)
                    -\mathbf c\,h(\tau(\mathbf x)).
\end{equation}
Subtracting \eqref{eq:perspective-expanded} from
$\mathbf x^\top\nabla f(\mathbf x)$ now gives
\begin{equation}\label{eq:second-block}
 f^*(\nabla f(\mathbf x))
 =d\,h(\tau(\mathbf x))-\mathbf b^\top\tau(\mathbf x).
\end{equation}
Consequently,
\[
 \mathbf M^\top
 \begin{pmatrix}\tau(\mathbf x)\\-h(\tau(\mathbf x))\end{pmatrix}
 =\begin{pmatrix}\nabla f(\mathbf x)\\
                  -f^*(\nabla f(\mathbf x))\end{pmatrix}.
\]
Invertibility of $\mathbf M$ proves \eqref{eq:block-reconstruction};
block multiplication proves \eqref{eq:component-reconstruction}.

\subsection{Proof of \texorpdfstring{\Cref{thm:dual-construction}}{Theorem 3}}
\label{app:proof:dual-construction}

By the regularity and minimality assumptions,
$f=\psi^*$ is closed, proper, and of Legendre type.
Moreover, $f\in C^\infty(\mathcal X)$,
$\nabla^2 f\succ0$ on $\mathcal X$, and
$\Theta=\dom\psi$ is open.
The Legendre correspondence gives mutually inverse
diffeomorphisms
\[
\nabla f:\mathcal X\longrightarrow\Theta, \qquad \nabla\psi:\Theta\longrightarrow\mathcal X. 
\]
Throughout the proof, write
\[
\mathbf M^{-\top} =
\begin{pmatrix}
\mathbf P&\mathbf q\\
\mathbf r^\top&s
\end{pmatrix},
\qquad
F(\mathbf x) = \langle\nabla f(\mathbf x),\mathbf x\rangle -f(\mathbf x).
\]
Then
\[
F(\mathbf x)=\psi(\nabla f(\mathbf x)).
\]
Define
\[
\tau(\mathbf x) = \mathbf P\nabla f(\mathbf x) -\mathbf q F(\mathbf x),  \qquad \eta(\mathbf x) = sF(\mathbf x) -\mathbf r^\top\nabla f(\mathbf x),
\]
and set
\[
\mathcal T=\tau(\mathcal X).
\]
The function $h$ is defined on $\mathbb R^n$ by
\[
h(\mathbf t) = \sup_{\mathbf x\in\mathcal X} \left\{ \langle\boldsymbol\xi(\mathbf x),\mathbf t\rangle -\frac{f(\mathbf x)}{g(\mathbf x)} \right\}.
\]
By the definition of $\mathbf M^{-\top}$,
\begin{equation}
\label{eq:dual-construction-block}
\begin{aligned}
\mathbf A^\top\tau(\mathbf x)
-\mathbf c\,\eta(\mathbf x)
&=\nabla f(\mathbf x),\\
\mathbf b^\top\tau(\mathbf x)
-d\,\eta(\mathbf x)
&=-F(\mathbf x).
\end{aligned}
\end{equation}

\noindent {\emph{i) Identification of $h\circ\tau$.}
For $\mathbf x\in\mathcal X$, let
\[
m_{\mathbf x}(\mathbf t) = \langle\boldsymbol\xi(\mathbf x),\mathbf t\rangle -\frac{f(\mathbf x)}{g(\mathbf x)}.
\]
Thus $h=\sup_{\mathbf x\in\mathcal X}m_{\mathbf x}$. For any $\mathbf x,\mathbf x_0\in\mathcal X$, the identities \eqref{eq:dual-construction-block}, evaluated at $\mathbf x_0$, yield
\begin{equation}
\langle \mathbf A\mathbf x+\mathbf b, \tau(\mathbf x_0) \rangle -\eta(\mathbf x_0)g(\mathbf x)= \langle\nabla f(\mathbf x_0),\mathbf x\rangle -F(\mathbf x_0).
\end{equation}
Using $F(\mathbf x_0) = \langle\nabla f(\mathbf x_0),\mathbf x_0\rangle -f(\mathbf x_0)$, we obtain
\begin{align}
m_{\mathbf x}(\tau(\mathbf x_0))
&=
\eta(\mathbf x_0)
-\frac{
f(\mathbf x)-f(\mathbf x_0)
-\langle\nabla f(\mathbf x_0),
\mathbf x-\mathbf x_0\rangle
}{g(\mathbf x)}
\nonumber\\
&=
\eta(\mathbf x_0)
-\frac{d_f(\mathbf x,\mathbf x_0)}
{g(\mathbf x)}.
\label{eq:dual-construction-key}
\end{align}
Since $f$ is strictly convex on $\mathcal X$ and $g>0$, the right-hand side is bounded above by $\eta(\mathbf x_0)$, with equality  if and only if $\mathbf x=\mathbf x_0$. Consequently, for $\mathbf x_0\in\mathcal X$,
\begin{equation}
\label{eq:dual-construction-value}
h(\tau(\mathbf x_0)) = \eta(\mathbf x_0).
\end{equation}
The supremum defining  $h(\tau(\mathbf x_0))$ is attained uniquely at $\mathbf x=\mathbf x_0$. In particular, $\tau$ is injective: if $\tau(\mathbf x_1)=\tau(\mathbf x_2)$, then both $\mathbf x_1$ and $\mathbf x_2$ maximize the same expression, hence
$\mathbf x_1=\mathbf x_2$. Furthermore, since $m_{\mathbf x_0}\leq h$ and $m_{\mathbf x_0}$ touches $h$ at $\tau(\mathbf x_0)$, 
\begin{equation}
\label{eq:dual-construction-subgradient}
\boldsymbol\xi(\mathbf x_0)
\in\partial h(\tau(\mathbf x_0)).
\end{equation}
Fenchel-Young equality then gives
\begin{equation}
\label{eq:dual-construction-fenchel}
h^*(\boldsymbol\xi(\mathbf x_0)) =  \frac{f(\mathbf x_0)}{g(\mathbf x_0)}.
\end{equation}

\noindent \emph{ii) Diffeomorphism $\tau$.}
Since $\nabla F(\mathbf x)  = \nabla^2f(\mathbf x)\mathbf x$, 
differentiating the definition of $\tau$
gives
\begin{equation}
\label{eq:dual-construction-Jtau}
J_\tau(\mathbf x)
=
(\mathbf P-\mathbf q\mathbf x^\top)
\nabla^2f(\mathbf x).
\end{equation}
The projective matrix identities give
\[
\det(\mathbf P-\mathbf q\mathbf x^\top)
=
\frac{g(\mathbf x)}{\det\mathbf M}.
\]
Therefore
\[
\det J_\tau(\mathbf x)
=
\frac{
g(\mathbf x)\det\nabla^2f(\mathbf x)
}{
\det\mathbf M
}
\neq0.
\]
Thus $\tau$ is a local $C^\infty$ diffeomorphism. Since it is injective by the identification of $h\circ \tau$ in part \emph{i)} above, it is a $C^\infty$ diffeomorphism from $\mathcal X$ onto the nonempty open set $\mathcal T=\tau(\mathcal X)$.\\

\noindent \emph{iii) Convexity of $\mathcal T$.} As a supremum of affine functions, $h$ is convex on $\mathbb R^n$.
Consequently, $\dom h$ is convex. By \eqref{eq:dual-construction-value}, $\mathcal T\subseteq\dom h$. We prove the reverse inclusion. Let $\mathbf t\in\dom h$ and set
\[
\sigma=h(\mathbf t),
\qquad
\btheta
=
\mathbf A^\top\mathbf t-\mathbf c\sigma.
\]
By the definition of $h$, for every
$\mathbf x\in\mathcal X$,
\[
\langle
\boldsymbol\xi(\mathbf x),\mathbf t
\rangle
-\frac{f(\mathbf x)}{g(\mathbf x)}
\leq\sigma.
\]
Multiplying by $g(\mathbf x)>0$ gives
\[
\langle\btheta,\mathbf x\rangle
-f(\mathbf x)
\leq
d\sigma-\mathbf b^\top\mathbf t,\quad \text{for all}\ \bx\in\X.
\]
That is, since $f$ is closed and $\X=\inte\dom f$ \cite[Thm.~7.5]{rockafellar1997convex}, and using $f^*=\psi$,
\begin{equation}
\label{eq:dual-construction-ineq}
\psi(\btheta)  \leq d\sigma-\mathbf b^\top\mathbf t 
\end{equation}
In particular, $\btheta\in\Theta$. We claim that equality holds in \eqref{eq:dual-construction-ineq}. Otherwise, since $\Theta$ is open and $\psi$ is continuous on $\Theta$, there exists $\varepsilon>0$ sufficiently small such that
\[
\psi(\btheta+\varepsilon\mathbf c) \leq d(\sigma-\varepsilon) -\mathbf b^\top\mathbf t.
\]
By conjugacy, this implies
\[
\langle
\btheta+\varepsilon\mathbf c,
\mathbf x
\rangle-f(\mathbf x)
\leq
d(\sigma-\varepsilon)
-\mathbf b^\top\mathbf t
\]
for every $\mathbf x\in\mathcal X$. Rearranging and dividing by
$g(\mathbf x)>0$ gives
\[
m_{\mathbf x}(\mathbf t)
\leq\sigma-\varepsilon.
\]
Taking the supremum over $\mathbf x$
would imply
\[
h(\mathbf t)\leq\sigma-\varepsilon,
\]
which contradicts $\sigma=h(\mathbf t)$.
Hence
\begin{equation}
\label{eq:dual-construction-implicit-equality}
\psi(\btheta)
=
d\sigma-\mathbf b^\top\mathbf t.
\end{equation}
Set $\mathbf x=\nabla\psi(\btheta)
\in\mathcal X$. 
By Legendre duality, one has $\nabla f(\mathbf x)=\btheta$ and $F(\mathbf x)=\psi(\btheta)$. It follows that
\[
\mathbf M^\top
\begin{pmatrix}
\mathbf t\\-\sigma
\end{pmatrix}
=
\begin{pmatrix}
\nabla f(\mathbf x)\\-F(\mathbf x)
\end{pmatrix}.
\]
Multiplying by $\mathbf M^{-\top}$ gives
\[
\mathbf t=\tau(\mathbf x), \qquad \sigma=\eta(\mathbf x).
\]
Thus $\mathbf t\in\mathcal T$, proving
\begin{equation}
\label{eq:dual-construction-domain}
\dom h=\mathcal T.
\end{equation}
Since $\dom h$ is convex and $\mathcal T$ is open by part (b), $\mathcal T$ is a nonempty open convex set.\\

\noindent \emph{iv) Regularity and strict convexity of $h$.}
By part \emph{i)}, one has $h=\eta\circ\tau^{-1}$ on $\mathcal T$. Hence $h$ is differentiable on $\mathcal T$. Since a differentiable convex function has a singleton subdifferential,
Eq. \eqref{eq:dual-construction-subgradient} implies 
\begin{equation}
\label{eq:dual-construction-gradient}
\nabla h(\tau(\mathbf x)) = \boldsymbol\xi(\mathbf x).
\end{equation}
Therefore $\nabla h = \boldsymbol\xi\circ\tau^{-1}$ on $\mathcal T$, which shows that $h\in C^2(\mathcal T)$. Differentiating \eqref{eq:dual-construction-gradient}
gives
\[
\nabla^2h(\tau(\mathbf x)) J_\tau(\mathbf x) = J_{\boldsymbol\xi}(\mathbf x).
\]
The projective identities yield
\[
J_{\boldsymbol\xi}(\mathbf x)
=
\frac1{g(\mathbf x)}
(\mathbf P-\mathbf q\mathbf x^\top)^{-\top}.
\]
Combining this with
\eqref{eq:dual-construction-Jtau},
we obtain
\begin{align}
\nabla^2h(\tau(\mathbf x))
&=
\frac1{g(\mathbf x)}
(\mathbf P-\mathbf q\mathbf x^\top)^{-\top}
[\nabla^2f(\mathbf x)]^{-1}
(\mathbf P-\mathbf q\mathbf x^\top)^{-1}.
\label{eq:dual-construction-Hessian}
\end{align}
Since $g(\mathbf x)>0$, $\nabla^2f(\mathbf x)\succ0$, and $\mathbf P-\mathbf q\mathbf x^\top$ is invertible,
\[
\nabla^2h(\mathbf t)\succ0,
\qquad
\mathbf t\in\mathcal T.
\]
Thus $h$ is strictly convex on the convex set $\mathcal T$.\\

\noindent \emph{v) Conformal identity.}
By parts \emph{i)}--\emph{iv)}, the pair $(h,\tau)$
satisfies the regularity and convexity assumptions of \Cref{thm:proj-characterization}. Moreover, Eq. \eqref{eq:dual-construction-gradient} gives
\[
\boldsymbol\xi=\nabla h\circ\tau,
\]
while Eq. \eqref{eq:dual-construction-fenchel} gives
\[
f(\mathbf x) = g(\mathbf x) h^*(\boldsymbol\xi(\mathbf x)).
\]
Thus the projective representation holds with no additional affine term. Applying \Cref{thm:proj-characterization}, we obtain, for $\mathbf x,\mathbf y\in\mathcal X.$
\begin{equation}
\label{eq:dual-construction-conformal-proof}
d_f(\mathbf y,\mathbf x) = g(\mathbf y) d_h\bigl( \tau(\mathbf x),\tau(\mathbf y) \bigr).
\end{equation}

\noindent \emph{vi) Legendre property.}
Since $h$ is a supremum of continuous affine functions, it is convex and lower semicontinuous on $\mathbb R^n$. By part \emph{i)}, it is finite on the nonempty set $\mathcal T$, and is
therefore proper. 

By part \emph{iii)},
\[
\dom h = \inte(\dom h) = \mathcal T.
\]
By part \emph{iv)}, $h$ is differentiable and strictly convex on $\mathcal T$. Since
\[
\dom\partial h \subseteq\dom h = \mathcal T,
\]
strict convexity on $\mathcal T$ implies essential strict convexity.

We now verify essential smoothness. Let $(\mathbf t_k)$ be a sequence in $\mathcal T$ converging to $\bar{\mathbf t}\in\partial\mathcal T$. Since $\dom h=\mathcal T$ is open,
\[
h(\bar{\mathbf t})=+\infty.
\]
By lower semicontinuity,
\[
h(\mathbf t_k)\longrightarrow+\infty.
\]
Suppose, by contradiction, that $\|\nabla h(\mathbf t_k)\|$ does not tend to infinity. Then there exists a subsequence, still denoted $(\mathbf t_k)$, along which the gradients remain
bounded. Fix $\mathbf t_0\in\mathcal T$. The convexity inequality gives
\[
h(\mathbf t_0) \geq h(\mathbf t_k) + \langle \nabla h(\mathbf t_k), \mathbf t_0-\mathbf t_k \rangle.
\]
Hence
\[
h(\mathbf t_k) \leq h(\mathbf t_0) + \|\nabla h(\mathbf t_k)\|\, \|\mathbf t_k-\mathbf t_0\|.
\]
The right-hand side remains bounded along the subsequence, contradicting $h(\mathbf t_k)\to+\infty$. Therefore
\[
\|\nabla h(\mathbf t_k)\| \longrightarrow+\infty. 
\]
Thus $h$ is essentially smooth. Together with essential strict convexity, this proves that $h$ is of Legendre type.\\

\noindent \emph{vii) Smoothness.}
Since $f\in C^\infty(\mathcal X)$, both $\nabla f$ and 
\[
F(\mathbf x)  = \langle\nabla f(\mathbf x),\mathbf x\rangle -f(\mathbf x)
\]
are $C^\infty$ on $\mathcal X$. Consequently, $\tau$ and $\eta$ are $C^\infty$. By part \emph{ii)}, $J_\tau$ is invertible everywhere on $\mathcal X$, and $\tau^{-1}$ is $C^\infty$ on
$\mathcal T$. Finally, by part \emph{i)},
\[
h=\eta\circ\tau^{-1} \quad\text{on }\mathcal T, \]
so
\[
h\in C^\infty(\mathcal T).
\]

\noindent \emph{viii) Uniqueness.} If $(h_1,\tau_1)$ is another such pair, \Cref{prop:matrix-reconstruction} gives $\tau_1=\tau$ and
$h_1\circ\tau=h\circ\tau$, so $h_1=h$ on $\T$, and both equal $+\infty$ outside $\dom h_1=\dom h=\T$.

\subsection{Proof of \texorpdfstring{\Cref{cor:explicit-construction}}{Corollary 2}}
\label{app:proof:explicit-construction}
By \Cref{thm:dual-construction}, $\tau$ is a bijection from $\X$ onto $\T$, $h$ is a closed proper convex function of
Legendre type with $\dom h=\T$, and $(h,\tau)$ satisfies the conditions of \Cref{thm:proj-characterization}, so that
\Cref{prop:matrix-reconstruction} applies. In particular, it gives $\tau=\mathbf P\nabla f-\mathbf q F$ and
\begin{equation}\label{eq:rowsh}
  \bA^{\top}\tau-\bc\,h\circ\tau=\nabla f,\qquad \bb^{\top}\tau-d\,h\circ\tau=-F,
\end{equation}
the two block rows of $$\bM^{\top}(\tau;-h\circ\tau)=(\nabla f;-F)$$

\noindent \emph{i) Expression of $h^*(u)$ .} 
Since $h$ is of Legendre type, \citet[Thm.~26.5]{rockafellar1997convex} shows that $h^*$ is of Legendre type, that $\nabla h$ is a bijection from $\T$ onto $\inte(\dom h^*)$, and that $\nabla h^*=(\nabla h)^{-1}$. Since $\tau$ is a bijection from $\X$ onto $\T$ and $\nabla h\circ\tau=\boldsymbol\xi$, $\boldsymbol\xi$ is a bijection from $\X$ onto $\nabla h(\T)=\inte(\dom h^*)$, and $\nabla h^*=\tau\circ{\boldsymbol\xi}^{-1}$ there. Finally, let $\bu={\boldsymbol\xi}(\bx)$. By~\eqref{eq:projective-data}, 
$$g(\bx)\,\begin{pmatrix}\bu\\1\end{pmatrix}=\bM \begin{pmatrix}\bx \\ 1\end{pmatrix}$$
and $\bM^{-1}=(\bM^{-\top})^{\top}$ has block rows $(\mathbf P^{\top}\ \mathbf r)$ and $(\mathbf q^{\top}\ s)$, so 
$$\begin{pmatrix}\bx\\1\end{pmatrix}=g(\bx)\,\begin{pmatrix}\mathbf P^{\top}\bu+\mathbf r\\\mathbf q^{\top}\bu+s\end{pmatrix}.$$ 
The inverse projective identities give
\[
\mathbf q^\top\boldsymbol\xi(\mathbf x)+s = \frac1{g(\mathbf x)},
\]
and
\[
\mathbf x
=
\frac{ \mathbf P^\top\boldsymbol\xi(\mathbf x) +\mathbf r }{\mathbf q^\top\boldsymbol\xi(\mathbf x)+s}.
\]
Substituting into \eqref{eq:proj-f}, we obtain, for every $\boldsymbol u\in\boldsymbol\xi(\mathcal X)$,
\begin{equation}
\label{eq:dual-construction-hstar}
h^*(\boldsymbol u) =
(\mathbf q^\top\boldsymbol u+s)
f\left(
\frac{ \mathbf P^\top\boldsymbol u+\mathbf r }{ \mathbf q^\top\boldsymbol u+s }
\right).
\end{equation}
which is ~\eqref{eq:construction-conjugate}.\\

\noindent \emph{ii) Expression of $h(t)$.} Since $h$ is closed, proper and convex, $h=h^{**}$ \cite[Thm.~12.2]{rockafellar1997convex}, that is, for $\bt\in\R^n$,
$$h(\bt)=\sup_{\bu}\bigl(\ip{\bu}{\bt}-h^*(\bu)\bigr).$$
The function $\bu\mapsto h^*(\bu)-\ip{\bu}{\bt}$ is closed, proper and convex, so its values on $\dom h^*$ are limits of its values on $\inte(\dom h^*)$ along segments \cite[Thm.~7.5]{rockafellar1997convex}, and the supremum may be taken over $\inte(\dom h^*)=\mathbf \xi(\X)$. By part~\emph{i)},
\[
  h(\bt)=\sup_{\bx\in\X}\bigl(\ip{\mathbf\xi(\bx)}{\bt}-h^*(\mathbf \xi(\bx))\bigr)=\sup_{\bx\in\X}\Bigl(\ip{\mathbf \xi(\bx)}{\bt}-\frac{f(\bx)}{g(\bx)}\Bigr),
\]
which is~\eqref{eq:construction-formulas}. In particular, $h$ is a supremum of affine functions, namely the conjugate of the restriction of $h^*$ to $\mathbf\xi(\X)$, given there by~\eqref{eq:construction-conjugate}.\\

\noindent \emph{iii) Implicit equation.}
Let $\mathbf t\in\mathcal T$.
There exists a unique
$\mathbf x\in\mathcal X$ such that $\mathbf t=\tau(\mathbf x).$ 
The block identities
\eqref{eq:rowsh}
therefore give
\[
\mathbf A^\top\mathbf t
-\mathbf c h(\mathbf t)
=
\nabla f(\mathbf x)
\]
and
\[
d h(\mathbf t)-\mathbf b^\top\mathbf t
=
F(\mathbf x)
=
\psi(\nabla f(\mathbf x)).
\]
Consequently,
\[
\psi\bigl(
\mathbf A^\top\mathbf t
-\mathbf c h(\mathbf t)
\bigr)
=
d h(\mathbf t)-\mathbf b^\top\mathbf t.
\]
Thus $\sigma=h(\mathbf t)$ solves
\begin{equation}
\label{eq:dual-construction-implicit-proof}
\psi(\mathbf A^\top\mathbf t-\mathbf c\sigma)
=
d\sigma-\mathbf b^\top\mathbf t.
\end{equation}
Conversely, suppose $\sigma\in\mathbb R$ satisfies \eqref{eq:dual-construction-implicit-proof}. Since the right-hand side is finite,
\[
\btheta = \mathbf A^\top\mathbf t-\mathbf c\sigma \in\Theta.
\]
Set $\mathbf x' = \nabla\psi(\btheta) \in\mathcal X$. Then
\[
\nabla f(\mathbf x')
=
\mathbf A^\top\mathbf t-\mathbf c\sigma
\]
and
\[
F(\mathbf x')
=
\psi(\btheta)
=
d\sigma-\mathbf b^\top\mathbf t.
\]
Therefore
\[
\mathbf M^\top
\begin{pmatrix}
\mathbf t\\-\sigma
\end{pmatrix}
=
\begin{pmatrix}
\nabla f(\mathbf x')\\-F(\mathbf x')
\end{pmatrix}.
\]
Multiplication by $\mathbf M^{-\top}$
yields
\[
\mathbf t=\tau(\mathbf x'),
\qquad
-\sigma=-h(\tau(\mathbf x')).
\]
Thus,
\[
\sigma
=
h(\tau(\mathbf x'))
=
h(\mathbf t).
\]
This proves uniqueness.

\subsection{Proof of \texorpdfstring{\Cref{cor:dual-readings}}{Corollary 3}}
\label{app:proof:dual-readings}
Let $\mathbf t=\tau(\mathbf x)$ and
$\btheta=\nabla f(\mathbf x)$.
The reconstruction identities give
\[
\btheta
=
\mathbf A^\top\mathbf t
-\mathbf c\,h(\mathbf t),
\qquad
f^*(\btheta)
=
d\,h(\mathbf t)-\mathbf b^\top\mathbf t.
\]
Substituting into the natural exponential-family
log-density yields
\begin{align*}
\log p(\mathbf y;\btheta)
&=
\langle\btheta,\mathbf y\rangle
-f^*(\btheta)
+\log p_0(\mathbf y)\\
&=
\langle\mathbf t,\mathbf A\mathbf y+\mathbf b\rangle
-(\mathbf c^\top\mathbf y+d)h(\mathbf t)
+\log p_0(\mathbf y).
\end{align*}
This proves \eqref{eq:dual-logdensity-exponential}. The normalization statement follows from \eqref{eq:construction-implicit}. Finally, for $\mathbf y\in\mathcal X$, combining \eqref{eq:likelihood} with \eqref{eq:construction-conformal} gives \eqref{eq:dual-logdensity-bregman}.

\subsection{Proof of \texorpdfstring{\Cref{cor:scaled-representations}}{Corollary 4}}
\label{app:proof:scaled-representations}
\noindent \emph{i)} For the $\beta$-scaled model, choose
\[
\mathbf M=
\begin{pmatrix}
I_n&0\\
0&1/\beta
\end{pmatrix},
\qquad
\mathbf M^{-\top}=
\begin{pmatrix}
I_n&0\\
0&\beta
\end{pmatrix}.
\]
The reconstruction formulas give $\tau=\nabla f$ and $h=\beta f^*$.\\

\noindent \emph{ii)} For the $\alpha$-scaled model, choose
\[
\mathbf M=
\begin{pmatrix}
0&-1\\
1/\alpha&0
\end{pmatrix},
\qquad
\mathbf M^{-\top}=
\begin{pmatrix}
0&-1\\
\alpha&0
\end{pmatrix}.
\]
The reconstruction formulas give
\[
\tau(x) = xf'(x)-f(x) = f^*(f'(x)), \qquad h(\tau(x)) = -\alpha f'(x).
\]
Since $f''(x)>0$ and $x>0$,
\[
\tau'(x)=xf''(x)>0.
\]
Furthermore, $f^*$ is strictly increasing on $f'(\mathcal X)= \inte(\dom f^*) = \dom f^*$, because
\[
(f^*)'(f'(x))=x>0.
\]
Thus $f^*$ is invertible and yields the stated expression
for $h$.\\

The two log-density representations follow from
\Cref{cor:dual-readings}.
The Legendre properties follow from
\Cref{thm:dual-construction}.

\subsection{Proof of \texorpdfstring{\Cref{prop:weighted:bregman:identity}}{Proposition 5}}
\label{app:weighted:bregman:identity}

We start by deriving the general form of the log-posterior in the following lemma.

\begin{lemma}\label{lem:derivation:log:posterior}
Consider the Bayesian model defined by the log-likelihood
\begin{equation}\label{log:likel:banerjee}
\ell(\mathbf y;\mathbf x) = -d_f(\mathbf y,\mathbf x)+\phi_f(\mathbf y)
\end{equation}
and the log-prior
\begin{equation}
\log p(\mathbf x|\alpha,\beta) = -\beta \,d_f(\frac{\alpha}{\beta},\mathbf x)+\varphi_f(\alpha,\beta).
\end{equation}
Then the log-posterior can be written as
\begin{align}
\log p(\mathbf x|\mathbf y,\alpha,\beta)
&=
-d_f(\mathbf y,\mathbf x)
-\beta\, d_f(\frac \alpha \beta,\mathbf x)
+\psi_f(\mathbf y,\alpha,\beta),
\label{eq:posterior}
\end{align}
where the term
$\psi_f(\mathbf y,\alpha,\beta)$
does not depend on the parameter $\mathbf x$
\end{lemma}
\begin{proof}
    We want to  show that combining the log prior  
\begin{equation}
\log p(\mathbf x|\alpha,\beta)
=
-\beta \,d_f(\frac \alpha \beta,\mathbf x)+\varphi_f(\alpha,\beta)
\end{equation}
 with the log-likelihood \eqref{log:likel:banerjee}
gives  the log-posterior
\begin{align}
\log p(\mathbf x|\mathbf y,\alpha,\beta)
&=
\ell(\mathbf y;\mathbf x)
+\log p(\mathbf x|\alpha,\beta)
\nonumber\\
&=
-d_f(\mathbf y,\mathbf x)
-\beta\, d_f(\frac \alpha \beta,\mathbf x)
+\psi_f(\mathbf{y},\alpha,\beta)
\end{align}
where the term
$\psi_f(\mathbf{y},\alpha,\beta)$
does not depend on the parameter $\mathbf x$. Starting from the likelihood and prior,
\begin{align*}
p(\mathbf{y}|\mathbf{x})
&=
\exp\left(
-d_f(\mathbf{y},\mathbf{x})+\phi_f(\mathbf{y})
\right),\\
p(\mathbf{x}|\alpha,\beta)
&=
\exp\left(
-\beta\, d_f(\frac \alpha \beta,\mathbf{x})
+\varphi_f(\alpha,\beta)
\right),
\end{align*}
applying Bayes' rule gives
\begin{align*}
p(\mathbf{x}|\mathbf{y},\alpha,\beta)
&=
\frac{
p(\mathbf{y}|\mathbf{x})
p(\mathbf{x}|\alpha,\beta)
}{
p(\mathbf{y}|\alpha,\beta)
}\\
&=
\frac{
\exp\left(
-d_f(\mathbf{y},\mathbf{x})
-\beta\, d_f(\frac \alpha \beta,\mathbf{x})
+\phi_f(\mathbf{y})
+\varphi_f(\alpha,\beta)
\right)
}{
p(\mathbf{y}|\alpha,\beta)
}.
\end{align*}
Taking the logarithm yields
\begin{align*}
\log p(\mathbf{x}|\mathbf{y},\alpha,\beta)
&=
-d_f(\mathbf{y},\mathbf{x})
-\beta\, d_f(\frac \alpha \beta,\mathbf{x})
+\phi_f(\mathbf{y})
+\varphi_f(\alpha,\beta)
-\log p(\mathbf{y}|\alpha,\beta).
\end{align*}
Hence,
\begin{equation*}
\log p(\mathbf{x}|\mathbf{y},\alpha,\beta)
=
-d_f(\mathbf{y},\mathbf{x})
-\beta\, d_f(\frac \alpha \beta,\mathbf{x})
+\psi_f(\mathbf{y},\alpha,\beta),
\end{equation*}
where
\begin{equation*}
\psi_f(\mathbf{y},\alpha,\beta)
=
\phi_f(\mathbf{y})
+\varphi_f(\alpha,\beta)
-\log p(\mathbf{y}|\alpha,\beta).
\end{equation*}
Furthermore, the marginal likelihood is
\begin{align*}
p(\mathbf{y}|\alpha,\beta)
&=
\int_{\mathcal{X}}
p(\mathbf{y}|\mathbf{x})
p(\mathbf{x}|\alpha,\beta)
\,d\mathbf{x}\\
&=
\exp\left(
\phi_f(\mathbf{y})+\varphi_f(\alpha,\beta)
\right)
\int_{\mathcal{X}}
\exp\left(
-d_f(\mathbf{y},\mathbf{x})
-\beta\, d_f(\frac \alpha \beta,\mathbf{x})
\right)
d\mathbf{x}.
\end{align*}
Substituting this expression into $\psi_f(\mathbf{y},\alpha,\beta)$ gives
\begin{equation*}
\psi_f(\mathbf{y},\alpha,\beta)
=
-\log
\int_{\mathcal{X}}
\exp\left(
-d_f(\mathbf{y},\mathbf{x})
-\beta\, d_f(\frac \alpha \beta,\mathbf{x})
\right)
d\mathbf{x}.
\end{equation*}
\end{proof}

Lemma \ref{lem:derivation:log:posterior} shows that the log-posterior can be written as a weighted sum of Bregman divergences with a common generator. We can now show that it can be rewritten as a single Bregman divergence. Indeed, using the definition of the Bregman divergence,
\begin{align*}
d_f(\mathbf{y},\mathbf{x})
+
\beta\, d_f(\frac \alpha \beta,\mathbf{x})
&=
f(\mathbf{y})-f(\mathbf{x})
-\left\langle
\mathbf{y}-\mathbf{x},
\nabla f(\mathbf{x})
\right\rangle
+\beta
\left[
f(\frac \alpha \beta)-f(\mathbf{x})
-\left\langle
\frac \alpha \beta-\mathbf{x},
\nabla f(\mathbf{x})
\right\rangle
\right]
\nonumber\\
&=
f(\mathbf{y})+\beta f(\frac \alpha \beta)
-(1+\beta)f(\mathbf{x})
-\left\langle
\mathbf{y}+\alpha
-(1+\beta)\mathbf{x},
\nabla f(\mathbf{x})
\right\rangle
\nonumber
\end{align*}
Since $\mathcal X$  is convex and
$\mathbf{y},\frac \alpha \beta\in\dom(f)$, their weighted average
\begin{equation*}
 \frac{\mathbf y}{1+\beta} + \frac \beta {1+\beta} \left(\frac \alpha \beta\right) = 
\frac{\mathbf{y}+\alpha}{1+\beta} : = \frac{\alpha^+}{\beta^+}
\end{equation*}
also belongs to $\dom(f)$. Thus
\begin{align*}
d_f(\mathbf{y},\mathbf{x})
+
\beta d_f\left(\frac{\alpha}{\beta},\mathbf{x}\right)
&=
\beta^+
\left[
f(\frac{\alpha^+}{\beta^+})-f(\mathbf{x})
-\left\langle
\frac{\alpha^+}{\beta^+}-\mathbf{x},
\nabla f(\mathbf{x})
\right\rangle
\right]
+
f(\mathbf{y})+\beta f(\frac{\alpha}{\beta})
-\beta^+\,f(\frac{\alpha^+}{\beta^+})
\nonumber\\
&=
\beta^+\,d_f(\frac{\alpha^+}{\beta^+},\mathbf{x})
+f(\mathbf{y})+\beta f(\frac{\alpha}{\beta})
-\beta^+\,f(\frac{\alpha^+}{\beta^+}).
\end{align*}
Finally, plugging this expansion into the expression of the log-posterior 
\begin{align}
\log p(\mathbf x|\mathbf y,\alpha,\beta) = 
-d_f(\mathbf y,\mathbf x)
-\beta\, d_f(\frac \alpha \beta,\mathbf x)
+\psi_f(\mathbf{y},\alpha,\beta)
\end{align}
exhibited in Lemma  \ref{lem:derivation:log:posterior}, then it follows
\begin{equation*}
\log p(\mathbf{x}|\mathbf{y},\alpha,\beta)
=
-\beta^+\,
d_f
\left(
\frac{\alpha^+}{\beta^+},
\mathbf x
\right)
+
\Psi_f(\mathbf y,\alpha,\beta)
\end{equation*}
where
\begin{equation*}
\Psi_f(\mathbf{y},\alpha,\beta)
= \psi_f(\mathbf{y},\alpha,\beta)
-f(\mathbf{y})
-\beta f(\frac{\alpha}{\beta})
+\beta^+\,f(\frac{\alpha^+}{\beta^+})
\end{equation*}
is independent of
$\mathbf x$.  

\subsection{Proof of \texorpdfstring{\Cref{prop:beta-relative-smoothness}}{Proposition 6} } \label{app:proof_beta-relative-smoothness} 
In natural coordinates, the negative log-density associated with a distribution belonging to the regular exponential family writes (see Definition \ref{def:natural_param_expo})
\begin{equation}
\lambda(\mathbf y;\mathbf t)
=
\psi(\mathbf t)
-
\langle \mathbf t,\mathbf y\rangle
- \log p_0(\mathbf{y}).
\end{equation}
According to Proposition \ref{theo:derivation:banerjee:term}, this log-density admits a Bregman representation once $\psi=f^*$ with $\mathbf{t} = \nabla f(\mathbf{x})$. Consider the Legendre function $h=\beta f^*$ and dual coordinates $\mathbf{t}$ associated with the $\beta$-scaled dual Bregman representation identified in \Cref{cor:beta_existence}. Then 
\begin{align}
\zeta(\mathbf t)
&=
Lh(\mathbf t)-\lambda(\mathbf y;\mathbf t)
\\
&=
L\beta f^*(\mathbf t)
-
f^*(\mathbf t)
+
\langle\mathbf t,\mathbf y\rangle
+
\log p_0(\mathbf{y})
\\
&=
(L\beta-1)f^*(\mathbf t)
+
\langle\mathbf t,\mathbf y\rangle
+
\log p_0(\mathbf{y}).
\end{align}
Since $f^*$ is convex and the remaining terms are affine in
$\mathbf t$, the function $\zeta$ is convex on $\mathcal T$ whenever
$ L\beta-1\geq0$. 
Hence, $\lambda(\mathbf y;\cdot)$ is $L$-smooth relative to $h$ for every $L\geq1/\beta$.

\subsection{Proof of \texorpdfstring{\Cref{prop:general-tau-tilde}}{Proposition 7}} \label{app:proof_general-tau-tilde}
In natural coordinates, the negative log-density associated with a distribution belonging to the regular exponential family writes (see Definition \ref{def:natural_param_expo})
\begin{equation} \label{eq:neglogdensity_alpha_natural}
\lambda( y; t)
=
\psi( t)
-
 t y
- \log p_0({y}).
\end{equation}
According to Proposition \ref{theo:derivation:banerjee:term}, this log-density admits a Bregman representation once $\psi=f^*$ with ${t} =  f'({x})$. Hence, the negative log-likelihood writes
\begin{equation}
-\ell( y; x)
=
f^*(f'({x}))
-
 f'({x})y
- \log p_0({y}).
\end{equation}
and
\[
-\frac{\partial}{\partial x}\ell( y; x)
=
x f''(x)-y f''(x)
=
(x-y)f''(x).
\]
Moreover, the dual coordinates induced by the $\alpha$-scaled Bregman representation identified in \Cref{cor:alpha_existence} are defined as ${t}={\tau}(x)$ with
\[
{\tau}(x) = f^*(f'(x)) = xf'(x)-f(x),
\]
which yields
\[
\frac{d {\tau}(x)}{dx}
=
x f''(x),
\]
and
\begin{equation}\label{eq:derivative_x_tildet}
    \frac{dx}{d{t}}=\frac{1}{x f''(x)}.
\end{equation}
By defining ${\lambda}(y;{t}) = -\ell(y;{\tau}^{-1}({t}))$, it follows that
\[
{\lambda}'(y;{t})
=
-\frac{\partial}{\partial x}\ell(y;x)
\frac{dx}{d{t}}
=
\frac{x-y}{x}
=
1-\frac{y}{x}.
\]
Differentiating once more gives
\begin{equation}\label{eq:2derivative_tilde_lambda}
    {\lambda}''(y;{t})
=
\frac{y}{x^3 f''(x)}.
\end{equation}
By setting ${h}({t}) = - \alpha f'({\tau}^{-1}({t}))$ and ${t} = {\tau}(x)$ as in \Cref{cor:alpha_existence}, one has
\begin{align*}
    {h}'({t}) &= -\alpha \frac{d f'(x)}{dx} \frac{dx}{d{t}}
\end{align*}
and, plugging \eqref{eq:derivative_x_tildet}, it yields
\begin{align*}
    {h}'({t}) &= -\frac{\alpha}{x}
\end{align*}
and
\begin{align}\label{eq:2derivative_tilde_h}
    h''({t}) &= \frac{\alpha}{x^3 f''(x)}.
\end{align}
Consider the function
\begin{equation}
    {\zeta}({t}) = L {h}({t})-{\lambda}(y;{t}).
\end{equation}
Then, using \eqref{eq:2derivative_tilde_lambda} and \eqref{eq:2derivative_tilde_h}
\begin{equation}
    {\zeta}''({t}) =\left( L - \frac{y}{\alpha}\right) {h}''({t}).
\end{equation}
Since ${h}$ is convex,
${h}''({t})\geq0$.
Therefore,
\begin{itemize}
    \item if $y\neq 0$, ${\zeta}$ is convex whenever
$
L\geq\frac{y}{\alpha}.
$
Thus, the function  ${\lambda}(y;\cdot)$ is $L$-smooth relative to
${h}$ for every $L\geq y/\alpha$.
 \item if $y= 0$, for any $\epsilon >0$, the function ${\lambda}(y;\cdot)$ is $L$-smooth relative to
${h}$ for every $L\geq \epsilon$.
\end{itemize}

\subsection{Proof of \texorpdfstring{\Cref{prop:multidimensional-separable}}{Corollary 10}} \label{app:proof_multidimensional-separable}
 Let define \[ {\mathbf{t}} = {\boldsymbol\tau}(\mathbf x). \] 
 For each component, 
 \[ {t}_i = {\tau}_i(x_i) = x_i f_i'(x_i)-f_i(x_i), \] 
 so that 
 \[ \frac{d{t}_i}{dx_i} = x_i f_i''(x_i), \qquad \frac{dx_i}{d{t}_i} = \frac{1}{x_i f_i''(x_i)}. \] 
 Since the model is separable, the negative log-likelihood can be written as 
 \[ \bar\ell(\mathbf y;\mathbf x) = \sum_{i=1}^{n} \left[ f_i^*(f_i'(x_i)) - y_i f_i'(x_i) \right] - \phi_f(\mathbf{y}), \]
 where $\phi_f(\mathbf{y})$ does not depend on $\mathbf x$. Hence, 
 \[ \frac{\partial}{\partial x_i} \bar\ell(\mathbf y;\mathbf x) = (x_i-y_i)f_i''(x_i). \]
 Therefore, \[ \frac{\partial {\lambda}} {\partial {t}_i} = \frac{\partial \bar\ell}{\partial x_i} \frac{dx_i}{d{t}_i} = \frac{(x_i-y_i)f_i''(x_i)} {x_i f_i''(x_i)} = 1-\frac{y_i}{x_i}. \] 
 Differentiating once more with respect to ${t}_i$ gives 
 \[ \frac{\partial^2 {\lambda}} {\partial {t}_i^2} = \frac{y_i}{x_i^2} \frac{dx_i}{d{t}_i} = \frac{y_i}{x_i^3 f_i''(x_i)}. \]
 On the other hand, by the one-dimensional $\alpha$-scaled construction, \[ {h}_i'({t}_i) = -\frac{\alpha_i}{x_i}, \] 
 and therefore \[ {h}_i''({t}_i) = \frac{\alpha_i}{x_i^2} \frac{dx_i}{d{t}_i} = \frac{\alpha_i}{x_i^3 f_i''(x_i)}. \] Hence, \[ \frac{\partial^2 {\lambda}} {\partial {t}_i^2} = \frac{y_i}{\alpha_i} {h}_i''({t}_i). \] 
 Since ${h}_{\mathrm{sep}}$ is separable and ${\lambda}(\mathbf y;\cdot)$ is separable in the transformed coordinates, all mixed second derivatives vanish. Consequently, 
 \[ \nabla^2 \left( L{h}_{\mathrm{sep}} - {\lambda}(\mathbf y;\cdot) \right) = \operatorname{Diag} \left[ \left( L-\frac{y_i}{\alpha_i} \right) {h}_i''({t}_i) \right]_{i=1}^{n}. \] 
 Since each ${h}_i$ is convex, ${h}_i''({t}_i)\geq0$.
 Thus, 
 \begin{itemize}
     \item if $\|\mathbf{y}\|_{\infty} \neq 0$, this matrix is positive semidefinite whenever \[ L \geq \max_{1\leq i\leq n} \frac{y_i}{\alpha_i}. \] 
 Hence, the function ${\lambda}(\mathbf y;\cdot)$ is $L$-smooth relative to ${h}_{\mathrm{sep}}$ for every \[ L\geq \left\| \frac{\mathbf y}{\boldsymbol{\alpha}} \right\|_\infty. \]
     \item if $\|\mathbf{y}\|_{\infty} = 0$, for any $\epsilon >0$, the function ${\lambda}(\mathbf y;\cdot)$ is $L$-smooth relative to ${h}_{\mathrm{sep}}$ for every \[ L\geq \epsilon. \]
 \end{itemize}

We now show that, since each ${h}_i$ is strictly convex, the threshold $ L_{\min}=\|\mathbf y/\boldsymbol\alpha\|_\infty$ is sharp. 
Indeed, let $i^\star \in
\operatorname*{argmax}_{1\leq i\leq n}
\frac{y_i}{\alpha_i}$. If  $ L<\|\mathbf y/\boldsymbol\alpha\|_\infty, $ then $  L-\frac{y_{i^\star}}{\alpha_{i^\star}}<0. $
Choose ${\mathbf{t}}$ and ${\mathbf{t}}^\prime$ that differ only in the $i^\star$-th coordinate. Since ${h}_{i^\star}$ is strictly convex,
$d_{{h}_{i^\star}} ({t}_{i^\star},{t}^\prime_{i^\star})>0 $
whenever ${t}_{i^\star}\neq{t}^\prime_{i^\star}$.
Moreover, since the two points differ only in the $i^\star$-th coordinate,
$
d_{{h}_j}
({t}_{j},{t}^\prime_{j})=0,\, j\neq i^\star.
$
Therefore,
\[
d_{{\zeta}}
({\mathbf{t}},{\mathbf{t}}^\prime)
=
\left(
L-\frac{y_{i^\star}}{\alpha_{i^\star}}
\right)
d_{{h}_{i^\star}}
({t}_{i^\star},{t}^\prime_{i^\star})
<0.
\]
Hence, ${\zeta}$ is not convex.
Thus, no
$L<\|\mathbf y / \boldsymbol{\alpha} \|_\infty$ can satisfy the relative smoothness
condition, which proves that the threshold is sharp. %
At the threshold  $  L_{\min}=\|\mathbf y / \boldsymbol{\alpha} \|_\infty $,
the coefficient associated with each component satisfying
$\frac{y_i}{\alpha_i}=\|\mathbf y/\boldsymbol\alpha\|_\infty$ vanishes, whereas it remains strictly  positive for every component satisfying
$\frac{y_i}{\alpha_i}<\left\|\mathbf y/\boldsymbol\alpha\right\|_\infty$. Consequently, ${\zeta}$ is convex but not strictly convex at the threshold. 

\subsection{Proof of \texorpdfstring{\Cref{Lsmooth_poisson_beta_AnotI}}{Lemma 2}} \label{app:proof_Lsmooth_poisson_beta_AnotI}
Let $\check\lambda(\by;\bt)$ the negative log-likelihood function associated with a Poisson linear inverse problem expressed in the dual coordinates $\bt =\log (\bx)$ induced by a component-wise $\beta$-scaled dual Bregman representation, i.e.,
\begin{align}
    \check\lambda(\mathbf{y};\mathbf{t}) &=-\ell(\by;{\linOperator} e^{\bt})\\
                                         &= d_f(\by;{\linOperator} e^{\bt}) - \phi_f(\by).
\end{align}
where the operator ${\linOperator} \in \R^m\times \R^n$ has non negative entries: $k_{ij}\ge0$ for $1 \le i \le m,\,1 \le j \le n$. Let denote by $s_j$ the sum of its $j^{\text{th}}$ column: $s_j = \sum_{i=1}^m k_{ij}$. Consider the intensity model, for all $1 \le i \le m$
\begin{equation}
u_i(\bt) = [\linOperator \exp \bt]_i = \sum_{j=1}^n k_{ij} e^{t_j} 
\end{equation}
such that the objective function $\check\lambda(\mathbf{y};\mathbf{t})$ writes
\begin{equation}
\check\lambda(\mathbf{y};\mathbf{t})=\sum_{i=1}^m \left( y_i \log y_i - y_i \log u_i(\bt) + u_i(\bt) - y_i \right)- \phi_f(\mathbf{y})
\end{equation}
For $\beta_j>0$, let define $h_\text{sep}(\mathbf{t}) = \sum_{j=1}^n \beta_j e^{t_j}$.
Since $\check\lambda(\by;\cdot)$ and $h$ are $C^2$ on $\R^{n}$, the convexity of $L h_\text{sep}-\check\lambda(\by;\cdot)$ is warranted for any $L > 0$ such that for any $\bv \in \R^n$ and $\bt\in \R^n$,
\begin{equation}
\bv^\top \nabla^2 \check\lambda(\mathbf{y};\bt) \, \bv \le L\, \bv^\top \nabla^2 h_\text{sep}(\bt) \, \bv 
\end{equation}

\noindent \textbf{$h$-Hessian quadratic form.}
Since $h_\text{sep}(\bt)$ is separable, its Hessian is diagonal 
\begin{equation}
\nabla^2 h_\text{sep}(\bt) = \mathrm{diag}\left(\beta_1 e^{t_1}, \dots,\, \beta_n e^{t_n}\right).
\end{equation}
For any vector $\bv \in \R^n$,
\begin{equation}\label{h_quad_form}
\bv^\top \nabla^2 h_\text{sep}(\bt) \, \bv = \sum_{j=1}^n \beta_j\, v_j^2\, e^{t_j}
\end{equation}
\noindent \textbf{$\check\lambda$-Hessian quadratic form.}
Differentiating $\check\lambda(\mathbf{y};\bt) $ twice with respect to $t_j$ and $t_k$ yields the quadratic form, for any $\bv \in \R^n$,
\begin{equation}
	\begin{aligned}
\bv^\top \nabla^2 \check\lambda(\mathbf{y};\bt) \, \bv &= \sum_{j=1}^n v_j^2 e^{t_j}\sum_{i=1}^m k_{ij} - \sum_{j=1}^n v_j^2 e^{t_j} \sum_{i=1}^m \frac{y_i k_{ij}}{u_i(\bt)} + \sum_{i=1}^m \frac{y_i}{u_i(\bt)^2} \left( \sum_{j=1}^n k_{ij}  v_j\exp{t_j} \right)^2\\
&= \sum_{j=1}^n s_j v_j^2 e^{t_j} + \sum_{i=1}^m y_i \left[\left( \sum_{j=1}^n \frac{k_{ij}e^{t_j}}{u_i(\bt)} v_j \right)^2 - \sum_{j=1}^n \frac{k_{ij} e^{t_j}}{u_i(\bt)} v_j^2 \right]
\end{aligned}
\end{equation} 

\noindent \textbf{$\check\lambda$-upper bound using Jensen's inequality.}
For each row $i$, define convex weights $\lambda_{ij} = \frac{k_{ij} e^{t_j}}{u_i(\bt)}$. Since $k_{ij} \ge 0$ and $u_i(\bt) = \sum_{j=1}^n k_{ij} e^{t_j}$, the weights satisfy $\lambda_{ij} \ge 0$ and $\sum_{j=1}^n \lambda_{ij} = 1$. Thus, by Jensen's Inequality,
\begin{equation}
	\left( \sum_{j=1}^n \lambda_{ij} v_j \right)^2 \le \sum_{j=1}^n \lambda_{ij} v_j^2 
\end{equation}
Since $y_i \ge 0$, multiplying by $y_i$ and summing over $i$ yields
\begin{equation}\label{eq:jensen_quad_form}
	\bv^T \nabla^2 \check\lambda(\mathbf y;\bt) \, \bv = \sum_{j=1}^n s_j v_j^2 e^{t_j} + \sum_{i=1}^m y_i \left[ \left( \sum_{j=1}^n \lambda_{ij} v_j \right)^2 - \sum_{j=1}^n \lambda_{ij} v_j^2 \right] \le \sum_{j=1}^n s_j v_j^2 e^{t_j}
\end{equation}

\noindent \textbf{Single weight.} Consider the choice $\beta_j=\beta=1$ for $1\le j\le n$,  such that
\begin{equation}
h_\text{sep}(\bt) = h_{\mathbf 1}(\bt)  = \sum_{j=1}^n e^{t_j}
\end{equation}
Then, \eqref{h_quad_form} writes
\begin{equation}
\bv^\top \nabla^2 h_{\mathbf 1} (\bt) \, \bv = \sum_{j=1}^n v_j^2\, e^{t_j}
\end{equation}
Bounding each column sum $s_j$ by the maximum column sum $\|{\linOperator}\|_1=\max_{1 \le j \le n} s_j$ in \eqref{eq:jensen_quad_form}, we get for all $\bv\in \R^n$,
\begin{equation}
	\bv^\top \nabla^2 \check\lambda(\mathbf{y};\bt)  \, \bv \le  \|{\linOperator}\|_1 \sum_{j=1}^n  v_j^2 e^{t_j} =  \|{\linOperator}\|_1\, \bv^\top \nabla^2 h_{\mathbf 1} (\bt) \, \bv
\end{equation}
Thus, the function  ${\check\lambda}(\mathbf{y};\bt)$ is $L$-smooth relative to ${h}_{\mathbf 1} $ for every $L\geq \|{\linOperator}\|_1$.\\

\noindent \textbf{Multiple weights.} If we choose $\beta_j=s_j$ for $1\le j\le n$, such that
\begin{equation}
h_\text{sep}(\bt) = h_{\mathbf s} (\bt) = \sum_{j=1}^n s_j e^{t_j}
\end{equation}
Then \eqref{h_quad_form} writes
\begin{equation}
\bv^\top \nabla^2 h_{\mathbf s} (\bt) \, \bv = \sum_{j=1}^n s_j v_j^2\, e^{t_j}
\end{equation}
and, from \eqref{eq:jensen_quad_form}, we immediately get that for all $\bv\in \R^n$,
\begin{equation}
	\bv^\top \nabla^2 \check\lambda(\mathbf{y};\bt)  \, \bv \le \bv^\top \nabla^2 h_{\mathbf s} (\bt) \, \bv.
\end{equation}
Thus, the function  ${\check\lambda}(\mathbf{y};\bt)$ is $L$-smooth relative to $h_{\mathbf s} $ for every $L\geq 1$.

\section{Primal Bregman representations of classical exponential family distributions}\label{app:summary_bregman_exponential}

Table~\ref{tab:bregman_families} summarizes the primal Bregman representations of several exponential family distributions widely encountered in the literature. For each model, it reports the minimal sufficient statistic, the expectation parameter, the corresponding Legendre generator $f$, and the associated Bregman divergence $d_f$ whenever it admits a convenient explicit expression in expectation coordinates. These geometries provide the starting point for the dual Bregman representations derived in \Cref{sec:dual_representations}.

\begin{table*}
\centering
\renewcommand{\arraystretch}{2.2}
\resizebox{\textwidth}{!}{%
\begin{tabular}{l|c|c|c|c}
\toprule
\textbf{Distribution}
& \textbf{Statistic $\mathbf y$}
& \textbf{Expectation parameter $\mathbf x$}
& \textbf{Generator $f(\mathbf x)$}
& \textbf{Bregman divergence $d_f(\mathbf y,\mathbf x)$}
\\
\hline

Gaussian $\mathcal{N}(\mu,\textcolor{gray}{\sigma^2})$ 
&
$y=\omega$
&
$x=\mu$
&
$\displaystyle
\frac{x^2}{2\sigma^2}$
&
$\displaystyle
\frac{(y-x)^2}{2\sigma^2}$
\\

Poisson  $\mathcal{P}(\lambda)$ 
&
$y=\omega$
&
$x=\lambda$
&
$\displaystyle
x\log x-x$
&
$\displaystyle
y\log\frac{y}{x}-y+x$
\\

Exponential $\mathcal{E}(b)$ 
&
$y=\omega$
&
$\displaystyle
x=\frac{1}{b}$
&
$\displaystyle
-\log x$
&
$\displaystyle
\frac{y}{x}-1-\log\frac{y}{x}$
\\

Gamma $\mathcal{G}(\textcolor{gray}{k},b)$
&
$y=\omega$
&
$\displaystyle
x=\frac{k}{b}$
&
$\displaystyle
-k\log x$
&
$\displaystyle
k\left(
\frac{y}{x}-1-\log\frac{y}{x}
\right)$
\\

Bernoulli  $\mathcal{B}er(p)$
&
$y=\omega$
&
$x=p$
&
$\displaystyle
x\log x+(1-x)\log(1-x)$
&
$\displaystyle
y\log\frac{y}{x}
+(1-y)\log\frac{1-y}{1-x}$
\\

Binomial   $\mathcal{B}(\textcolor{gray}{k},p)$
&
$y=\omega$
&
$x=kp$
&
$\displaystyle
x\log x+(k-x)\log(k-x)$
&
$\displaystyle
y\log\frac{y}{x}
+(k-y)\log\frac{k-y}{k-x}$
\\

Negative Binomial $\mathcal{NB}(\textcolor{gray}{r},p)$
&
$y=\omega$
&
$\displaystyle
x=\frac{rp}{1-p}$
&
$\displaystyle
x\log x-(r+x)\log(r+x)$
&
$\displaystyle
y\log\frac{y}{x}
+(r+y)\log\frac{r+x}{r+y}$
\\

Inverse-Gaussian $\mathcal{IN}(\mu,\textcolor{gray}{\lambda})$ 
&
$y= \omega $
&
$\displaystyle
x= \mu $
&
$\displaystyle
\frac{\lambda}{2x}  $
&
$\displaystyle
\frac{\lambda}{2} \left(\frac1y- \frac1x + \frac{y-x}{x^2}\right)  $
\\

Gaussian $\mathcal{N}(\mu,{\sigma^2})$ 
&
$\displaystyle
\mathbf y=
\left(
\begin{smallmatrix}
\omega\\
\omega^2
\end{smallmatrix}\right)$
&
$\displaystyle
\mathbf x= \left(
\begin{smallmatrix}
\mu\\
\mu^2+\sigma^2
\end{smallmatrix}\right)$
&
$\displaystyle
-\frac12\log(x_2-x_1^2)$
&
--
\\

Gamma $\mathcal{G}(k,b)$ 
&
$\displaystyle
\mathbf y=
\left(
\begin{smallmatrix}
\log\omega\\
\omega
\end{smallmatrix}\right)$
&
$\displaystyle
\mathbf x=
\left(
\begin{smallmatrix}
\psi(k)-\log b\\
k/b
\end{smallmatrix}\right)$
&
\text{no closed-form}
&
--
\\

Beta $\mathcal{B}e(a,b)$
&
$\displaystyle
\mathbf y=
\left(
\begin{smallmatrix}
\log\omega\\
\log(1-\omega)
\end{smallmatrix}\right)$
&
$\displaystyle
\mathbf x=
\left(
\begin{smallmatrix}
\psi(a)-\psi(a+b)\\
\psi(b)-\psi(a+b)
\end{smallmatrix}\right)$
&
\text{no closed-form}
&
--
\\

\bottomrule
\end{tabular}%
}
\caption{Primal Bregman representations of classical regular exponential family models. Fixed parameters appear in gray.}
\label{tab:bregman_families}
\end{table*}

For the Gamma model with unknown shape and rate parameters and the Beta model with both shape parameters unknown, the Legendre conjugate $f^*$ of the generator (i.e.,  log-partition function) is explicit in natural coordinates, whereas the generator itself $f$ does not admit a simple closed-form expression in expectation coordinates. For the Gaussian model with unknown mean and variance, the Gamma model with unknown shape and rate, and the Beta model with both shape parameters unknown, a single observed sufficient statistic lies on the boundary of the corresponding mean-parameter space. Accordingly, the standard finite Bregman divergence $d_f(\mathbf y,\mathbf x)$ is not reported in the last column.

\section{Dual Bregman representations of classical exponential family distributions: detailed derivations}
\label{appendix:additional-examples}

\subsection{Gaussian distribution with known variance}

\noindent \textbf{$\beta$-scaled dual Bregman representation --}
Consider the Gaussian model with known variance $\sigma^2>0$. In the expectation parametrization, the parameter is the mean $x\in\mathbb R$, and the associated convex generator is
\[
f(x)=\frac{x^{2}}{2\sigma^{2}}, \qquad x\in\mathbb R.
\]
Its Legendre conjugate is
\[
f^{*}(t)=\frac{\sigma^{2}t^{2}}{2},
\qquad
t\in\mathbb R,
\]
while the dual coordinate mapping is
\[
\tau(x)= f'(x)=\frac{x}{\sigma^{2}}.
\]
By \Cref{cor:beta_existence}, the resulting dual Bregman representation of the log-density writes
\begin{align}
    d_f(z,x) &= d_h(\tau(x),\tau(z))\\
&=
\frac{\sigma^{2}}{2}
\left(
\frac{x}{\sigma^{2}}
-
\frac{z}{\sigma^{2}}
\right)^2
=
\frac{(x-z)^2}{2\sigma^{2}}.
\end{align}
Consequently, the primal and dual representations of the density share the same quadratic Bregman geometry, up to the fixed scaling determined by the known variance $\sigma^2$. In this case, the primal and dual generators are both quadratic, reflecting the self-dual character of the Gaussian
geometry under Legendre transformation.\\

\noindent \textbf{$\alpha$-scaled dual Bregman representation --}
  The alternative representation of \Cref{cor:alpha_existence} does not apply here, since its
assumptions require a positive expectation-parameter domain $\mathcal X\subseteq\mathbb R_{+}^*$, whereas the Gaussian mean parameter ranges over $\mathbb R$.

\subsection{Poisson distribution}\label{app:subsec:poisson}
\noindent \textbf{$\beta$-scaled dual Bregman representation --}
Consider the Poisson model, for which the expectation parameter is the
Poisson intensity itself. The corresponding Bregman generator is
\[
f(x)=x\log x-x,
\qquad
x\in\mathbb R_{+}^{*}.
\]
The corresponding Legendre conjugate is
\[
f^{*}(t)=e^{t}
\]
and the dual coordinate mapping is
\[
\tau(x)= f'(x)=\log x.
\]
By \Cref{cor:beta_existence}, the resulting dual Bregman representation of the log-density writes
\begin{align}
    d_f(z,x) &= d_h(\tau(x),\tau(z))=
x-z-z\log\!\left(\frac{x}{z}\right),
\end{align}
which can be identified as the generalized Kullback--Leibler divergence ${\text{KL}}(z\|x)$.\\

\noindent \textbf{$\alpha$-scaled dual Bregman representation --}
Adopting the alternative parametrization of \Cref{cor:alpha_existence} leads to a different dual geometry defined by
\[
{\tau}(x)=x,
\qquad
{h}(t)=-\log t,
\]
leading to
\begin{align}
\frac{1}{z}d_f(z,x) &= d_{{h}}
({\tau}(x),{\tau}(z))=
\frac{x}{z}
-
\log\!\left(\frac{x}{z}\right)
-
1,
\end{align}
which can be identified as  the Itakura-Saito divergence $d_{\mathrm{IS}}(x,z)$. Thus, the alternative parametrization replaces the generalized Kullback-Leibler geometry by the Itakura-Saito geometry while preserving the geometric structure.

\subsection{Exponential distribution}
\noindent \textbf{$\beta$-scaled dual Bregman representation --}
Considering the Exponential model, the expectation parameter belongs to
$\mathbb R_{+}^{*}$ and the associated generator is

\[
f(x)=-\log x.
\]
The corresponding gradient mapping is
\[
\tau(x)
=f'(x)=
-\frac 1 x,
\]
while the Legendre conjugate reads
\[
f^{*}(t)
=
-1-\log(-t),\quad t<0.
\]
\Cref{cor:beta_existence} therefore yields
\[
h(t)
=
-1-\log(-t),
\]
and
\[
d_h(\tau(x),\tau(z))
=
\frac{z}{x}
-
\log\!\left(\frac{z}{x}\right)
-
1,
\]
which is the reverse Itakura-Saito divergence.\\

\noindent \textbf{$\alpha$-scaled dual Bregman representation --}
For the alternative parametrization,
\[
{\tau}(x)=\log x-1,
\qquad
{h}(t)=e^{-t-1},
\]
which gives
\[
d_{{h}}
({\tau}(x),{\tau}(z))
=
\frac1x
-
\frac1z
-
\frac1z
\log\!\left(\frac{z}{x}\right).
\]
It is worth noting that the convex generator $f(x)=-\log x$ also arises for other exponential-family models, including the centered Laplace and Weibull (with known shape parameter $k$) distributions, when expressed in the expectation coordinate associated with their sufficient statistic ($y=|\omega|$ and $y=\omega^k$ respectively). These models therefore inherit the same dual Bregman geometry in the corresponding expectation parametrization.

\subsection{Gamma distribution with known shape parameter}
\noindent \textbf{$\beta$-scaled dual Bregman representation --}
We now consider the Gamma distribution with known shape parameter $k>0$. The
expectation parameter belongs to $\mathbb{R}_{+}^{*}$ and the associated
Bregman generator is 

\[
f(x)=-k\log x, \quad x>0.
\]
 Consequently, the dual coordinates are  given by

\[
\tau(x)
=f'(x)=
-\frac{k}{x},
\]
while the Legendre conjugate is
\[
f^{*}(t)
=
-k\log(-t) +k\log k -k, \qquad t<0.
\]
By \Cref{cor:beta_existence}, $h=f^*$. But since additive constants do not affect Bregman divergences, we may equivalently take
\[
h(t)
=
-k\log(-t),
\]
and the corresponding divergence
\[
d_h(\tau(x),\tau(z))
=
k\left(\frac{z}{x}
-
\log\!\left(\frac{z}{x}\right)
-
1\right).
\]

\noindent \textbf{$\alpha$-scaled dual Bregman representation --}
For the alternative parametrization, ${\tau}(x)= k\log x -k$, ${h}(t) = k\exp(- \frac {t+k} k)$,
	\begin{equation}
		d_{{h}}({\tau}(x),{\tau}(z))= k\left(\frac 1 x - \frac 1 z + \frac 1 z\,\log \frac x z\right).	
        \end{equation}
Hence, the canonical dual geometry of the Gamma family is described by a scaled reverse Itakura-Saito divergence. This example illustrates that exponential-family models whose convex generators differ only by a positive scaling factor share the same underlying Bregman geometry, up to the corresponding rescaling of the divergence.

\subsection{Bernoulli distribution}
\noindent \textbf{$\beta$-scaled dual Bregman representation --}
Consider now the Bernoulli model.
The expectation parameter is the success probability
$x\in(0,1)$
and the corresponding convex generator is

\[
f(x)
=
x\log x
+
(1-x)\log(1-x).
\]
The associated  mapping is the logit transformation
\[
\tau(x)
=
\log x
-
\log(1-x)=\log\frac{x}{1-x},
\]
with Legendre conjugate 
\[
f^{*}(t)
=
\log(1+e^{t}),
\]
Application of \Cref{cor:beta_existence} gives
\begin{align*}
 d_f(z,x) &=   d_h(\tau(x),\tau(z))\\
&=
-z\log x
-
(1-z)\log(1-x)
+
z\log z
+
(1-z)\log(1-z),
\end{align*}
which is the binary cross-entropy, up to an additive constant independent of $x$.\\

\noindent \textbf{$\alpha$-scaled dual Bregman representation --}
For the alternative $\alpha$-scaled parametrization, one has
\[
{\tau}(x)
=
-\log(1-x)
\]
and
\[
{h}(t)
=
-\log(e^{t}-1)
\]
yielding
\begin{align*}
\frac{1}{z} d_f(z,x)&=d_{{h}}
({\tau}(x),{\tau}(z))\\
&=
-\log x
-
\frac{1-z}{z}
\log(1-x)
+
\log z
+
\frac{1-z}{z}
\log(1-z).
\end{align*}
This example illustrates how the proposed construction recovers the geometry underlying the classical Beta-Bernoulli conjugate model adopted in Bayesian inference within the same general framework.

\subsection{Binomial distribution with know trial parameter}
\noindent \textbf{$\beta$-scaled dual Bregman representation --}
Considering the Binomial distribution with a fixed number of trials
$k\in\mathbb N^*$, the expectation parameter is the mean number of successes,
$x\in(0,k)$, and the associated convex generator is

\[
f(x)
=
x\log x
+
(k-x)\log(k-x).
\]
Its gradient is
\[
\tau(x)
=
\log x
-
\log(k-x),
\]
The Legendre conjugate  is
\[
f^{*}(t)
=
k\log(1+e^{t}) - k\log k.
\]
The conjugate generator is therefore 
\[
h(t)
=
k\log(1+e^{t}),
\]
which immediately yields
\[
d_h(\tau(x),\tau(z))
=
-z\log x
-(k-z)\log(k-x) 
+z\log z
+(k-z)\log(k-z).
\]
As in the Bernoulli case, the divergence corresponds to the Binomial cross-entropy, up to an additive constant independent of the unknown
parameter.\\

\noindent \textbf{$\alpha$-scaled dual Bregman representation --}
For the alternative parametrization,
\[
{\tau}(x)
=
-k\log(k-x),
\]
and
\[
{h}(t)
=
-\log\!\left(k e^{t/k}-1\right),
\]
leading to
\[
d_{{h}}
({\tau}(x),{\tau}(z))
=-\log x
-\frac{k-z}{z}\log(k-x) 
+\log z
+\frac{k-z}{z}\log(k-z).
\]
This example shows that the same geometric construction extends naturally from the Bernoulli to the Binomial family, with the number of trials $k$ entering explicitly into the corresponding generators and coordinate mappings.

\subsection{Negative Binomial distribution with known number of failure}
\noindent \textbf{$\beta$-scaled dual Bregman representation --}
Consider now the Negative Binomial distribution with a known number of
failures $r$. The corresponding generator is
\[
f(x)
=
x\log x
-
(r+x)\log(r+x),
\]
defined on $\mathbb R_{+}^{*}$. 
The associated gradient mapping is
\[
\tau(x)
=
\log x
-
\log(r+x),
\]
while its Legendre conjugate is
\[
f^{*}(t)
=
r\log  r-r\log(1-e^{t}),\quad t<0.
\]
By \Cref{cor:beta_existence},
\[
h(t)
=
-r\log(1-e^{t}),
\]
and the induced divergence becomes
\[
d_h(\tau(x),\tau(z))=
-z\log x
+(r+z)\log(r+x) 
+z\log z
-(r+z)\log(r+z).
\]

\noindent \textbf{$\alpha$-scaled dual Bregman representation --}
For the alternative parametrization,
\[
{\tau}(x)
=
r\log(r+x),
\]
and
\[
{h}(t)
=
\frac{t}{r}
-
\log\!\left(e^{t/r}-r\right),
\]
yielding
\[
d_{{h}}
({\tau}(x),{\tau}(z))
=
-\log x
+\frac{r+z}{z}\log(r+x) 
+\log z
-\frac{r+z}{z}\log(r+z).
\]
This example shows that the same Legendre-dual construction applies to the Negative Binomial family, thereby extending the proposed geometric representation to overdispersed count models.

\subsection{Inverse-Gaussian distribution with known shape parameter}\label{app:inverse_gaussian}
The Inverse-Gaussian model with known shape $\lambda$ is a classical non-regular family whose log-partition function is essentially smooth but $\Theta=(-\infty,0]$ is not open \cite[Ch.~8]{barndorff2014information}. According to \Cref{rem:role-assumptions}, all conclusions of \Cref{cor:scaled-representations} do not apply. Nevertheless, expressions of $(h,\tau)$ can still be derived for the $\beta$- and $\alpha$-scaled models. Its expectation parameter is the mean $x \in \mathbb R_+ ^*$ and the corresponding convex generator is
\[
f(x)
=
\frac \lambda {2\,x},
\qquad
x\in\mathbb R_{+}^{*}.
\]
\noindent \textbf{$\beta$-scaled dual Bregman representation --}
The Legendre conjugate of $f$ writes
\[
h(t)=f^{*}(t)=-\sqrt{-2\lambda \,t}
\]
while the dual coordinate mapping is
\[
\tau(x)= f'(x)=-\frac \lambda {2 x^2}.
\]
Application of \Cref{cor:beta_existence} gives
\begin{align*}
 d_f(z,x) &=   d_h(\tau(x),\tau(z))\\
&=\frac{\lambda\, (x-z)^2}{2\,x^2z},
\end{align*}
Note that $h$ is of Legendre type since $f$ is Legendre, but the equality $\dom h=\tau(\X)$ fails: $\tau(\X)=\inte\Theta = (-\infty,0)$ misses the boundary point $0$ of $\dom h=\Theta = (-\infty,0]$.
\\

\noindent \textbf{$\alpha$-scaled dual Bregman representation --}
For the alternative $\alpha$-scaled parametrization, one has
\[
{\tau}(x)
=
-\frac \lambda x.
\]
So $\tau(\X)=(-\infty,0)$, and from \Cref{cor:explicit-construction}
\[
  h(t)=\sup_{x>0}\Bigl(-\frac{t}{x}-\frac{\lambda}{2x^2}\Bigr)=
  \begin{cases}t^2/(2\lambda), & t<0,\\ 0, & t\ge0.\end{cases}
\]

Note that  for $t<0$ the supremum is attained at $x=-\lambda/t=\tau^{-1}(t)$. For $t\ge0$, it is approached as $x\to\infty$, that
is, as the natural parameter $f'(x)=-\lambda/(2x^2)$ tends to the boundary point $0$ of $\Theta$, where $f^*(0)=0$ is
finite. This boundary point, absent in the regular case, is why $\dom h=\R$ is larger than $\tau(\X)$. Moreover, the
restriction of $h$ to $\tau(\X)$ is not essentially smooth (hence not Legendre), since $h'(t)\to0$ as $t\uparrow0$.

Finally, $(h,\tau)$ expressions yield
\begin{align*}
\frac{1}{z} d_f(z,x)&=d_{{h}}
({\tau}(x),{\tau}(z))\\
&=\frac \lambda 2\,\left(\frac 1 x-\frac 1 z\right)^2.
\end{align*}

\subsection{Gaussian distribution}
\noindent \textbf{$\beta$-scaled dual Bregman representation --}
The proposed dual representation of the log-density is not restricted to exponential families defined by a one-dimensional parameter and naturally extends to distributions with vector-parameter. To illustrate this extension, we consider the Gaussian distribution with unknown mean and variance. The minimal
sufficient statistic is
\[
\mathbf y=(\omega,\omega^{2}),
\]
and the corresponding expectation parameter $\mathbf x=(x_{1},x_{2}) \in \mathcal{X} $ is
\[
\mathbf x=(\mathbb E[\omega],\mathbb E[\omega^{2}])
\]
where the parameter space is 
\[ \mathcal{X} = \{ (x_1,x_2) \in \mathbb{R}\times \mathbb{R}_+ : x_2>x_1^2\}.\]
The associated  generator is
\[
f(\mathbf x)
=
-\frac12
\log(x_{2}-x_{1}^{2}).
\]
whose Legendre conjugate is
\begin{equation}\label{eq:gaussian_h}
f^*(\mathbf t)
=
-\frac{t_{1}^{2}}{4t_{2}}
-\frac12
\log(-2t_{2})-\frac{1}{2},\quad t_2<0.
\end{equation}
By \Cref{cor:beta_existence}, the dual representation of the log-density writes
\begin{align*}
d_f(\mathbf z,\mathbf x) &= d_h(\tau(\mathbf x),\tau(\mathbf z))\\
&=
\frac12
\Bigg[
\frac{(x_{1}-z_{1})^{2}}
{x_{2}-x_{1}^{2}}
+
\frac{z_{2}-z_{1}^{2}}
{x_{2}-x_{1}^{2}}
+\log
\frac{x_{2}-x_{1}^{2}}
{z_{2}-z_{1}^{2}}
-1
\Bigg].   
\end{align*}
In the usual parametrization
\[
(\mu,\sigma^{2})
=
(x_{1},\,x_{2}-x_{1}^{2}).
\]
the resulting Bregman divergence has the characteristic quadratic-over-variance structure underlying the classical Normal-Inverse-Gamma conjugate model adopted in Bayesian inference.

\subsection{Gamma distribution}
\noindent \textbf{$\beta$-scaled dual Bregman representation --}
Another example of the extension to distributions with vector-valued parameter is the Gamma distribution when both the shape $a>0$ and rate $b>0$ are unknown. A minimal sufficient statistic is
\[
\mathbf y=(\log\omega,\omega).
\]
The corresponding expectation parameter $\mathbf x=(x_1,x_2)\in \mathcal{X}$ is
\[
\mathbf x=
\bigl(
\mathbb E[\log\omega],
\mathbb E[\omega]
\bigr)
\]
with
\[
x_1=\psi_0(a)-\log b,
\qquad
x_2=\frac{a}{b},
\]
where $\psi_0$ denotes the digamma function and  the expectation-parameter space is defined as
\[
\mathcal X
=
\left\{
(x_1,x_2)\in\mathbb R\times\mathbb R_{+}^{*}
:
x_1<\log x_2
\right\}.
\]
The natural coordinates $\mathbf t=(t_1,t_2)$ are
\[
t_1=a-1>-1,
\qquad
t_2=-b<0.
\]
The Legendre conjugate of the likelihood generator $f$, which
coincides with the log-partition function, is
\begin{equation}\label{eq:gamma_h}
f^*(\mathbf t)
=
\log\Gamma(t_1+1)
-
(t_1+1)\log(-t_2),
\qquad
t_1>-1,\quad t_2<0,    
\end{equation}
with gradient
\[
\nabla f^*(\mathbf t)
=
\left(
\psi_0(t_1+1)-\log(-t_2),\,
-\frac{t_1+1}{t_2}
\right)^T.
\]
The likelihood generator $f$ and the dual coordinate mapping $\tau=\nabla f$ cannot be expressed with elementary closed forms in expectation coordinates, since inversion of the mean map requires solving a (scalar) nonlinear equation involving the digamma function. Indeed, since $b=a/x_2$, the natural coordinates $\mathbf t=(t_1,t_2)$ can be rewritten as functions of $\mathbf x\in\mathcal X$ as
\[
t_1=a(\mathbf{x})-1,
\qquad
t_2=-\frac{a(\mathbf{x})}{x_2}
\]
where $a(\mathbf x)>0$ denotes the unique
solution of
\begin{equation}\label{eq:gamma_nonlinear_equation}
    \psi_0(a)-\log a = x_1-\log x_2.
\end{equation}
Hence,
\[
\tau(\mathbf x)
= \left(a(\mathbf x)-1,\,-\frac{a(\mathbf x)}{x_2}\right)^T
\]
and, by Legendre duality $f(\mathbf x) = \langle\mathbf t,\mathbf x\rangle-f^*(\mathbf t)$, 
\[
f(\mathbf x)
=
(a(\mathbf x)-1)x_1-a(\mathbf x)-\log\Gamma(a(\mathbf x)) + a(\mathbf x)\log\left(\frac{a(\mathbf x)}{x_2}\right).
\]
Thus, both $\tau(\mathbf x)$ and $f(\mathbf x)$ can be evaluated once the scalar equation \eqref{eq:gamma_nonlinear_equation} defining $a(\mathbf x)$ has been solved. \Cref{cor:beta_existence} gives
\[
h(\mathbf t)=f^*(\mathbf t)
=
\log\Gamma(t_1+1)
-
(t_1+1)\log(-t_2),
\]
and consequently
\[
d_h\bigl(\tau(\mathbf x),\tau(\mathbf z)\bigr)
=
d_{f^*}\bigl(\tau(\mathbf x),\tau(\mathbf z)\bigr)
=
d_f(\mathbf z,\mathbf x).
\]
Although $d_h$ can be evaluated explicitly once the corresponding
natural coordinates are known, its evaluation from expectation
coordinates requires the numerical computation of
$\tau(\mathbf x)$ and $\tau(\mathbf z)$.\\

\noindent \textbf{$\alpha$-scaled dual Bregman representation --}
    The alternative representation of \Cref{cor:alpha_existence} does not apply directly to this
model, since that result is restricted to one-dimensional positive expectation parameters.

\subsection{Beta distribution}
\noindent \textbf{$\beta$-scaled dual Bregman representation --}
We finally consider the Beta distribution when both shape parameters $a$ and $b$ are
unknown.
A minimal sufficient statistic is
\[
\mathbf y=
\bigl(
\log\omega,
\log(1-\omega)
\bigr),
\qquad
\omega\in(0,1).
\]
The corresponding expectation parameter is
\[
\mathbf x=(x_1,x_2)
=
\bigl(
\mathbb E[\log\omega],\,
\mathbb E[\log(1-\omega)]
\bigr).
\]
For the Beta distribution,
\[
x_1=\psi_0(a)-\psi_0(a+b),
\qquad
x_2=\psi_0(b)-\psi_0(a+b),
\]
where $\psi_0$ denotes the digamma function.

The expectation-parameter space is
\footnote{For any mapping $g:\mathcal A\to\mathcal B$, we denote by
$\operatorname{ran}g=\{g(a):a\in\mathcal A\}$ the range of $g$.}
\[
\mathcal X
=
\operatorname{ran}\nabla f^*
\subset
(\mathbb R_{-}^{*})^2.
\]
A useful necessary condition for membership in $\mathcal X$ follows  from Jensen's inequality. Indeed,
\[
e^{x_1}
=
\exp\bigl(\mathbb E[\log\omega]\bigr)
<
\mathbb E[\omega],
\qquad
e^{x_2}
=
\exp\bigl(\mathbb E[\log(1-\omega)]\bigr)
<
\mathbb E[1-\omega],
\]
and therefore
\[
e^{x_1}+e^{x_2}<1.
\]

The natural coordinates are
\[
\mathbf t=(t_1,t_2),
\qquad
t_1=a-1>-1,
\qquad
t_2=b-1>-1.
\]
The Legendre conjugate of the likelihood generator $f$, which
coincides with the log-partition function, is
\begin{equation}\label{eq:beta_h}
f^*(\mathbf t)
=
\log\Gamma(t_1+1)
+
\log\Gamma(t_2+1)
-
\log\Gamma(t_1+t_2+2),
\end{equation}
with
\[
\inte(\dom f^*)
=
(-1,+\infty)^2.
\]
Its gradient is
\[
\nabla f^*(\mathbf t)
=
\left(
\psi_0(t_1+1)-\psi_0(t_1+t_2+2),
\,
\psi_0(t_2+1)-\psi_0(t_1+t_2+2)
\right).
\]

The likelihood generator $f$ and the dual coordinate mapping
$\tau=\nabla f$ do not admit simple closed-form expressions directly
in expectation coordinates. More precisely, for
$\mathbf x\in\mathcal X$, let $(a(\mathbf x),b(\mathbf x))$ denote
the unique solution in $(0,+\infty)^2$ of
\[
\begin{cases}
\psi_0(a)-\psi_0(a+b)=x_1,\\[1mm]
\psi_0(b)-\psi_0(a+b)=x_2.
\end{cases}
\]
Existence and uniqueness follow from the Legendre property of $f^*$,
since $\nabla f^*$ is a bijection from the natural-parameter space
onto $\mathcal X$. Hence,
\[
\tau(\mathbf x)
=
\nabla f(\mathbf x)
=
(\nabla f^*)^{-1}(\mathbf x)
=
\bigl(
a(\mathbf x)-1,\,
b(\mathbf x)-1
\bigr),
\]
which must generally be evaluated numerically.

Similarly, Legendre duality gives
\[
f(\mathbf x)
=
\langle\mathbf t,\mathbf x\rangle-f^*(\mathbf t),
\]
and therefore
\[
\begin{aligned}
f(\mathbf x)
={}&
(a-1)x_1+(b-1)x_2
-\log\Gamma(a)-\log\Gamma(b)
+\log\Gamma(a+b),
\end{aligned}
\]
where
\[
a=a(\mathbf x),
\qquad
b=b(\mathbf x).
\]
Thus, $f$ is explicit once $(a(\mathbf x),b(\mathbf x))$ has been
determined.

For $\beta=1$, \Cref{cor:beta_existence} gives
\[
h(\mathbf t)=f^*(\mathbf t)
=
\log\Gamma(t_1+1)
+
\log\Gamma(t_2+1)
-
\log\Gamma(t_1+t_2+2),
\]
and consequently
\[
d_h\bigl(\tau(\mathbf x),\tau(\mathbf z)\bigr)
=
d_{f^*}\bigl(\tau(\mathbf x),\tau(\mathbf z)\bigr)
=
d_f(\mathbf z,\mathbf x).
\]
Although the divergence is explicit in natural coordinates, its
evaluation in expectation coordinates requires the numerical
computation of $\tau(\mathbf x)$ and $\tau(\mathbf z)$.\\

\noindent \textbf{$\alpha$-scaled dual Bregman representation --}
The alternative representation of
\Cref{cor:alpha_existence} does not apply directly to this
model, since that result is restricted to one-dimensional positive
expectation parameters.

\bibliographystyle{plainnat}
\bibliography{bregman_conjugacy}

\end{document}